\documentclass[journal]{IEEEtran}
\usepackage{epsfig}
\usepackage{graphicx}
\usepackage{amsmath}
\usepackage{amssymb}
\usepackage{multirow}
\usepackage{cite}
\usepackage{subfigure}
\usepackage{adjustbox}
\usepackage{booktabs}
\usepackage[table,xcdraw]{xcolor}
\usepackage[normalem]{ulem}

\useunder{\uline}{\ul}{}
\ifCLASSINFOpdf
\else
\fi
\begin{document}
%
\title{Snow Mask Guided Adaptive Residual Network for Image Snow Removal}

\author{Bodong Cheng, Juncheng Li*, Ying Chen*, Shuyi Zhang, and Tieyong Zeng

\thanks{\textsuperscript{$\ast$}Corresponding author}
\thanks{B. Cheng is with the School of Computer Science and Technology, Xidian University, Xian, Shanxi 710071, China, and the Department of Cyberspace Security, Beijing Electronic Science an Technology Institute, Beijing, China. (E-mail:bdcheng@stu.xidian.edu.cn)}
\thanks{J. Li, and T. Zeng are with the Department of Mathematics, The Chinese University of Hong Kong, New Territories, Hong Kong. (E-mail: cvjunchengli@gmail.com, zeng@math.cuhk.edu.hk)}
\thanks{Y. Chen is with the Department of Cyberspace Security, Beijing Electronic Science and Technology Institute, Beijing, China. (E-mail:ychen@besti.edu.cn)}
\thanks{S. Zhang is with Guangxi Key Lab of Multi-source Information Mining and Security, Guangxi Normal University, Guilin 541004, China. (E-mail: zhangshuyi@stu.gxnu.edu.cn)}
}

\markboth{IEEE Transactions}
{Shell \MakeLowercase{\textit{et al.}}: Bare Demo of IEEEtran.cls for IEEE Journals}
%



\maketitle

\begin{abstract}
Image restoration under severe weather is a challenging task. Most of the past works focused on removing rain and haze phenomena in images. However, snow is also an extremely common atmospheric phenomenon that will seriously affect the performance of high-level computer vision tasks, such as object detection and semantic segmentation. Recently, some methods have been proposed for snow removing, and most methods deal with snow images directly as the optimization object. However, the distribution of snow location and shape is complex. Therefore, failure to detect snowflakes / snow streak effectively will affect snow removing and limit the model performance. To solve these issues, we propose a Snow Mask Guided Adaptive Residual Network (SMGARN). Specifically, SMGARN consists of three parts, Mask-Net, Guidance-Fusion Network (GF-Net), and Reconstruct-Net. Firstly, we build a Mask-Net with Self-pixel Attention (SA) and Cross-pixel Attention (CA) to capture the features of snowflakes and accurately localized the location of the snow, thus predicting an accurate snow mask. Secondly, the predicted snow mask is sent into the specially designed GF-Net to adaptively guide the model to remove snow. Finally, an efficient Reconstruct-Net is used to remove the veiling effect and correct the image to reconstruct the final snow-free image. Extensive experiments show that our SMGARN numerically outperforms all existing snow removal methods, and the reconstructed images are clearer in visual contrast. All codes will be available.
\end{abstract}


%
\IEEEpeerreviewmaketitle

\section{Introduction}
As a common atmospheric phenomenon, snow is often inevitably captured in images by camera lenses, which will affect the accuracy of high-level computer vision tasks such as image classification~\cite{ramachandran2019stand, dosovitskiy2020image}, object detection~\cite{carion2020end, bochkovskiy2020yolov4}, and facial recognition~\cite{deng2020retinaface, gao2020hierarchical, hu2021orthogonal}. Different from other weather phenomena, snow is more complex that includes opaque and translucent snowflakes and snow streaks, and will cause veiling effects. According to previous work~\cite{chen2021all}, images affected by snow can be modeled as
\begin{equation}
    I(x) = K(x)T(x) + A(x)(1 - T(x)),
\end{equation}
where $I(x)$ denotes the snowy image, $T(x)$ is the media transmission, and $A(x)$ is the atmospheric light. Meanwhile, $K(x)$ represents a snow scene image without veiling effect, which can be obtained by the following formula
\begin{equation}
    K(x) = J(x)(1-Z(x)R(x)) + C(x)Z(x)R(x), 
\end{equation}
where $J(x)$ is the scene radiance, $R(x)$ is a binary mask that presents the snow location information, $C(x)$ and $Z(x)$ are the chromatic aberration map for the snow image and the snow mask, respectively.

\begin{figure}[t]
\centering
\begin{minipage}[c]{0.23\textwidth}
\includegraphics[width=4.1cm, height=3.2cm]{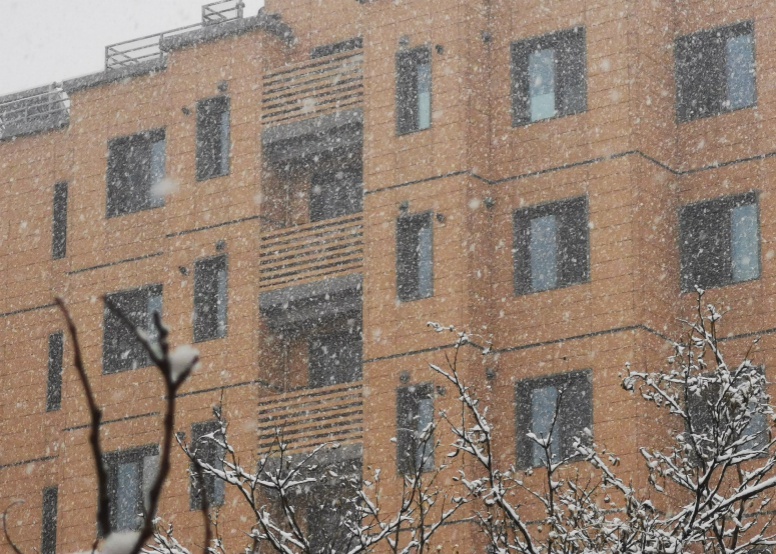}
\centerline{Snowy Image}
\centerline{}
\includegraphics[width=4.1cm, height=3.2cm]{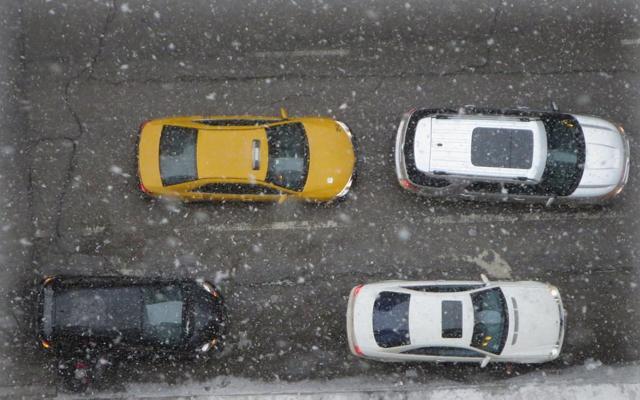}
\centerline{Snowy Image}
\centerline{}
\end{minipage}
\begin{minipage}[c]{0.23\textwidth}
\includegraphics[width=4.1cm, height=3.2cm]{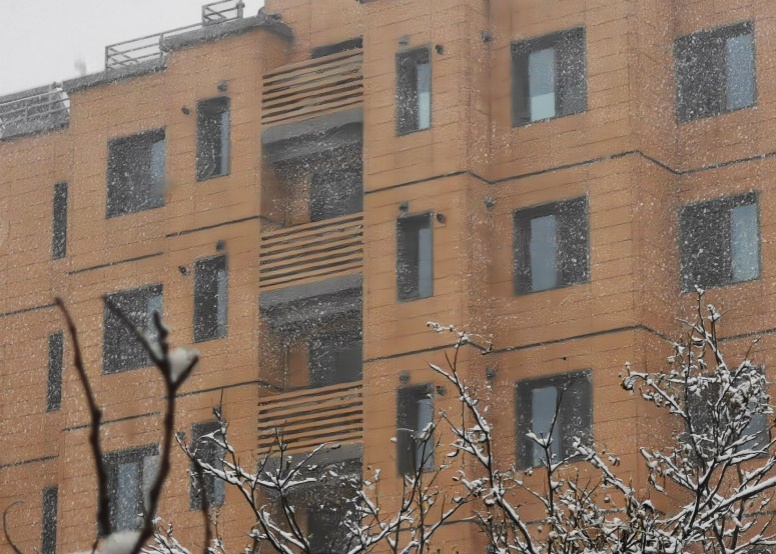}
\centerline{SMGARN (Ours)}
\centerline{}
\includegraphics[width=4.1cm, height=3.2cm]{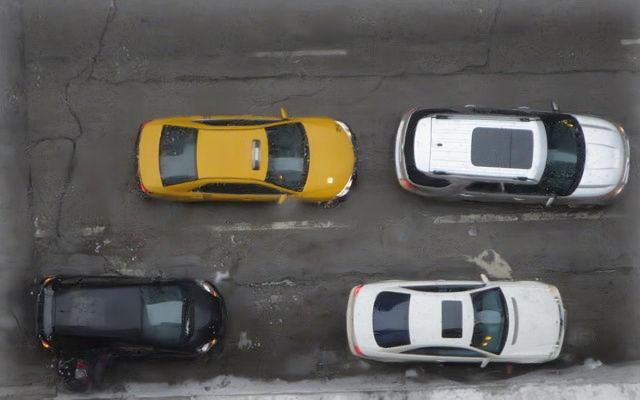}
\centerline{SMGARN (Ours)}
\centerline{}
\end{minipage}
\caption{Examples of real image snow removal by our SMGARN. All of these snowy images are real images captured by cameras, which contains different shapes and sizes of snowflakes / snow streaks. It can be see that our SMGARN can reconstruct high-quality snow-free images in real scenes.} 
\label{Examples}
\end{figure}

In the past, in order to better deal with the problem of single image desnowing, many methods~\cite{xu2012improved, zheng2013single, pei2014removing, wang2017hierarchical, yu2014content} have been proposed with manually extracting features. However, most of them rely on human intuition and do not have the ability to learn the deep features of images, so the snow removal performance is limited and the generalization ability is weak. Recently, with the powerful feature extraction capabilities of convolutional neural networks (CNNs), more and more CNN-based image desnowing methods~\cite{liu2018desnownet, chen2020jstasr, chen2021all} have been proposed. Among them, Liu et al.~\cite{liu2018desnownet} proposed the first CNN-based image snow removal method, called DesnowNet. It produces snow-free images by sequentially processing complex translucent and opaque snow particles. Chen et al.~\cite{chen2020jstasr} took the removal of the veiling effect as part of the image snow removal and inpainted the image. In order to better guide the model to learn the information of snow particles and pay more attention to the heavy snow. Jaw et al.~\cite{jaw2020desnowgan} proposed an efficient modular snow removal network, and introduced a generative adversarial network (GAN) to further improve the snow removal ability. Although JSTASR also predicts snow mask as part of joint snow removal, its performance is not excellent and the number of model parameters is huge. Chen et al.~\cite{chen2021all} proposed a dual-tree complex wavelet transform-based image snow removal model, named HDCWNet, and used contradictory channel loss to improve the desnowing performance. Using Dual-Tree Complex Wavelet Transform (DTCWT) to locate high-frequency snow information in images enables HDCWNet to obtain better snow removal results. However, wavelet transform cannot accurately distinguish snow from other high-frequency information in the original image, so part of useful information will also be removed. Moreover, the loss of information in the up-sampling process of wavelet transform will inevitably result in the change of the information of these sub-bands, which will affect the quality of the reconstructed images.


According to our research on the generation mechanism of snow images, we found that the distribution of snow can be predicted through a specially designed network. Meanwhile, we believe that the predicted snow mask can accurately reflect the distribution of snow in the image and the reconstructed images guided by snow mask will have less snow residue. To achieve this, we propose a Snow Mask Guided Adaptive Residual Network (SMGARN) for image snow removal. Specifically, SMGARN adopts an end-to-end modular design, and the overall model consists of three parts: Mask-Net, Guidance-Fusion Network (GF-Net), and Reconstruct-Net. Among them, Mask-Net is proposed to predict the snow mask, GF-Net is specially designed to adaptively remove snow with the guidance of the learned snow mask, and Reconstruct-Net is used to suppress the veiling effect and reconstruct the final snow-free image. In summary, the main contributions of this paper are as follows
\begin{itemize}
\item We build an efficient Mask-Net to directly predict the snow mask from the snowy image. With the help of Self-pixel Attention (SA) and Cross-pixel Attention (CA), Mask-Net can capture the features of snowflakes and accurately localize the shape and location of snow.
\item We design a multi-level Guidance-Fusion Network (GF-Net) to adaptively remove snow from the image with the guidance of the learned snow mask. 
\item We design a Reconstruct-Net to reconstruct the final snow-free image by the specially designed multi-scale aggregated residual blocks. 
\item We propose a novel Snow Mask Guided Adaptive Image Residual Network (SMGARN) for Image Snow Removal. Meanwhile, we construct a new real snow image test dataset, named SnowWorld24, which contains 24 different scenes of real snow images from all over the world. 
\end{itemize}

The rest of this paper is organized as follows. Related works are reviewed in Section~\ref{RW}. A detailed explanation of the proposed SMGARN is given in Section~\ref{MN}. The experimental results, ablation studies, and discussion are presented in Section~\ref{EX}, ~\ref{AS}, and ~\ref{DS} respectively. Finally, we draw a conclusion in Section~\ref{DS}.

\section{Related Works}~\label{RW}
Images are easily disturbed by atmospheric phenomena such as rain, fog, and snow during the acquisition process. This will contaminate and blur the acquired images, thus affecting subsequent analysis of the image. Therefore, designing efficient models to restore images distorted by extreme atmospheric phenomena into clear images is necessary.

\subsection{Single Image Dehazing}
Early image dehazing algorithms~\cite{tan2008visibility, chen2016edge, fattal2008single, fattal2014dehazing, wang2021tms} took the physical properties of hazy images, such as contrast, reflectivity, etc., as research objects, to remove haze in images. For example, He et al.~\cite{he2010single} analyzed a large number of haze-free images and proposed a well-known image dehazing algorithm guided by dark channel priors. After that, a considerable part of works take the dark channel as the research focus of image dehazing~\cite{huang2014efficient, chen2015advanced}. However, due to the lack of generalization ability of manual setting priors, the generality of these methods is greatly limited. Recently, many deep learning-based methods~\cite{zhang2018densely, cai2016dehazenet, ren2016single, berman2016non, zhang2018density, li2018single, he2016deep, mei2018progressive, ancuti2018ntire, yi2021efficient} have been proposed for image dehazing. For instance, Cai et al.~\cite{cai2016dehazenet} proposed the first learning-based image dehazing model called Dehazenet. Zhang et al.~\cite{zhang2018densely} proposed a Densely Connected Pyramid Dehazing Network (DCPDN) that can simultaneously estimate the transmission map and atmospheric light intensity. Although these models have achieved promising results, they cannot play the role of image snow removal task since the haze phenomenon is quite different from the snow phenomenon. Image dehazing methods generally assume haze particles to be translucent and that there are no fully occluded opaque regions. This is not useful for snow removal since some snowflakes completely cover parts of the image and are difficult to remove. 

\subsection{Single Image Deraining}
Traditional image deraining methods usually use gradient features~\cite{kang2011automatic} or sparse coding~\cite{luo2015removing, son2016rain, chen2014visual} to reconstruct rain-free images. However, these methods have limited performance and applicability. Recently, deep learning-based ~\cite{fu2017clearing, yasarla2019uncertainty, wei2019semi, jiang2020multi, yang2017deep, yi2021structure} also promoted the development of image deraining. For example, Fu et al.~\cite{fu2017clearing} proposed a deep neural network-based image deraining method, called DerainNet, which train the model on the high-pass layer rather than in the image domain. Yang et al.~\cite{yang2017deep} proposed a recurrent deep network to progressively remove rain streaks. Yi et al.~\cite{yi2021structure} proposed a Structure Preserving Deraining Network (SPDNet) with the guidance of the residue channel prior. Although rain images are similar to snow images, directly using existing rain removal methods cannot completely remove them from images due to the diverse states of snow.

\begin{figure*}[t]
\begin{center}
\includegraphics[width=0.95\linewidth]{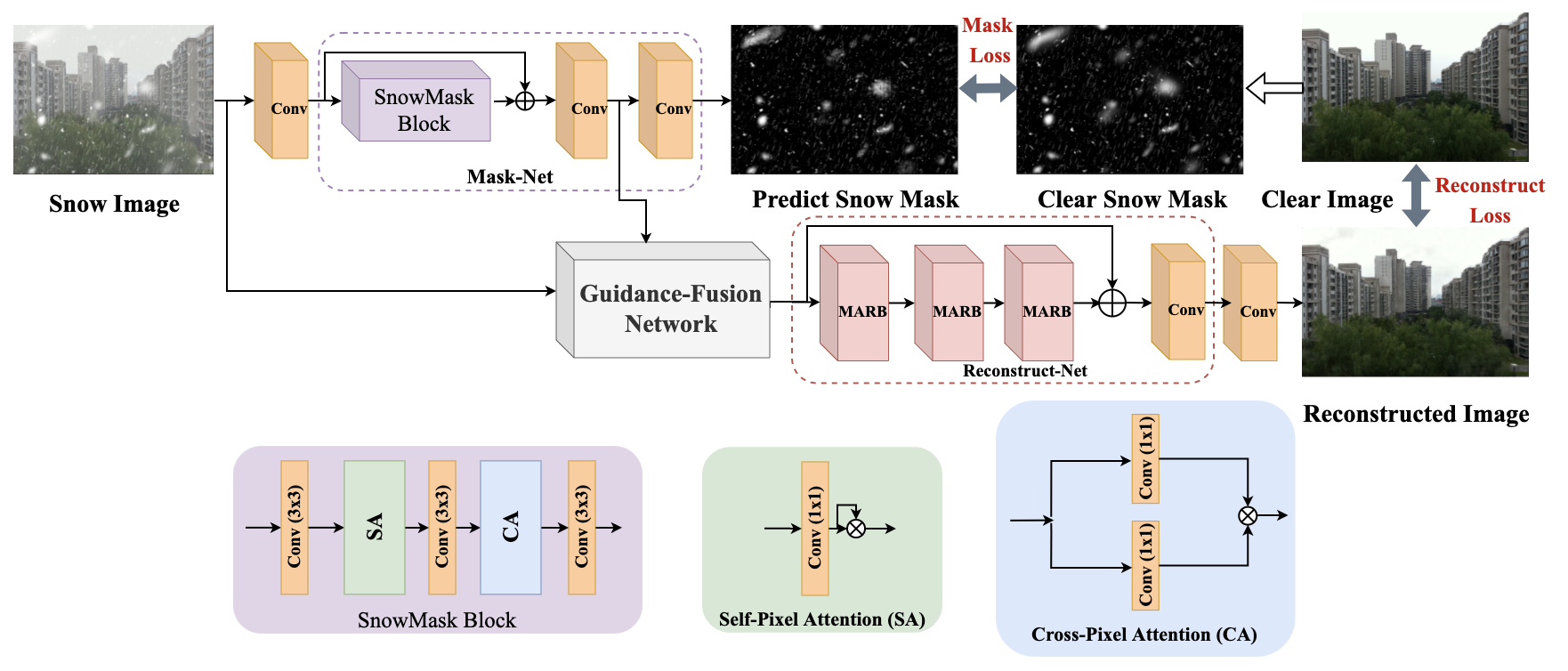}
\end{center}
\caption{The complete architecture of the proposed Snow Mask Guided Adaptive Residual Network (SMGARN). SMGARN consists of three parts: Mask-Net, GF-Net, and Reconstruct-Net.} 
\label{model}
\end{figure*}

\subsection{Single Image Snow Removal}
As another representation of atmospheric phenomena, the changes of snow are more complex, and the spatial states are more abundant. Traditional image snow removal methods still use feature priors to model snow particle information, such as histogram of gradients (HOG)~\cite{bossu2011rain, pei2014removing}, frequency separation ~\cite{rajderkar2013removing}, etc. However, these methods cannot guarantee the generalization of snow removal and usually have poor performance. In order to further improve the ability of snow removal, Liu et al.~\cite{liu2018desnownet} proposed the first deep neural network-based snow removal method, called DesnowNet. DesnowNet adopts the multi-scale pyramid model of Inception-v4 as the backbone and performs well in its proposed Snow100K dataset. Chen et al.~\cite{chen2020jstasr} proposed a model named JSTASR that takes into account the veiling effect and opaque snow particles. It removes the effects caused by snow phenomena using convolution and slightly darker channel priors. After that, Chen et al.~\cite{chen2021all} proposed a hierarchical network based on the Dual-tree Complex Wavelet Transform (DTCWT), named HDCWNet, which solved the problem that JSTASR cannot remove snow patterns and color distortion.

Although HDCWNet has achieved state-of-the-art results, we observe that it will produces blurry images. This is because the downsampling unit is excessively used in HDCWNet to remove snowflakes and snow streaks. With the help of downsampling operations, snowflakes and snow streaks can be effectively removed, but also cause the loss of useful information, which is not conducive to reconstructing clear images. In this work, we aim to explore a more effective snow removal method that can remove snow from the image while preserving the quality of the image.

\section{Methodology}~\label{MN}
Predicting an accurate snow mask can help the model to accurately locate and remove snow. More importantly, this method will not mistakenly remove the original information of the image as snow, which greatly improves the quality of reconstructed image. To achieve efficient snow removal, we propose a Snow Mask Guided Adaptive Residual Network (SMGARN). As shown in Fig.~\ref{model}, SMGARN consists of three parts: Mask-Net, Guidance-Fusion Network (GF-Net), and Reconstruct-Net. Among them, Mask-Net is specially designed to predict the snow mask of the snowy image. With the help of Self-pixel Attention (SA) and Cross-pixel Attention (CA) mechanisms, Mask-Net can quickly and accurately capture snowflakes and snow streaks in the image, thus can predict accurate snow mask. However, snow affects different regions of the image differently. Directly using snow masks and snow images to do residuals will generate a large number of dark areas, which will affect the quality of the reconstructed images. Therefore, we propose a multi-level Guidance-Fusion Network (GF-Net) to adaptively remove snow with the guidance of the predicted snow mask. Finally, in order to remove the haze phenomenon caused by the veiling effect, we design a multi-scale based Reconstruct-Net to achieve the effect of dehazing and further correct the area affected by snow. Each sub-network will be described in detail in the following chapters.

\subsection{Mask-Net}
As we all know, the key for image snow removal is to accurately capture snowflakes and snow streaks in the image. In HDCWNet~\cite{chen2021all}, the authors use wavelet transform to decompose the high and low-frequency information to model snow. However, this method will confuse the information of non-snow targets in the image, and training with such features will affect the final result. To address this issue, we propose a mask prediction network (Mask-Net) to directly predict snow mask by using of the convolutional neural network. Specifically, Mask-Net ($\mathcal{M}$) takes the snowy image $\mathcal{I}_{snowy}$ as input and output the snow mask $m$ and the mask feature map $\mathcal{F}_{mask}$
\begin{equation}
    \mathcal{F}_{mask}, m = \mathcal{M}(\mathcal{I}_{snowy}).
\end{equation}

As shown in Fig.~\ref{model}, the backbone of Mask-Net consists of one SnowMaskBlock and two convolutional layers. In SnowMaskBlock, we design two kinds of pixel attention units, namely Self-pixel Attention (SA) unit and Cross-pixel Attention (CA) unit. It should be noted that the attention used in this paper is different from the operation performed by constructing a similarity matrix in the self-attention mechanism. The pixel attention we use here aims to enhance the activations of snow-covered regions in images with snow to help the model better extract snow features.
Furthermore, we add residual connections on SnowMaskBlock to facilitate feature expression. Among them, SA is used to enhance the representation of important features and suppress unimportant features in the image. CA adopts two different encoding functions to improve the adaptability of the network. The operations of CA and SA can be defined as follows
\begin{equation}
    \mathcal{SA}(\mathcal{X}) = f_{0}(\mathcal{X})\odot f_{0}(\mathcal{X}), \label{self_pixel attention}
\end{equation}
\begin{equation}
    \mathcal{CA}(\mathcal{X}) = f_{1} (\mathcal{X})\odot f_{2}(\mathcal{X}), \label{cross_pixel attention}
\end{equation}
where $\mathcal{SA}(\cdot)$ and $\mathcal{CA(\cdot)}$ denote the Self-pixel Attention and the Cross-pixel Attention, respectively. $\mathcal{X}$ represents the feature matrix. $f_{0}(\cdot)$, $f_{1}(\cdot)$ and $f_{2}(\cdot)$ represent the encoding functions, and $\odot$ represents the Hadamard product operation. To adequately encode features to extract deep information, we replace the encoding functions in Eq.(\ref{self_pixel attention}) and Eq.(\ref{cross_pixel attention}) with convolutional layers
\begin{equation}
    \mathcal{SA}(\mathcal{X}) =  Conv_{0}(\mathcal{X})\odot Conv_{0}(\mathcal{X}),
\end{equation}
\begin{equation}
    \mathcal{CA}(\mathcal{X}) =  Conv_{1}(\mathcal{X})\odot Conv_{2}(\mathcal{X}). 
\end{equation}
Specifically, we first square each element in the feature matrix, which will highlight regions with high activation value (regions with snow) and suppress regions with low activation (regions without snow). To further enhance the representation of features, we operate on the feature matrix through two parallel convolutional units, and then use the Hadamard porduct to intersect the two features.

The Mask-Net can be used as part of any image snow removal model to provide the snow mask or works independently to predict snow mask from the snowy image directly. The goal of Mask-Net is to learn a prediction function that can predict an accurate snow mask from the corresponding snowy input. To achieve this, we apply a mask loss $\mathcal{L}_{mask}$ on Mask-Net
\begin{equation}
    \mathcal{L} _{mask} = \left \| \mathcal{M}(\mathcal{I}_{snowy}) -  \mathcal{I}_{mask} \right \|_{1}, 
\end{equation}
where $\mathcal{M}(\cdot)$ denotes the Mask-Net, $\mathcal{M}(\mathcal{I}_{snowy})$ represents the predicted mask, and $\mathcal{I}_{mask}$ is the corresponding ground-truth snow mask. It is worth noting that we embed the Mask-Net as part of the SMGARN to provide the snow mask prior for snow-free images reconstruction in this work.

\subsection{Guidance-Fusion Network (GF-Net)}
\begin{figure}[t]
\begin{center}
\includegraphics[width=1\linewidth]{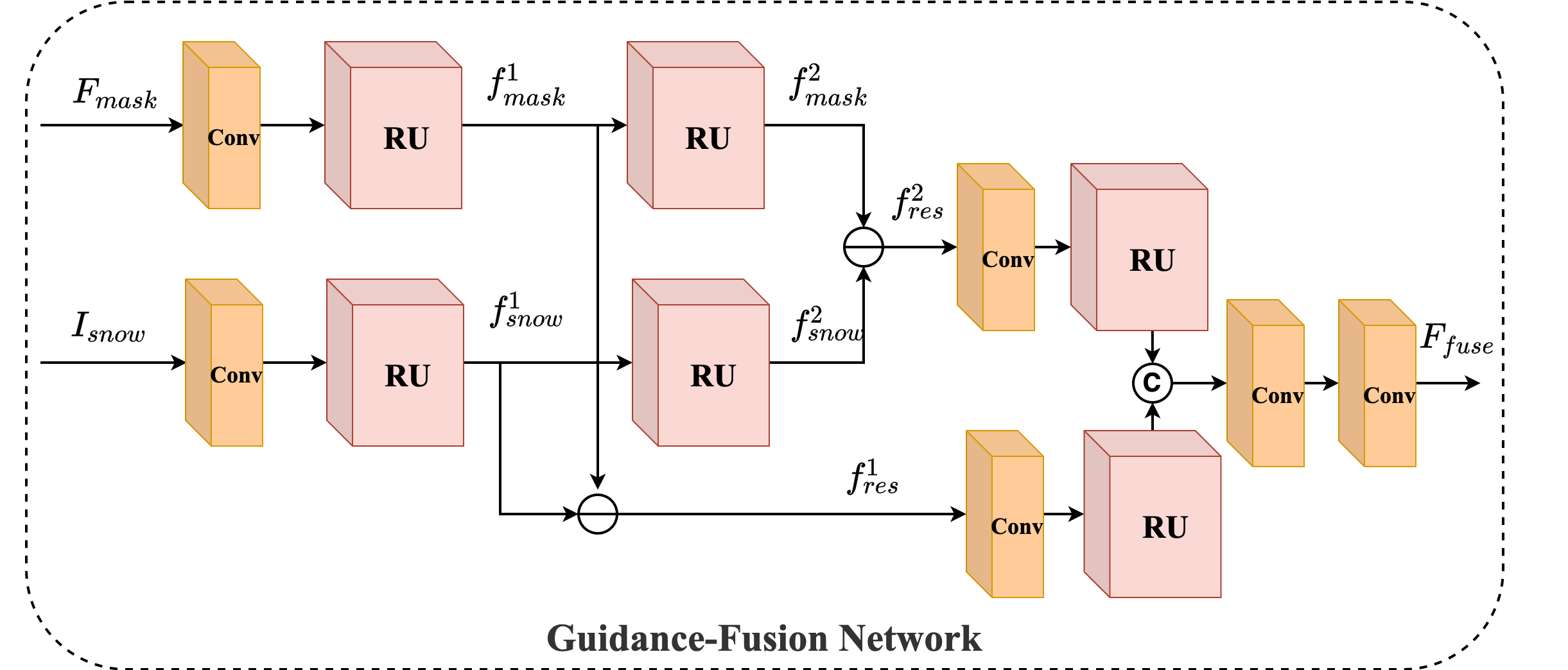}
\end{center}
\caption{The complete architecture of the proposed Guidance-Fusion Network.} 
\label{gfn}
\end{figure}

As an important bridge connecting snow images and snow masks, the proposed Guidance-Fusion Network (GF-Net, Fig.~\ref{gfn}) plays a crucial role in improving the snow removal performance of the model. In GF-Net, we design a multi-level residual network to fully utilize the predicted snow mask to guide the model for snow removing. Specifically, we found that subtracting the snow image with the adaptively weighted snow mask (the $\ominus$ symbol in Fig.~\ref{gfn} represents the residual operation) can effectively remove snow streaks. Meanwhile, we observe that it is impossible to remove all snow by directly applying the predicted snow mask to the snowy image as a residual. The reason is that the snowy image is not a simple addition of the snow mask and the clear image, and the predicted snow mask is just an enhanced schematic diagram of the position and shape of the snow in the image. Therefore, GF-Net adopts an adaptive way to guide the model to remove snow according to the information provided by the snow mask. In addition, to better preserve the features of the snow mask, we use the feature map of the snow mask as input rather than the snow mask itself.

\begin{figure}[t]
\begin{center}
\includegraphics[width=0.8\linewidth]{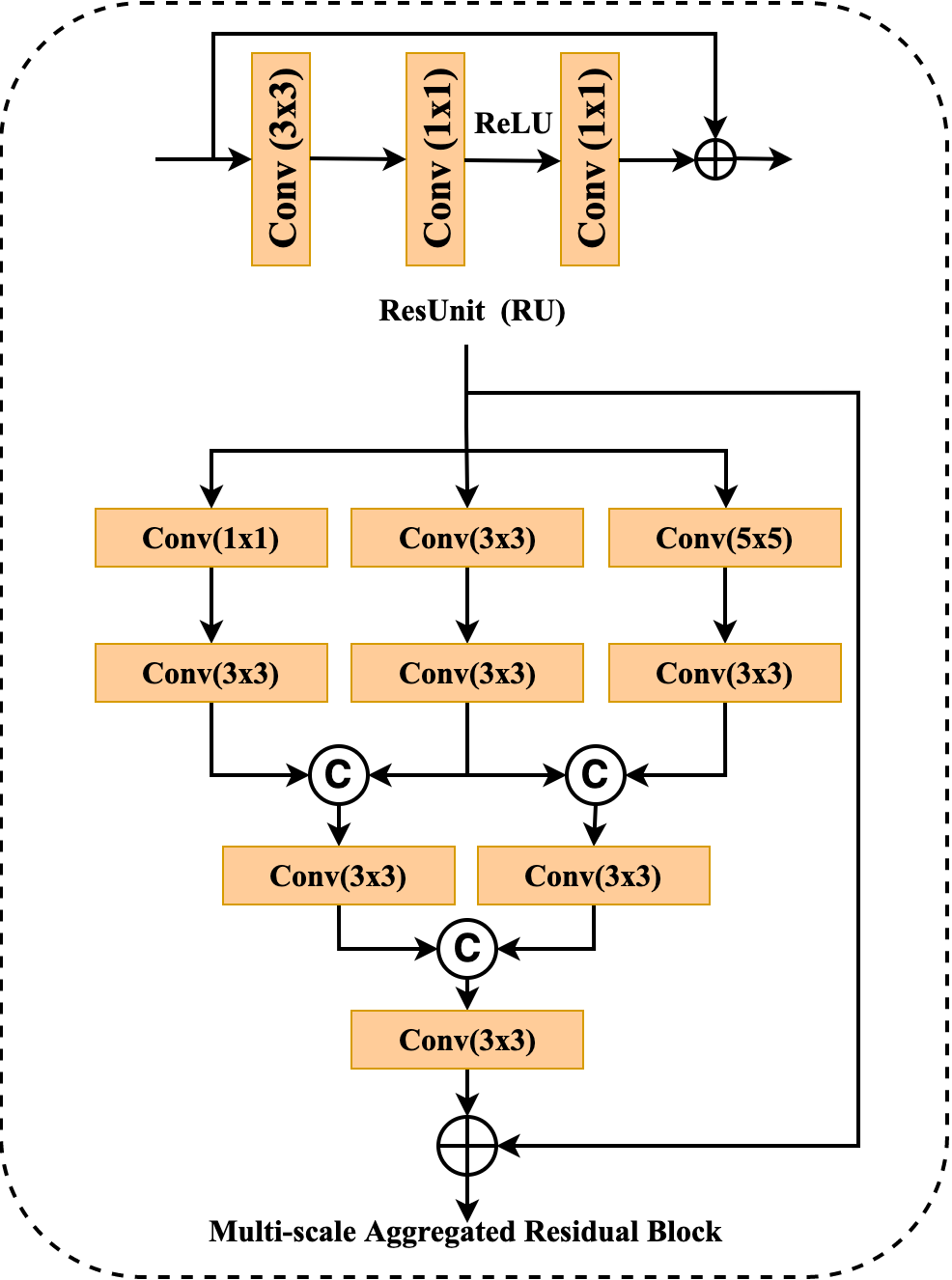}
\end{center}
\caption{The complete architecture of the proposed ResUnit (RU) and Multi-scale Aggregated Residual Block (MARB).}
\label{block}
\end{figure}

As shown in Fig.~\ref{block}, we also design a ResUnit (RU) for high-dimensional features encoding. Since snow has different effects in different regions, RU encodes these differences into features, making the model more adaptive. Meanwhile, we adopted a multi-level residual method to fuse the residual information of the shallow layer and deep layer to further improve the snow removal effect. Therefore, the adaptive residual block in GF-Net has two levels, and the residual of the $i$-th level can be expressed as
\begin{equation}
    f_{res}^{i}  = f_{snowy}^{i} - f_{mask}^{i},
\end{equation}
where $f_{mask}^{i}$ and $f_{snowy}^{i}$ represent the feature encoding of the snow mask and the snowy image at the $i$-th level, respectively.

The specific process of GF-Net is intuitively reflected in Fig.~\ref{gfn}. Firstly, it performs the first-level adaptive coding on the input mask feature $\mathcal{F}_{mask}$ and the snowy image $\mathcal{I}_{snowy}$ to obtain $f_{mask}^{1}$ and $f_{snowy}^{1}$
\begin{equation}
    f_{snow}^{1}=RU^{11}(Conv^{1}(\mathcal{I}_{snow})),
\end{equation}
\begin{equation}
    f_{mask}^{1}=RU^{12}(Conv^{2}(\mathcal{F}_{mask})),
\end{equation}
where $RU^{ij}(\cdot)$ represents the $j$-th ResUnit of the $i$-th level. After that, the residuals of these two features are used to obtain the first-level residual $f_{res}^{1}$. Secondly, the network performs second-level adaptive coding on $f_{mask}^{1}$ and $f_{snow}^{1}$ to obtain $f_{mask}^{2}$ and $f_{snow}^{2}$
\begin{equation}
    f_{snow}^{2}=RU^{21}(f_{snow}^{1}),
\end{equation}
\begin{equation}
    f_{mask}^{2}=RU^{22}(f_{mask}^{1}).
\end{equation}
Similarly, the residuals of $f_{mask}^{2}$ and $f_{snow}^{2}$ are used to obtain the second-level residual $f_{res}^{2}$. Finally, we encode and fuse $f_{res}^{1}$ and $f_{res}^{2}$ through two convolutional layers to obtain the final feature $\mathcal{F}_{fuse}$ in a relatively snow-free state.

\subsection{Reconstruct-Net}
After GF-Net, we can obtained features $\mathcal{F}_{fuse}$ that are close to snow-free. However, in addition to snow phenomena such as snowflakes and snow streaks, the veiling effect often appears in snow images. This phenomenon is similar to haze in that its effect on an object depends on its distance. Therefore, image snow removal must also include removal of the veiling effect. Not only that, there are also some imperfections and residual snow after GF-Net. To solve these problems, we also propose a Reconstruct-Net to correct these areas to ensure the quality of the reconstructed images.

As shown in Fig.~\ref{model}, Reconstruct-Net consists of three Multi-scale Aggregated Residual Blocks (MARBs) and a convolutional layer. Among them, MARB (Fig.~\ref{block}) is specially designed to capture the multi-scale features with different receptive fields, which is inspired by~\cite{szegedy2015going, li2018multi, li2020mdcn}. The haze effect caused by the veiling effect and the residual snow removal from the previous part of the network is unevenly distributed in the image. Using the multi-scale receptive field of MARB can well remove and repair the pollution in these areas. Specifically, MARB consists of $1\times1$, $3\times3$, and $5\times5$ parallel convolutional layers, and all of them will pass through a $3\times3$ convolutional layer respectively. Then, the branch of the $3\times3$ kernel size will be connected with the branches of the $1\times1$ and $5\times5$ kernel size to form two new branches. After the $3\times3$ convolutional layer is performed on these two new branches, they will also be connected as a new branch, and a $3\times3$ convolutional layer will be performed again. Finally, the input of MARB is summed with output features to enhance the information flow. In Reconstruct-Net, a global residual connection is performed after the three MARBs are stacked to enhance feature representation. The operation of Reconstruct-Net can be defined as
\begin{equation}
    \mathcal{I}_{clear}^{'} = \mathcal{R}(\mathcal{F}_{fuse}),
\end{equation}
where $\mathcal{R}(\cdot)$ denotes the Reconstruct-Net and $\mathcal{I}_{clear}^{'}$ is the reconstructed snow-free image.

\subsection{Loss Function}
During training, we use $L1$ loss as the reconstruct loss to minimize the difference between the reconstructed snow-free image and the corresponding clear image $\mathcal{I}_{clear}$
\begin{equation}
    \mathcal{L}_{reconstruct} = \left \| \mathcal{I}_{clear}^{'} -  \mathcal{I}_{clear} \right \|_{1}.
\end{equation}

In summary, Mask-Net, GF-Net, and Reconstruct-Ne form the complete Snow Mask Guided Adaptive Residual Network (SMGARN). In this work, Mask-Net served as a sub-network to provide the snow mask and SMGARN adopts an end-to-end training mode. Therefore, the reconstruct loss $\mathcal{L}_{reconstruct}$ and mask loss $\mathcal{L}_{mask}$ form the complete mask-assisted loss $\mathcal{L}_{mas}$
\begin{equation}
    \mathcal{L}_{mas} = \mathcal{L}_{reconstruct}+\lambda \mathcal{L}_{mask},
\end{equation}
where $\lambda$ is a hyper-parameter used to control the composition of the mask loss. To ensure that $\mathcal{L}_{reconstruct}$ and $\mathcal{L}_{mask}$ receive equal attention, we set $\lambda$ to $1$ in this work.

\section{Experiments}~\label{EX}
In this part, we provide relevant experimental details, descriptions, and results to verify the effectiveness and excellence of the proposed SMGARN.

\subsection{Datasets}
In this work, we trained our SMGARN with the CSD~\cite{chen2021all} train set (8000 images) for synthetic images and use the Snow100K~\cite{liu2018desnownet} train set (50000 images) to train the model for real images. For evaluation, we use three benchmark test sets, including SRRS~\cite{chen2020jstasr}, CSD~\cite{chen2021all}, and Snow100K~\cite{liu2018desnownet}. Specifically, we adopted the last 2000 images in the SRRS test set for evaluation. For CSD, we use its 2000 test images for verification. In addition, we use the three test sets Snow100K-S, Snow100K-M and Snow100K-L provided by Snow100K for evaluation. 

\begin{figure*}[t]
\begin{center}
\includegraphics[width=0.9\linewidth]{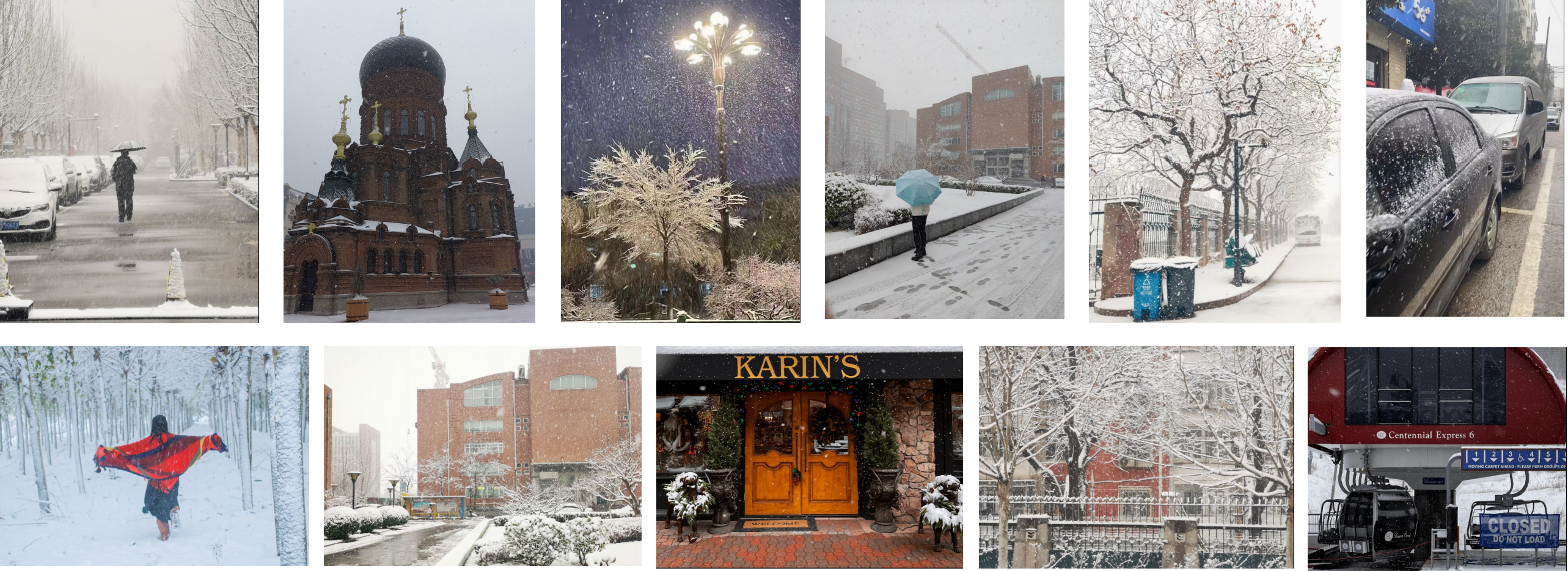}
\end{center}
\caption{Some sample images from the proposed SnowWorld24 dataset. All of these images were taken with different cameras on different snowy days.} 
\label{Dataset}
\end{figure*}

The training and testing datasets of the existing datasets (SRRS, CSD, and Snow100K) are all generated by synthetic techniques, so there are many abnormal images, such as snowflakes in summer or sunny days, and the effect of directly using them for evaluation is limited. Although Snow100K has proposed a real snow scene dataset, most of the images are extracted directly from web pages, and many images are not clear and have watermarks. More importantly, there are also a large number of post-processed artistic photos in this dataset. Since there are no labels and snow masks in the real scene, images need to be more clear and able to represent the complex situation of the real world. Therefore, Only the data provided by Snow100K cannot effectively evaluate the performance of the model. In order to further evaluate the performance and generalization of the snow removal model more comprehensively, we construct a new real snow image dataset, named \textbf{SnowWorld24}. 

\textbf{SnowWorld24}. As shown in Fig.~\ref{Dataset}, SnowWorld24 contains 24 snow images of different scenes, which are directly obtained under snow weather without any processing, and there are no traces such as watermarks that are not related to the images themselves. Meanwhile, all of these images were taken with different cameras on different snowy days, and all of these scenes are from different countries and regions. The purpose of this is to increase the diversity of the snow images, so as to better verify the generalization of the model. In summary, SnowWorld24 is a real dataset specially proposed to verify model validity and generalization.

\subsection{Implementation Detail}
\subsubsection{Training Details}
Following previous works, we randomly extract $16$ snowy patches with the size of $128\times128$ as inputs. Meanwhile, we augment the training data with flips and rotates operations to further improve the generalization ability of the model. The learning rate is initialized to $10^{-4}$ and halved every 100 epochs. We implement our model with the PyTorch framework and update it with the Adam optimizer. 

\begin{table}[t]
	\centering
	\setlength{\tabcolsep}{3mm}
    \caption{Quantitative comparison with other advanced methods on Snow100K-S, Snow100K-M, and Snow100K-L. Among them, best results are \textbf{highlight} and the second best results are \textit{ITALIC}.}
		\begin{tabular}{c|ccccc}
			\toprule
			Method / Dataset &Snow100K-S	        & Snow100K-M         & Snow100K-L        \\  \midrule 
			Zheng\cite{zheng2013single}                     & 20.29 / 0.73      & 20.18 / 0.71      & 18.83 / 0.66      \\ 
			DehazeNet~\cite{cai2016dehazenet}               & 22.06 / 0.78      & 21.54 / 0.74      & 20.19 / 0.70      \\ 
			HDCWNet~\cite{cai2016dehazenet}                 & 22.13 / 0.78      & 21.66 / 0.72      & 19.91 / 0.69      \\
			DerainNet~\cite{fu2017clearing}                 & 25.74 / 0.86      & 23.36 / 0.84      & 19.18 / 0.74      \\ 
			DeepLab~\cite{chen2017deeplab}                  & 25.94 / 0.87      & 24.36 / 0.85      & 21.29 / 0.77      \\ 
			JORDER~\cite{yang2017deep}                      & 25.62 / 0.88      & 24.97 / 0.87      & 23.40 / 0.80      \\ 
			DuRN-S-P~\cite{liu2019dual}                     & 32.27 / 0.94      & 30.92 / 0.93      & 27.21 / 0.88      \\ 
			DeSnowNet~\cite{liu2018desnownet}               & 32.23 / \textit{0.95}      & 30.86 / \textit{0.94}      & 27.16 / 0.89      \\
			DS-GAN~\cite{jaw2020desnowgan}                  & \textit{33.43} / \textbf{0.96}      & \textit{31.87} / \textbf{0.95}      & 28.06 / \textit{0.92}      \\
			All in One~\cite{li2020all}                     & - / -             & - / -             & \textit{28.33} / 0.88      \\ \midrule
			SMGARN(Ours)                                    & \textbf{34.46} / \textit{0.95}      & \textbf{33.22} / \textbf{0.95}      & \textbf{29.44} / \textbf{0.93}       \\
			\bottomrule 
		\end{tabular}
		\label{sota1}
\end{table}

\subsubsection{Model Details}
In our final model, we use three MARBs in the Reconstruct-Net. Meanwhile, we set the embedding dimension of the model to $112$ and the kernel size of the convolutional layers in Fig.~\ref{model} are set to $3 \times 3$. In addition, between the second and third convolutional layers of ResUnit, we expand the feature embedding dimension by 2 times, and the channel of the final output will be reduced to the original dimension.

\subsection{Comparisons with State-of-the-Art Methods}
In this section, we compare our proposed SMGARN with other advanced methods to verify the validity of the proposed model. Since image snow removal is an emerging field, the test data and test method of some method are confusing. In this work, we  unify multiple classic snow removal models, and all of which are trained on the same dataset and conditions. It is worth noting that we re-produced or re-tested some models whose experimental details were unclear. Besides, due to the commonality of image restoration and atmospheric phenomena, we also provide the results of classic models for image restoration (SRCNN~\cite{dong2015image}), image dehazing (DehazeNet~\cite{cai2016dehazenet}), and deraining (RESCAN~\cite{li2018recurrent}). All these methods are trained under the same experimental setting and training set. For the validation method, we adopt PSNR and SSIM to evaluate the final results.

\begin{figure*}[t]
\begin{minipage}[c]{0.135\textwidth}
\includegraphics[width=2.45cm,height=2cm]{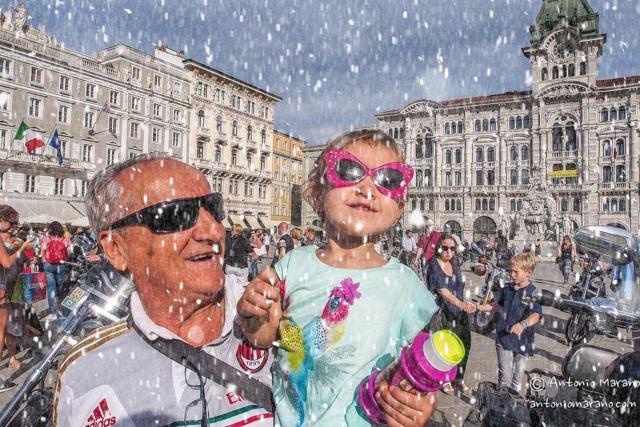}
\end{minipage}
\begin{minipage}[c]{0.135\textwidth}
\includegraphics[width=2.45cm,height=2cm]{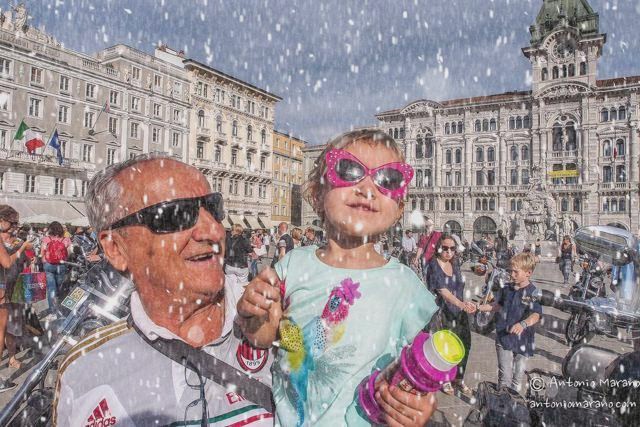}
\end{minipage}
\begin{minipage}[c]{0.135\textwidth}
\includegraphics[width=2.45cm,height=2cm]{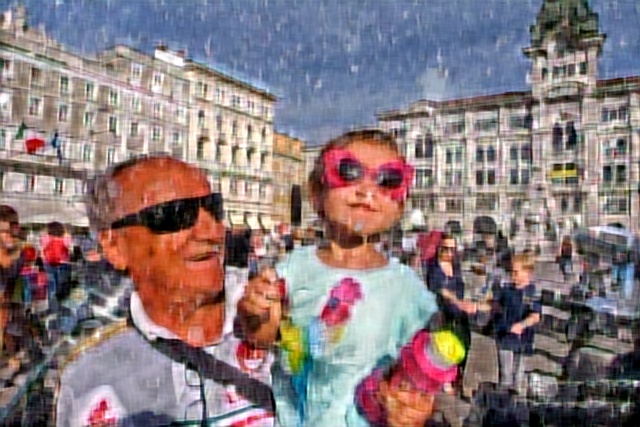}
\end{minipage}
\begin{minipage}[c]{0.135\textwidth}
\includegraphics[width=2.45cm,height=2cm]{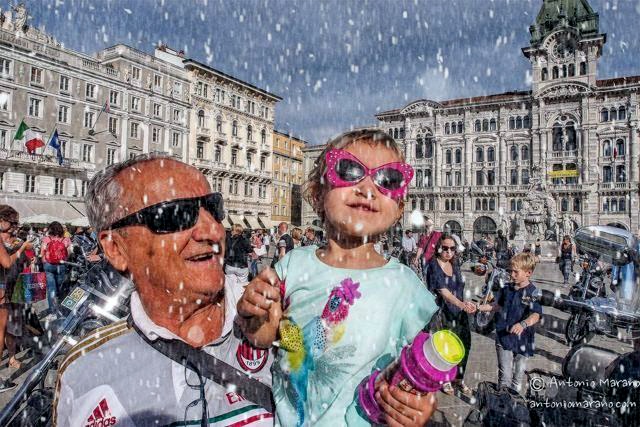}
\end{minipage}
\begin{minipage}[c]{0.135\textwidth}
\includegraphics[width=2.45cm,height=2cm]{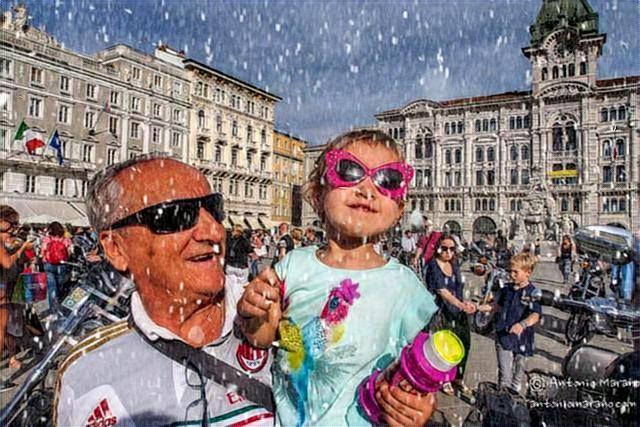}
\end{minipage}
\begin{minipage}[c]{0.135\textwidth}
\includegraphics[width=2.45cm,height=2cm]{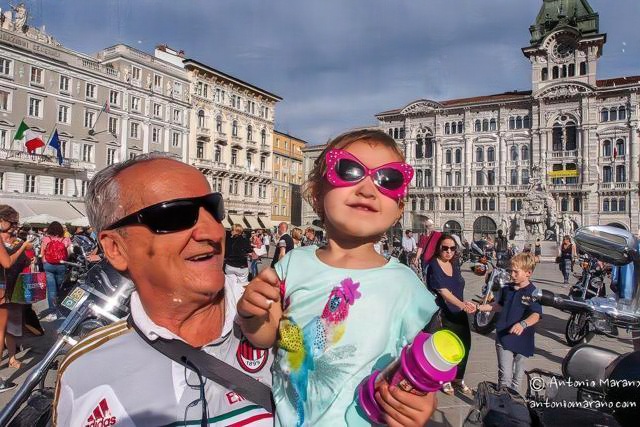}
\end{minipage}
\begin{minipage}[c]{0.135\textwidth}
\includegraphics[width=2.45cm,height=2cm]{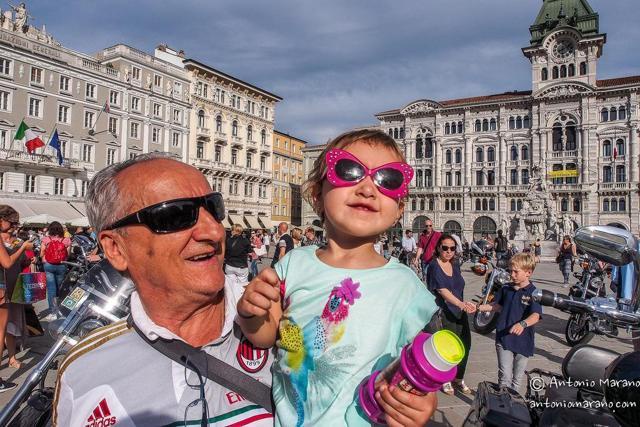}
\end{minipage}

\begin{minipage}{1\textwidth}
\end{minipage}

\begin{minipage}[c]{0.135\textwidth}
\includegraphics[width=2.45cm,height=2cm]{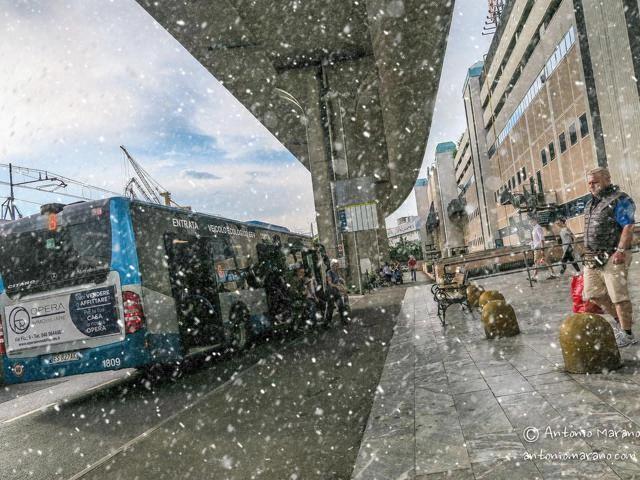}
\centerline{Snowy Image}
\centerline{}
\end{minipage}
\begin{minipage}[c]{0.135\textwidth}
\includegraphics[width=2.45cm,height=2cm]{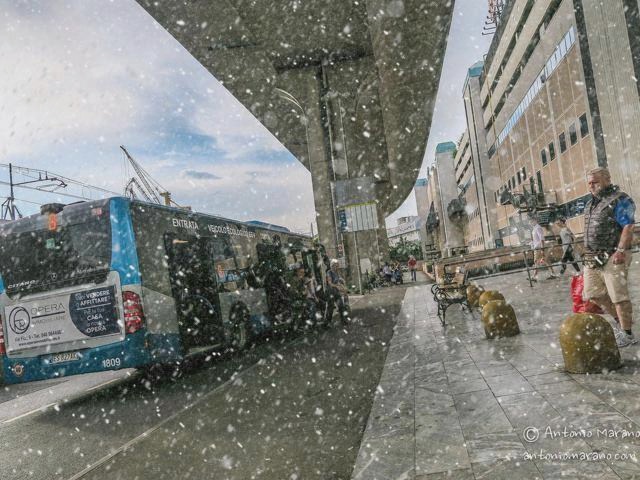}
\centerline{Zheng~\cite{zheng2013single}}
\centerline{}
\end{minipage}
\begin{minipage}[c]{0.135\textwidth}
\includegraphics[width=2.45cm,height=2cm]{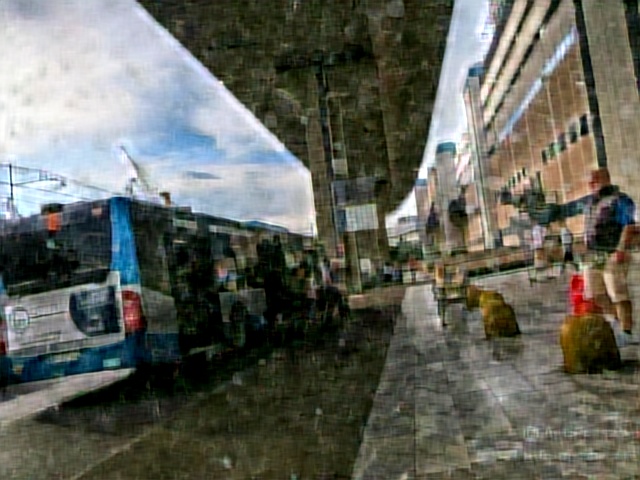}
\centerline{DehazeNet~\cite{cai2016dehazenet}}
\centerline{}
\end{minipage}
\begin{minipage}[c]{0.135\textwidth}
\includegraphics[width=2.45cm,height=2cm]{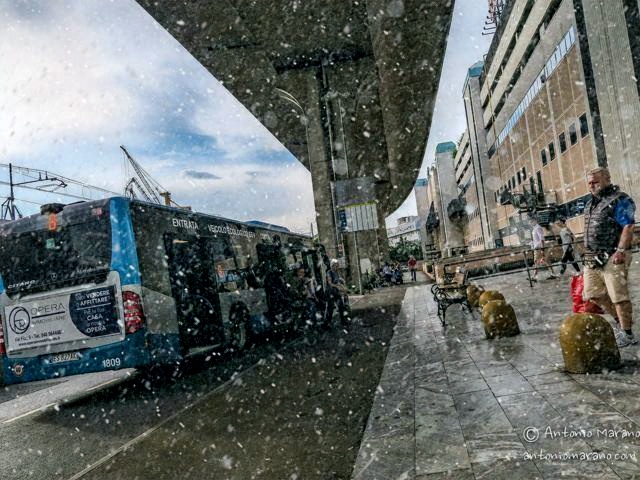}
\centerline{RESCAN~\cite{li2018recurrent}}
\centerline{}
\end{minipage}
\begin{minipage}[c]{0.135\textwidth}
\includegraphics[width=2.45cm,height=2cm]{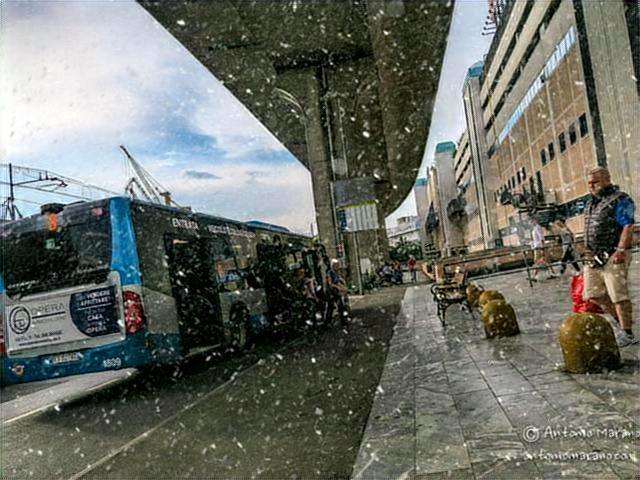}
\centerline{HDCWNet~\cite{chen2021all}}
\centerline{}
\end{minipage}
\begin{minipage}[c]{0.135\textwidth}
\includegraphics[width=2.45cm,height=2cm]{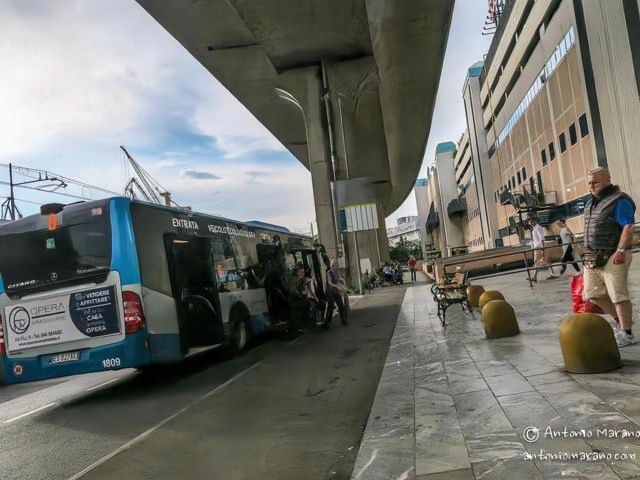}
\centerline{\textbf{SMGARN(Ours)}}
\centerline{}
\end{minipage}
\begin{minipage}[c]{0.13\textwidth}
\includegraphics[width=2.45cm,height=2cm]{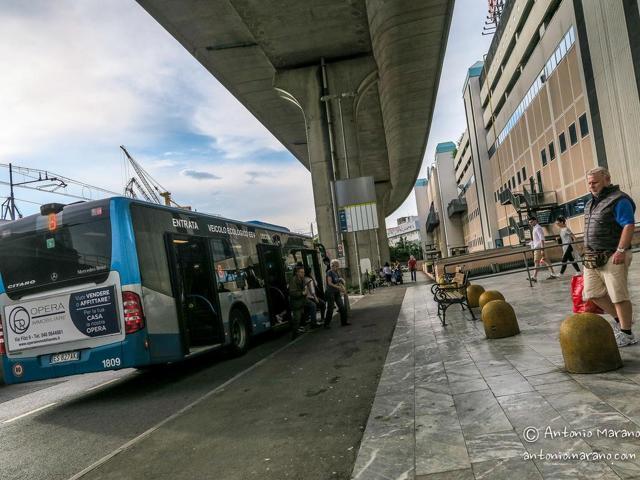}
\centerline{GT}
\centerline{}
\end{minipage}
\caption{Visual comparison with other advanced models on Snow100K. Obviously, our proposed method can reconstruct high-quality snow-free images.}
\label{snow100k}
\end{figure*}

\begin{figure*}[t]
\begin{minipage}[c]{0.16\textwidth}
\includegraphics[width=2.9cm,height=1.8cm]{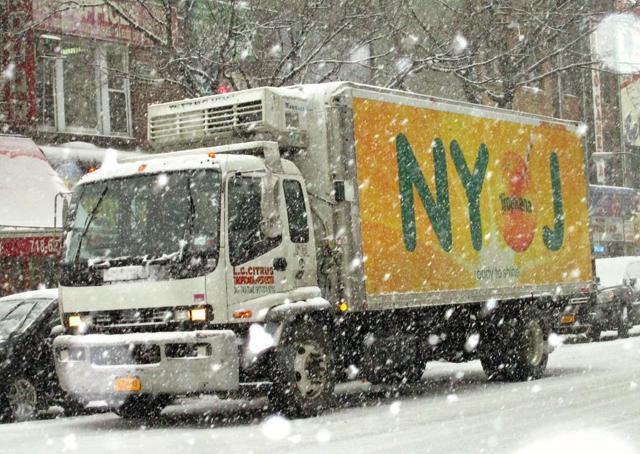}
\end{minipage}
\begin{minipage}[c]{0.16\textwidth}
\includegraphics[width=2.9cm,height=1.8cm]{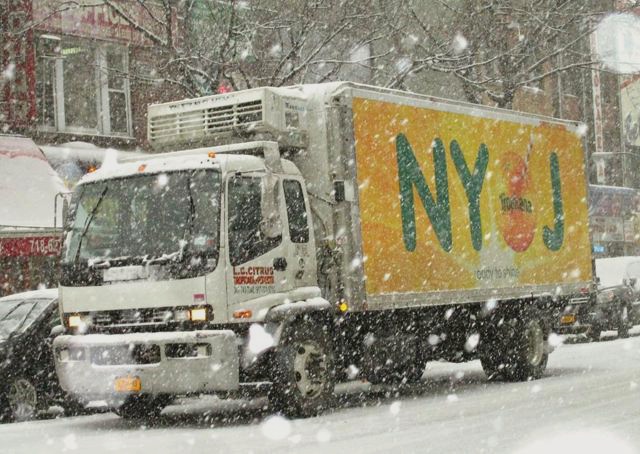}
\end{minipage}
\begin{minipage}[c]{0.16\textwidth}
\includegraphics[width=2.9cm,height=1.8cm]{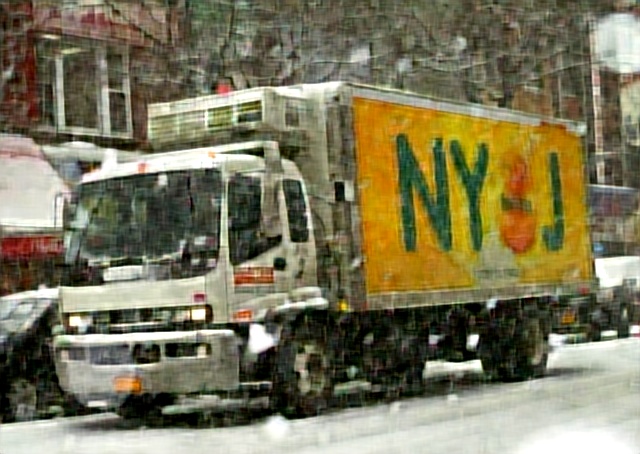}
\end{minipage}
\begin{minipage}[c]{0.16\textwidth}
\includegraphics[width=2.9cm,height=1.8cm]{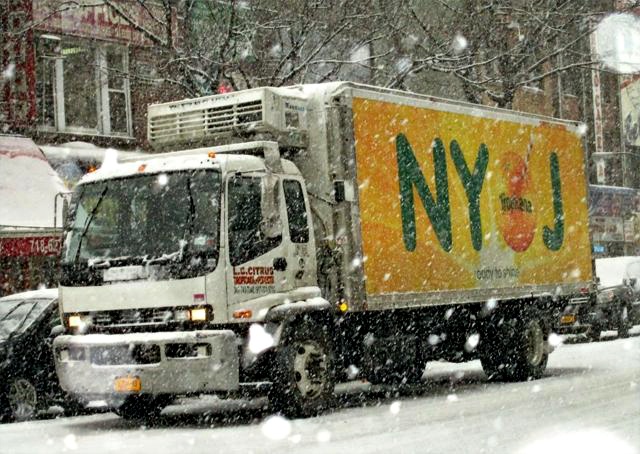}
\end{minipage}
\begin{minipage}[c]{0.16\textwidth}
\includegraphics[width=2.9cm,height=1.8cm]{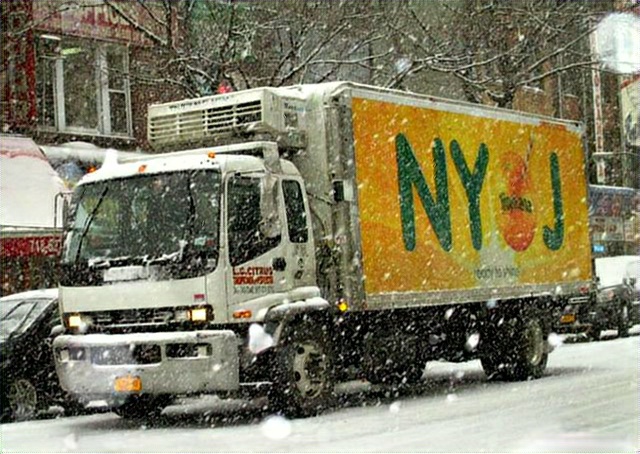}
\end{minipage}
\begin{minipage}[c]{0.16\textwidth}
\includegraphics[width=2.9cm,height=1.8cm]{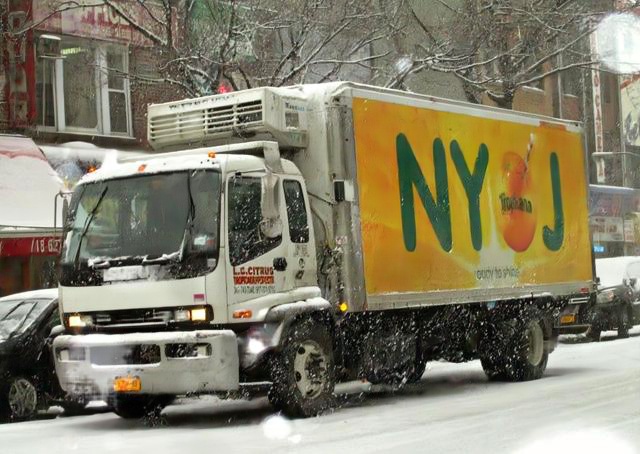}
\end{minipage}

\begin{minipage}[c]{1\textwidth}
\end{minipage}

\begin{minipage}[c]{0.16\textwidth}
\includegraphics[width=2.9cm,height=1.8cm]{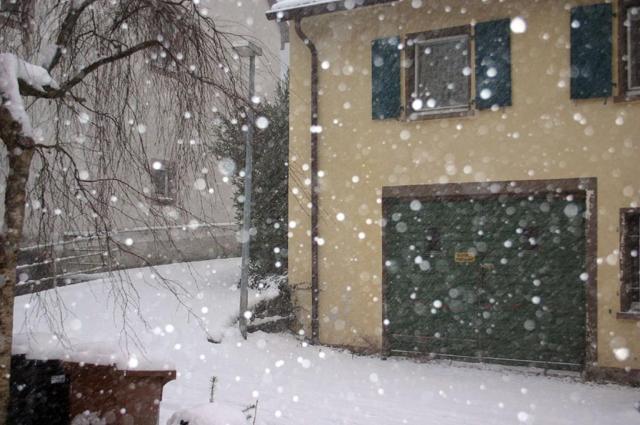}
\end{minipage}
\begin{minipage}[c]{0.16\textwidth}
\includegraphics[width=2.9cm,height=1.8cm]{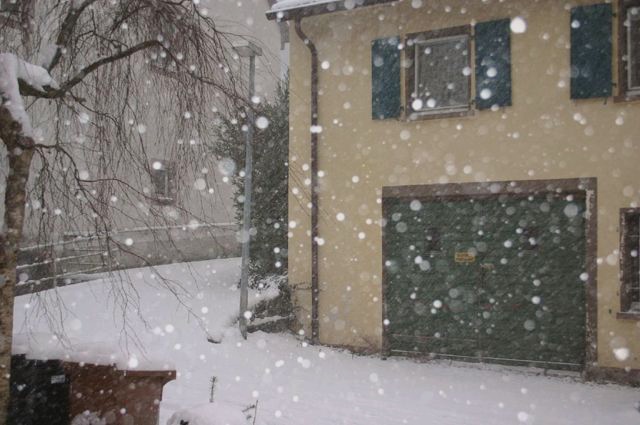}
\end{minipage}
\begin{minipage}[c]{0.16\textwidth}
\includegraphics[width=2.9cm,height=1.8cm]{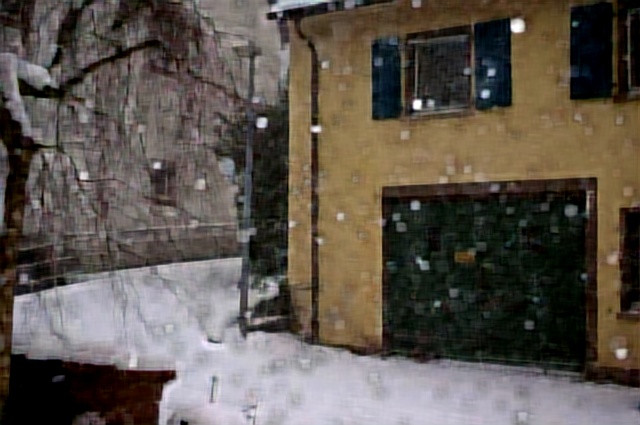}
\end{minipage}
\begin{minipage}[c]{0.16\textwidth}
\includegraphics[width=2.9cm,height=1.8cm]{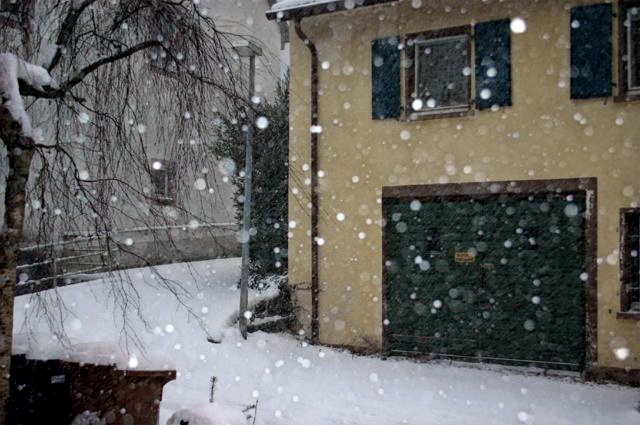}
\end{minipage}
\begin{minipage}[c]{0.16\textwidth}
\includegraphics[width=2.9cm,height=1.8cm]{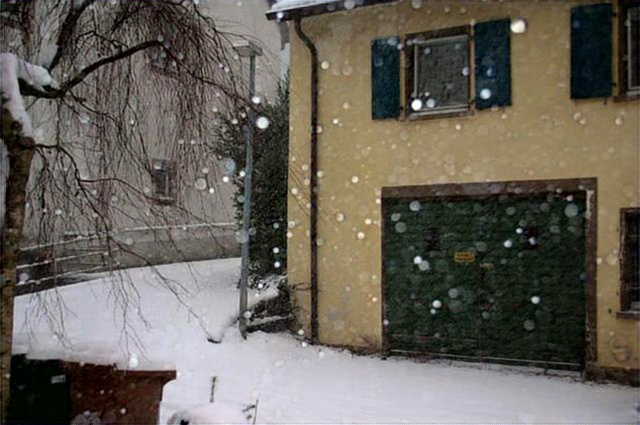}
\end{minipage}
\begin{minipage}[c]{0.16\textwidth}
\includegraphics[width=2.9cm,height=1.8cm]{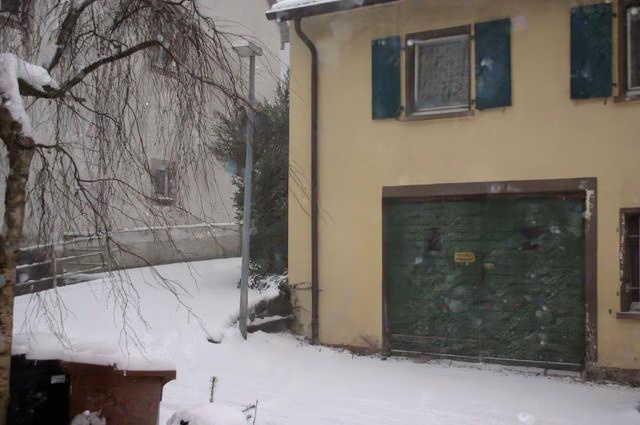}
\end{minipage}

\begin{minipage}[c]{1\textwidth}
\end{minipage}

\begin{minipage}[c]{0.16\textwidth}
\includegraphics[width=2.9cm,height=1.8cm]{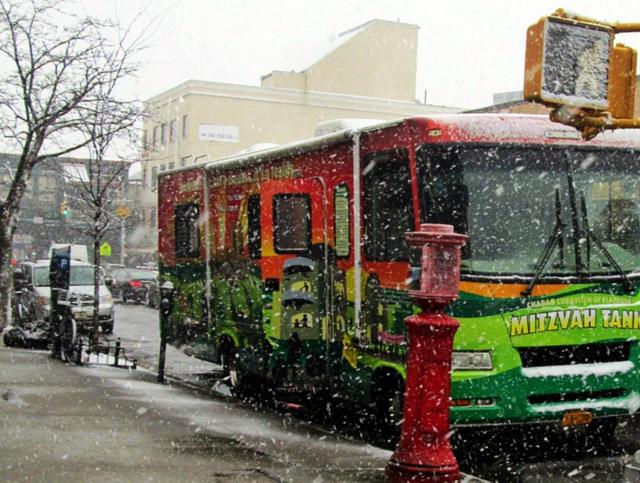}
\centerline{Snowy Image}
\centerline{}
\end{minipage}
\begin{minipage}[c]{0.16\textwidth}
\includegraphics[width=2.9cm,height=1.8cm]{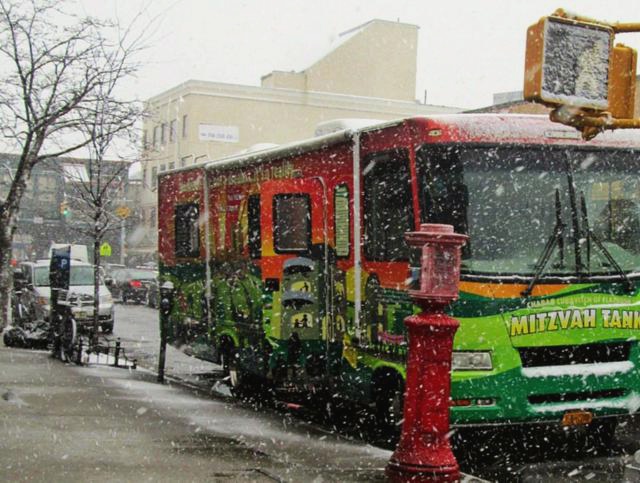}
\centerline{Zheng~\cite{zheng2013single}}
\centerline{}
\end{minipage}
\begin{minipage}[c]{0.16\textwidth}
\includegraphics[width=2.9cm,height=1.8cm]{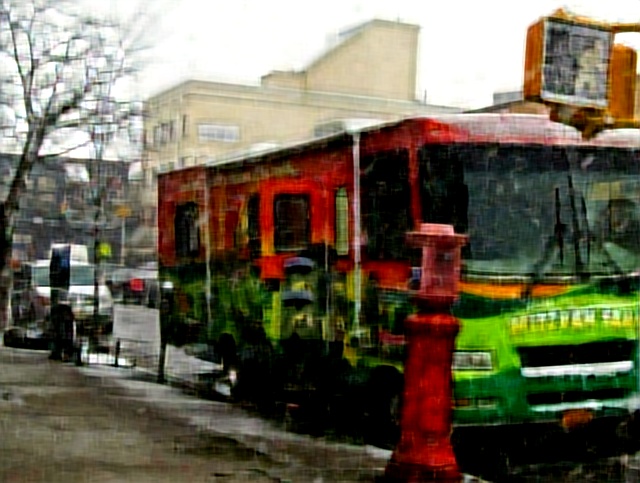}
\centerline{DehazeNet~\cite{cai2016dehazenet}}
\centerline{}
\end{minipage}
\begin{minipage}[c]{0.16\textwidth}
\includegraphics[width=2.9cm,height=1.8cm]{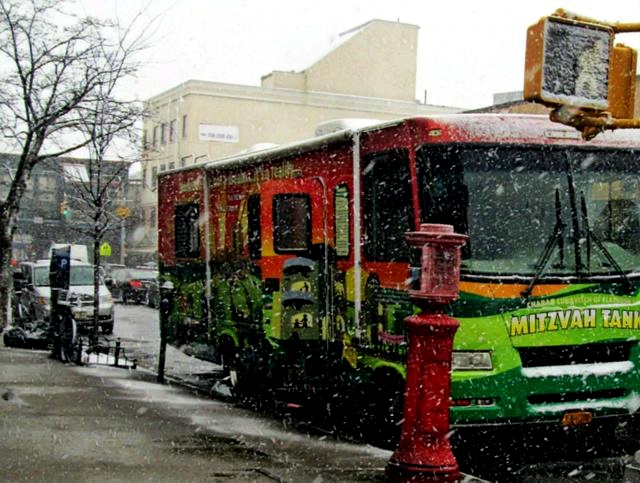}
\centerline{RESCAN~\cite{li2018recurrent}}
\centerline{}
\end{minipage}
\begin{minipage}[c]{0.16\textwidth}
\includegraphics[width=2.9cm,height=1.8cm]{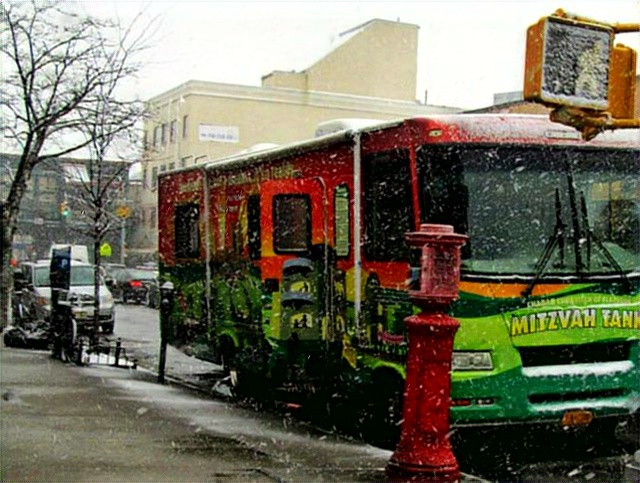}
\centerline{HDCWNet~\cite{chen2021all}}
\centerline{}
\end{minipage}
\begin{minipage}[c]{0.16\textwidth}
\includegraphics[width=2.9cm,height=1.8cm]{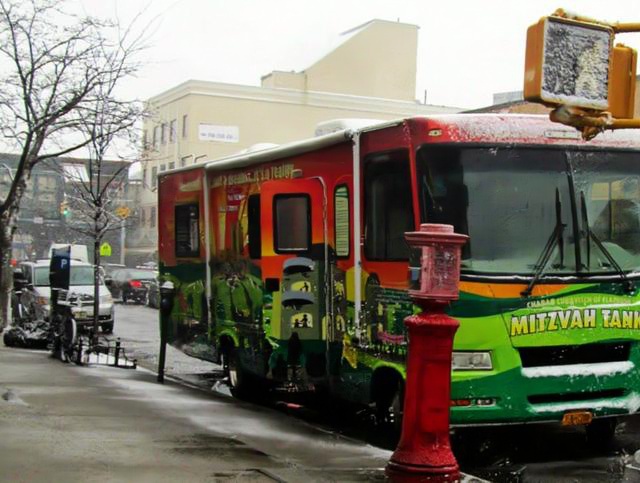}
\centerline{\textbf{SMGARN(Ours)}}
\centerline{}
\end{minipage}
\caption{Visual comparison on real snow images from Snow100K. Obviously, our proposed method can reconstruct clear images with less snow residue.}
\label{snow100k_real}
\end{figure*}

\begin{table}[t]
\centering
	\setlength{\tabcolsep}{6mm}
    \caption{Quantitative comparison with other advanced methods on CSD and SRRS. Among them, best results are \textbf{highlight} and the second best results are \textit{ITALIC}.}
		\begin{tabular}{c|cc}
			\toprule
			Method / Dataset                                 & CSD~\cite{chen2021all}                   &SRRS~\cite{chen2020jstasr} \\ \midrule
			Zheng~\cite{zheng2013single}            &14.21/0.61             &16.34/0.69 \\
			SRCNN~\cite{dong2015image}              &22.25/0.82             &22.46/0.85 \\
			DehazeNet~\cite{cai2016dehazenet}       &20.91/0.72             &21.22/0.75 \\
			RESCAN~\cite{li2018recurrent}           &22.11/0.81             &22.79/0.86 \\
			DeSnowNet~\cite{liu2018desnownet}       &20.13/0.81             &20.38/0.84 \\
			CycleGAN~\cite{engin2018cycle}          &20.98/0.80             &20.21/0.74 \\
			DAD~\cite{zou2020deep}                  &24.33/0.85             &24.31/0.86 \\
			All in One~\cite{li2020all}             &26.31/0.87             &24.98/0.88 \\
			JSTASR~\cite{chen2020jstasr}            &27.96/0.88             &- \\
			HDCWNet~\cite{chen2021all}              & \textit{28.62/0.89}             & \textit{25.03/0.89} \\
			\midrule
			SMGARN (Ours)                           &\textbf{29.94/0.94}    &\textbf{25.43/0.92} \\ 
			\bottomrule 
		\end{tabular}
		\label{sota2}
\end{table}

\subsubsection{Quantitative Comparison}
In Tables~\ref{sota1} and~\ref{sota2}, we show the quantitative comparison between our method and other advanced methods. Among them, TABLE~\ref{sota1} shows the results of these methods on three subsets of the Snow100K test set (Snow100K-S, Snow100K-M, Snow100K-L). According to the table, we can observe that the PSNR performance of our method are significantly outperforms previous methods. Compared with DS-GAN~\cite{jaw2020desnowgan}, SMGARN improves PSNR by \textbf{1.03dB} and \textbf{1.35dB} in Snow100K-S and Snow100K-M, respectively. Notably, on the hardest Snow100K-L, our method achieves a PSNR improvement of \textbf{1.11dB} over the previous best model (All in One~\cite{li2020all}). It can be seen that our model improves the PSNR results more than \textbf{1dB} compared to the current SOTA model on these three datasets. This fully demonstrated that the proposed SMGARN can significantly improve the snow removal performance.

Since Snow100K does not take into account the influence of the veiling effect, we tested it on the other snow removal datasets, CSD and SRRS, to further evaluate the proposed method. In TABLE~\ref{sota2}, we provide PSNR and SSIM results of these methods. It is clear that SMGARN outperforms all previous methods. Especially on the CSD test set, our proposed method improves the PSNR performance by \textbf{1.32dB}, which is a huge improvement. All these results further verify the effectiveness of the proposed SMGARN.

\begin{figure*}[h]
\begin{minipage}[c]{0.12\textwidth}
\includegraphics[width=2.2cm,height=1.8cm]{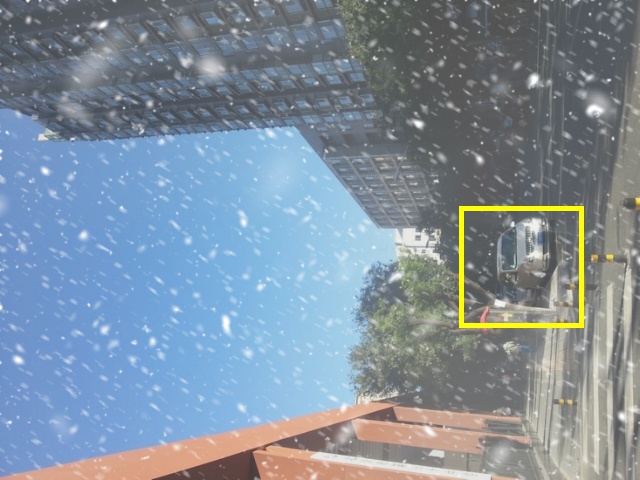}
\centerline{CSD-8066}
\end{minipage}
\begin{minipage}[c]{0.12\textwidth}
\includegraphics[width=2.2cm,height=1.8cm]{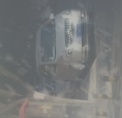}
\centerline{12.27dB}
\end{minipage}
\begin{minipage}[c]{0.12\textwidth}
\includegraphics[width=2.2cm,height=1.8cm]{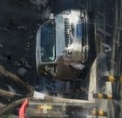}
\centerline{22.22dB}
\end{minipage}
\begin{minipage}[c]{0.12\textwidth}
\includegraphics[width=2.2cm,height=1.8cm]{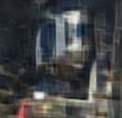}
\centerline{20.70dB}
\end{minipage}
\begin{minipage}[c]{0.12\textwidth}
\includegraphics[width=2.2cm,height=1.8cm]{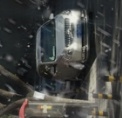}
\centerline{20.05dB}
\end{minipage}
\begin{minipage}[c]{0.12\textwidth}
\includegraphics[width=2.2cm,height=1.8cm]{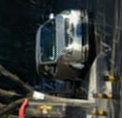}
\centerline{27.28dB}
\end{minipage}
\begin{minipage}[c]{0.12\textwidth}
\includegraphics[width=2.2cm,height=1.8cm]{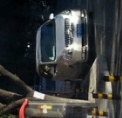}
\centerline{\textbf{29.08dB}}
\end{minipage}
\begin{minipage}[c]{0.12\textwidth}
\includegraphics[width=2.2cm,height=1.8cm]{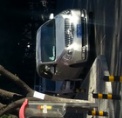}
\centerline{PSNR}
\end{minipage}

\begin{minipage}[c]{0.12\textwidth}
\includegraphics[width=2.2cm,height=1.8cm]{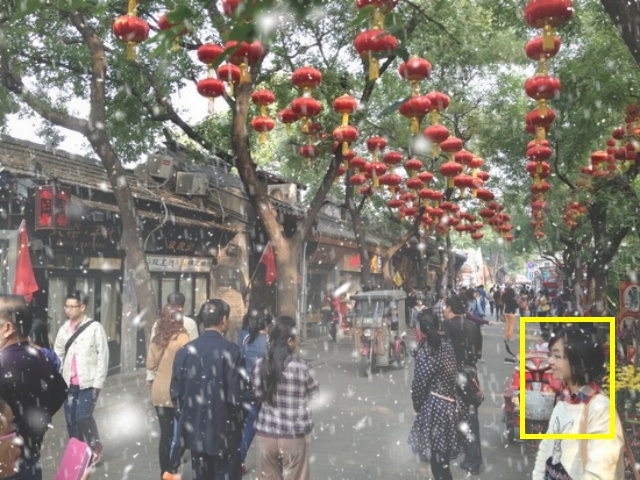}
\centerline{CSD-8132}
\end{minipage}
\begin{minipage}[c]{0.12\textwidth}
\includegraphics[width=2.2cm,height=1.8cm]{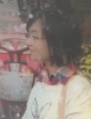}
\centerline{13.85dB}
\end{minipage}
\begin{minipage}[c]{0.12\textwidth}
\includegraphics[width=2.2cm,height=1.8cm]{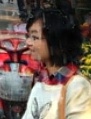}
\centerline{22.60dB}
\end{minipage}
\begin{minipage}[c]{0.12\textwidth}
\includegraphics[width=2.2cm,height=1.8cm]{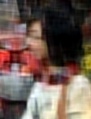}
\centerline{19.61dB}
\end{minipage}
\begin{minipage}[c]{0.12\textwidth}
\includegraphics[width=2.2cm,height=1.8cm]{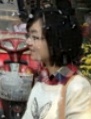}
\centerline{21.66dB}
\end{minipage}
\begin{minipage}[c]{0.12\textwidth}
\includegraphics[width=2.2cm,height=1.8cm]{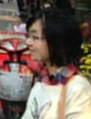}
\centerline{25.17dB}
\end{minipage}
\begin{minipage}[c]{0.12\textwidth}
\includegraphics[width=2.2cm,height=1.8cm]{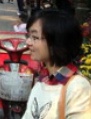}
\centerline{\textbf{28.79dB}}
\end{minipage}
\begin{minipage}[c]{0.12\textwidth}
\includegraphics[width=2.2cm,height=1.8cm]{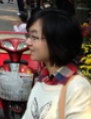}
\centerline{PSNR}
\end{minipage}

\begin{minipage}[c]{0.12\textwidth}
\includegraphics[width=2.2cm,height=1.8cm]{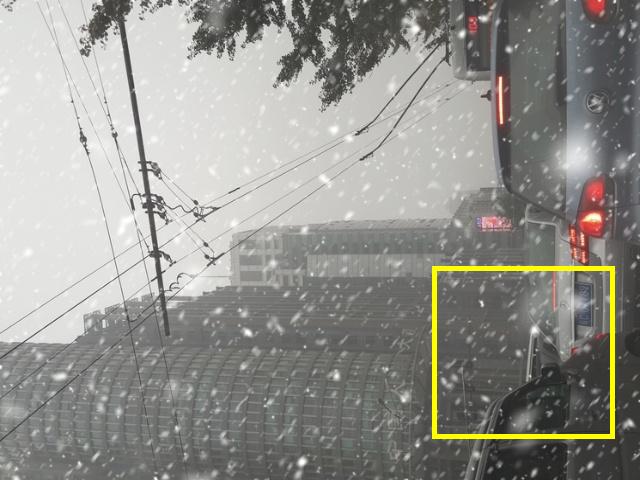}
\centerline{CSD-9983}
\end{minipage}
\begin{minipage}[c]{0.12\textwidth}
\includegraphics[width=2.2cm,height=1.8cm]{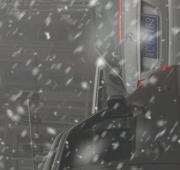}
\centerline{14.61dB}
\end{minipage}
\begin{minipage}[c]{0.12\textwidth}
\includegraphics[width=2.2cm,height=1.8cm]{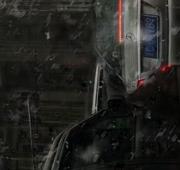}
\centerline{22.77dB}
\end{minipage}
\begin{minipage}[c]{0.12\textwidth}
\includegraphics[width=2.2cm,height=1.8cm]{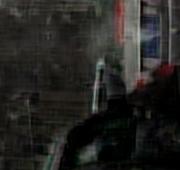}
\centerline{21.23dB}
\end{minipage}
\begin{minipage}[c]{0.12\textwidth}
\includegraphics[width=2.2cm,height=1.8cm]{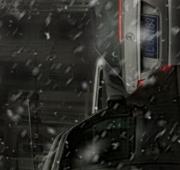}
\centerline{23.81dB}
\end{minipage}
\begin{minipage}[c]{0.12\textwidth}
\includegraphics[width=2.2cm,height=1.8cm]{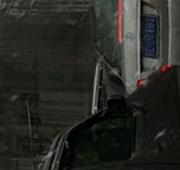}
\centerline{27.56dB}
\end{minipage}
\begin{minipage}[c]{0.12\textwidth}
\includegraphics[width=2.2cm,height=1.8cm]{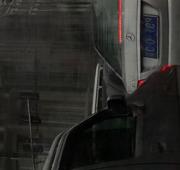}
\centerline{\textbf{31.65dB}}
\end{minipage}
\begin{minipage}[c]{0.12\textwidth}
\includegraphics[width=2.2cm,height=1.8cm]{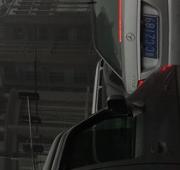}
\centerline{PSNR}
\end{minipage}

\begin{minipage}[c]{0.12\textwidth}
\includegraphics[width=2.2cm,height=1.8cm]{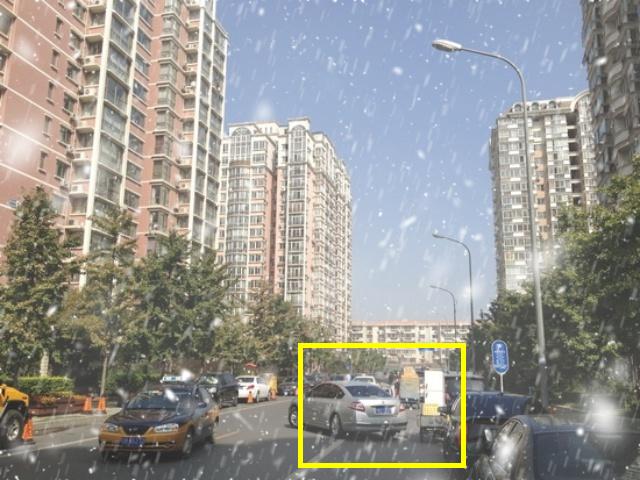}
\centerline{CSD-9993}
\end{minipage}
\begin{minipage}[c]{0.12\textwidth}
\includegraphics[width=2.2cm,height=1.8cm]{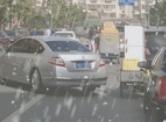}
\centerline{14.56dB}
\end{minipage}
\begin{minipage}[c]{0.12\textwidth}
\includegraphics[width=2.2cm,height=1.8cm]{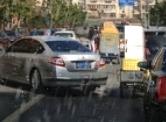}
\centerline{23.59dB}
\end{minipage}
\begin{minipage}[c]{0.12\textwidth}
\includegraphics[width=2.2cm,height=1.8cm]{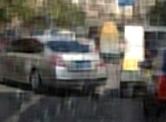}
\centerline{20.92dB}
\end{minipage}
\begin{minipage}[c]{0.12\textwidth}
\includegraphics[width=2.2cm,height=1.8cm]{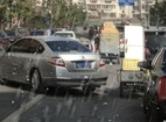}
\centerline{22.17dB}
\end{minipage}
\begin{minipage}[c]{0.12\textwidth}
\includegraphics[width=2.2cm,height=1.8cm]{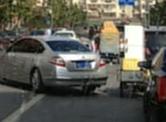}
\centerline{27.20dB}
\end{minipage}
\begin{minipage}[c]{0.12\textwidth}
\includegraphics[width=2.2cm,height=1.8cm]{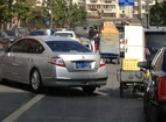}
\centerline{\textbf{30.47dB}}
\end{minipage}
\begin{minipage}[c]{0.12\textwidth}
\includegraphics[width=2.2cm,height=1.8cm]{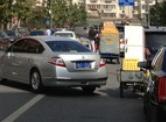}
\centerline{PSNR}
\end{minipage}

\begin{minipage}[c]{0.12\textwidth}
\includegraphics[width=2.2cm,height=1.8cm]{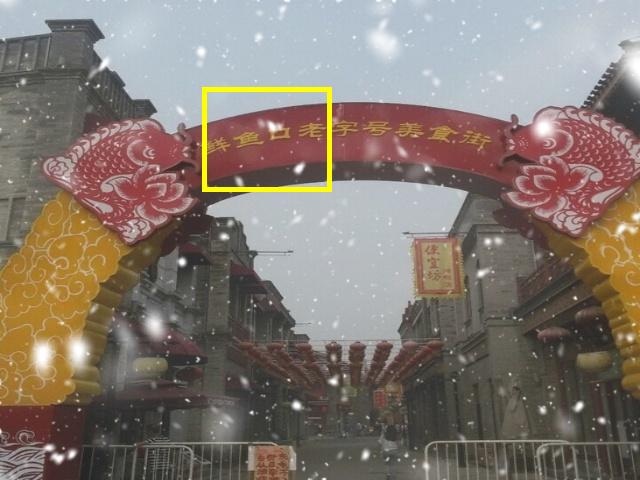}
\centerline{SRRS-14358}
\end{minipage}
\begin{minipage}[c]{0.12\textwidth}
\includegraphics[width=2.2cm,height=1.8cm]{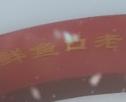}
\centerline{16.16dB}
\end{minipage}
\begin{minipage}[c]{0.12\textwidth}
\includegraphics[width=2.2cm,height=1.8cm]{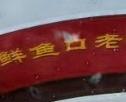}
\centerline{24.58dB}
\end{minipage}
\begin{minipage}[c]{0.12\textwidth}
\includegraphics[width=2.2cm,height=1.8cm]{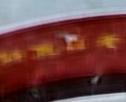}
\centerline{22.23dB}
\end{minipage}
\begin{minipage}[c]{0.12\textwidth}
\includegraphics[width=2.2cm,height=1.8cm]{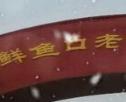}
\centerline{23.06dB}
\end{minipage}
\begin{minipage}[c]{0.12\textwidth}
\includegraphics[width=2.2cm,height=1.8cm]{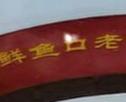}
\centerline{27.54dB}
\end{minipage}
\begin{minipage}[c]{0.12\textwidth}
\includegraphics[width=2.2cm,height=1.8cm]{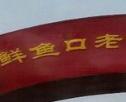}
\centerline{\textbf{30.39dB}}
\end{minipage}
\begin{minipage}[c]{0.12\textwidth}
\includegraphics[width=2.2cm,height=1.8cm]{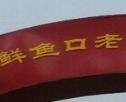}
\centerline{PSNR}
\end{minipage}

\begin{minipage}[c]{0.12\textwidth}
\includegraphics[width=2.2cm,height=1.8cm]{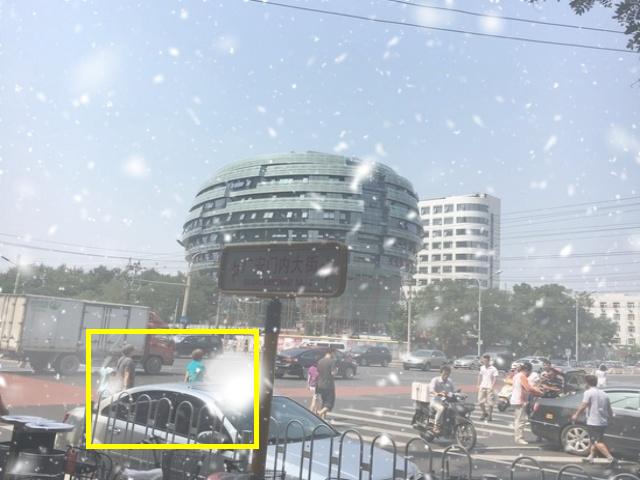}
\centerline{SRRS-14367}
\centerline{Snowy}
\centerline{}
\end{minipage}
\begin{minipage}[c]{0.12\textwidth}
\includegraphics[width=2.2cm,height=1.8cm]{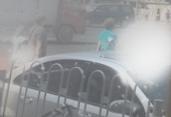}
\centerline{14.55dB}
\centerline{Zheng~\cite{zheng2013single}}
\centerline{}
\end{minipage}
\begin{minipage}[c]{0.12\textwidth}
\includegraphics[width=2.2cm,height=1.8cm]{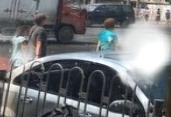}
\centerline{22.96dB}
\centerline{SRCNN~\cite{dong2015image}}
\centerline{}
\end{minipage}
\begin{minipage}[c]{0.12\textwidth}
\includegraphics[width=2.2cm,height=1.8cm]{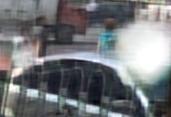}
\centerline{21.12dB}
\centerline{DehazeNet~\cite{cai2016dehazenet}}
\centerline{}
\end{minipage}
\begin{minipage}[c]{0.12\textwidth}
\includegraphics[width=2.2cm,height=1.8cm]{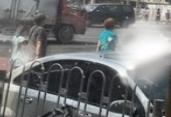}
\centerline{22.31dB}
\centerline{RESCAN~\cite{li2018recurrent}}
\centerline{}
\end{minipage}
\begin{minipage}[c]{0.12\textwidth}
\includegraphics[width=2.2cm,height=1.8cm]{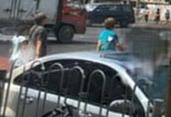}
\centerline{25.32dB}
\centerline{HDCWNet~\cite{chen2021all}}
\centerline{}
\end{minipage}
\begin{minipage}[c]{0.12\textwidth}
\includegraphics[width=2.2cm,height=1.8cm]{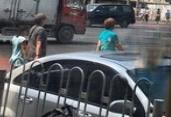}
\centerline{\textbf{26.92dB}}
\centerline{\textbf{Ours}}
\centerline{}
\end{minipage}
\begin{minipage}[c]{0.12\textwidth}
\includegraphics[width=2.2cm,height=1.8cm]{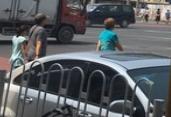}
\centerline{PSNR}
\centerline{GT}
\centerline{}
\end{minipage}
\caption{Visual comparison with other advanced models on CSD and SRRS. Obviously, our SMGARN can reconstruct high-quality snow-free images.}
\label{CSD&SRRS}
\end{figure*}

\subsubsection{Visual Comparison}
In Figs.~\ref{snow100k} and~\ref{CSD&SRRS}, we provide the visual comparison with other advance models on Snow100K~\cite{liu2018desnownet}, CSD~\cite{chen2021all}, and SRRS~\cite{chen2020jstasr}, respectively (please zoom in to see details). According to Fig.~\ref{snow100k}, we can clearly observed that our SMGARN can reconstruct high-quality snow-free image very close to GT. According to Fig.~\ref{CSD&SRRS}, we can found that the snow-free images reconstructed by our SMGARN are clearer and have less snow residue. Specifically, the first 4 rows of results in Fig.~\ref{CSD&SRRS} show that SMGARN can effectively eliminate the interference of snow streaks and fine snowflakes on the image. Rows 5 and 6 show that our method is more helpful in removing large snowflakes without creating large areas of shading.

Further, we demonstrate the snow removal capability of SMGARN in real scene in Fig.~\ref{snow100k_real} to verify the generalization ability of the proposed method (please zoom in to see details). From the figure, we can clearly see that SMGARN can remove both large particles of snow and dense snowflakes well. This is benefit from the proposed Mask-Net, which can predict accurate location and shape of the snow. With the help of the predicted snow mask,  the images reconstructed by our SMGARN have almost no snow and texture details can be better preserved.

\subsubsection{Model Size and Performance Comparison}

\begin{figure}[t]
\includegraphics[width=1\linewidth]{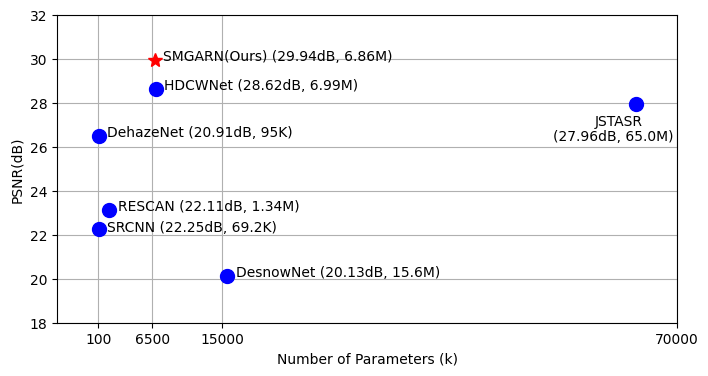}
\caption{Model Size and Performance Comparison. The results in the figure show that SMGARN achieves a more perfect balance between model parameters and performance.} 
\label{mcc}
\end{figure}

We also provide the trade-off analysis of the proposed SMGARN and other classic snow removal methods in the number of parameters and performance in Fig.~\ref{mcc}. Among them, JSTASR also uses the information of the snow mask for snow removal, but its parameters quantity is $\mathbf{10}$ times that of our SMGARN. Meanwhile, our method improves the performance of PSNR and SSIM by $\mathbf{1.98dB}$ and $\mathbf{0.06}$ respectively compared to JSTASR. Compared to HDCWNet, our method improves snow removal performance by more than $\mathbf{1dB}$ with fewer parameters. The above results fully demonstrate that the proposed SMGARN can achieve better performance with few parameters. Therefore, we can draw a conclusion that the proposed SMGARN achieves a good balance between the model performance and parameter quantity.

\begin{figure*}[t]
\begin{minipage}[c]{0.135\textwidth}
\includegraphics[width=2.5cm,height=2.5cm]{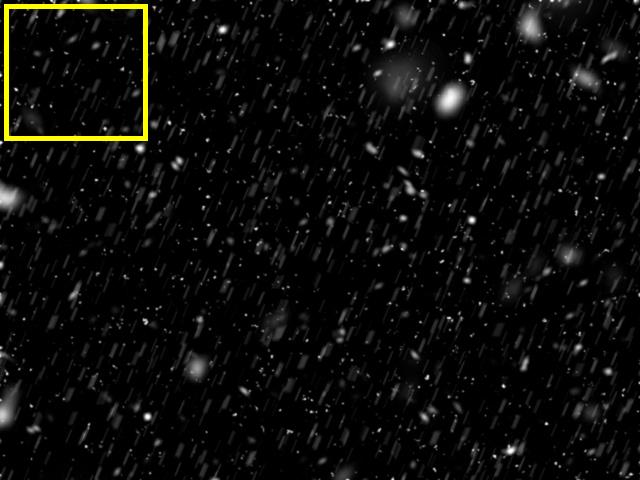}
\centerline{CSD-8002-Mask}
\centerline{}
\end{minipage}
\begin{minipage}[c]{0.135\textwidth}
\includegraphics[width=2.5cm,height=2.5cm]{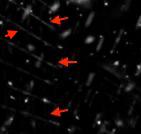}
\centerline{Baseline}
\centerline{}
\end{minipage}
\begin{minipage}[c]{0.135\textwidth}
\includegraphics[width=2.5cm,height=2.5cm]{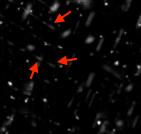}
\centerline{SA}
\centerline{}
\end{minipage}
\begin{minipage}[c]{0.135\textwidth}
\includegraphics[width=2.5cm,height=2.5cm]{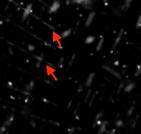}
\centerline{CA}
\centerline{}
\end{minipage}
\begin{minipage}[c]{0.135\textwidth}
\includegraphics[width=2.5cm,height=2.5cm]{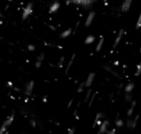}
\centerline{SA+CA (Ours)}
\centerline{}
\end{minipage}
\begin{minipage}[c]{0.135\textwidth}
\includegraphics[width=2.5cm,height=2.5cm]{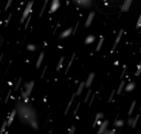}
\centerline{GT}
\centerline{}
\end{minipage}
\begin{minipage}[c]{0.135\textwidth}
\includegraphics[width=2.5cm,height=2.5cm]{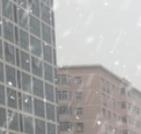}
\centerline{Snow image (input)}
\centerline{}
\end{minipage}
\caption{The effect of different attention mechanism on the quality of the snow mask predicted by Mask-Net. Obviously, lacking either SA or CA, some areas (marked by red arrows) will be disturbed by the edge of the building in the original image, resulting in the wrong snow pattern in the predicted snow mask.}
\label{snow_mask}
\end{figure*}

\begin{figure*}
\begin{minipage}[c]{0.135\textwidth}
\includegraphics[width=2.5cm,height=2.5cm]{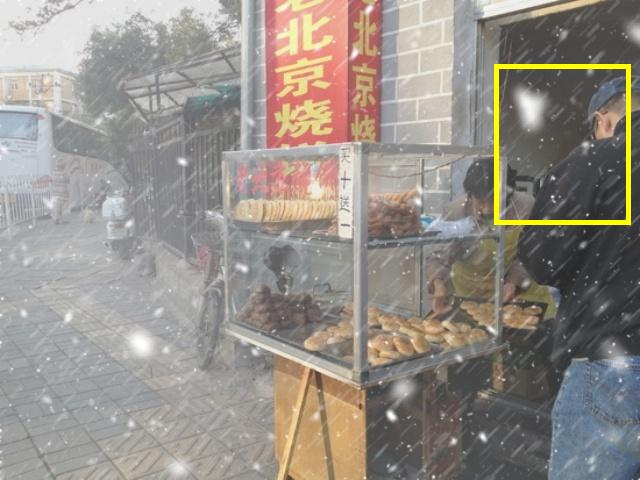}
\centerline{CSD-8602}
\end{minipage}
\begin{minipage}[c]{0.135\textwidth}
\includegraphics[width=2.5cm,height=2.5cm]{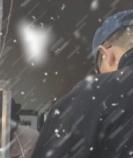}
\centerline{Snow}
\end{minipage}
\begin{minipage}[c]{0.135\textwidth}
\includegraphics[width=2.5cm,height=2.5cm]{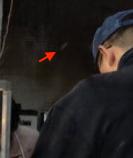}
\centerline{Case1}
\end{minipage}
\begin{minipage}[c]{0.135\textwidth}
\includegraphics[width=2.5cm,height=2.5cm]{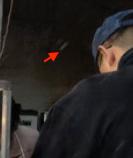}
\centerline{Case2}
\end{minipage}
\begin{minipage}[c]{0.135\textwidth}
\includegraphics[width=2.5cm,height=2.5cm]{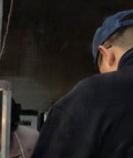}
\centerline{Case3}
\end{minipage}
\begin{minipage}[c]{0.135\textwidth}
\includegraphics[width=2.5cm,height=2.5cm]{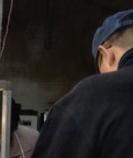}
\centerline{Case4}
\end{minipage}
\begin{minipage}[c]{0.135\textwidth}
\includegraphics[width=2.5cm,height=2.5cm]{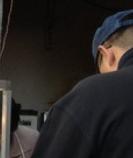}
\centerline{GT}
\end{minipage}
\caption{The effect of snow mask guidance on snow-free images reconstructed by the model. Due to the lack of guidance by the snow mask, there are residual snow patterns in Case1 and Case2 (marked by red arrows).}
\label{mask}
\end{figure*}

\section{Investigation}~\label{AS}
This section conducts a comprehensive ablation study on all modules of SMGARN to verify the effect of different modules on snow removing. We use PSNR and SSIM as evaluation metrics, all experiments are evaluated on the CSD dataset, and a patch size of $64\times64$ is used for training. In addition to the ablation studies, we also investigate multiple aspects of the model to fully validate the effectiveness of the model.

\subsection{Study on Mask-Net}
As a lightweight snow mask prediction network, Mask-Net plays an important role in extracting snow information. In order to verify the rationality of the Mask-Net, we evaluate the effect of SA (Self-Pixel Attention) and CA (Cross-Pixel Attention) in Mask-Net from both visual and quantization perspectives. In Fig.~\ref{snow_mask}, we compare the effect of different attention mechanisms on the snow mask predicted by Mask-Net. From the figure we can clearly see that the results predicted by Mask-Net with both SA and CA are very close to the ground-truth snow mask. On the contrary, if the Mask-Net without any attention unit or composed of only one attention unit, the generated snow mask will be disturbed by other objects in the image. Meanwhile, we also provide the quantitative comparison in TABLE~\ref{Attention}. According to the table, we can observed that SA and CA can significantly promote the final snow removal performance of the model. In addition, the performance of the model can be further improved when both SA and CA are adopted. This fully demonstrated the rationality of the Mask-Net.

Moreover, we provide the effect of snow mask on model performance in Fig.~\ref{mask} and TABLE~\ref{ISM}. Specifically, we use the model without Mask-Net as the baseline model, denoted as Case 1. Case 2 represents the model after removing $\mathcal{L}_{mask}$ and Case 3 is the final SMGARN model. For Case 4, we directly use the ground truth snow mask as a guide to further verify the effect of snow mask on the snow removal ability of the model. These results show that the snow mask plays a crucial role in model performance, and the quality of the snow mask will determine the effectiveness of snow removal. The higher the quality of the provided snow mask, the better the model performance. \textbf{This experiment proves the decisive role of the snow mask in the image snow removal task, making up for the deficiency of the previous research.}

\begin{table}
	\centering
	\setlength{\tabcolsep}{6mm}
    \caption{Study of Two Pixel Attention Mechanisms in Mask-Net.}
		\begin{tabular}{c|cc|c}
			\toprule
			Model Case          & SA	    & CA       &PSNR  	    \\ \midrule
			Mask-Net-baseline              &    $\times$	        &   $\times$                   &28.83 \\ 
		    Mask-Net-SA               &    \checkmark            &   $\times$                   &28.88\\
			Mask-Net-CA               &    $\times$            &   \checkmark                  &28.90\\
			Mask-Net-CASA               &    \checkmark	        &   \checkmark                   &\textbf{29.01} \\ 
			\bottomrule 
		\end{tabular}
		\label{Attention}
\end{table}

\begin{table}[t]
	\centering
	\setlength{\tabcolsep}{2mm}
    \caption{Study on the importance of snow mask. Among them, best results are \textbf{highlight} and the second best results are \textit{ITALIC}.}
		\begin{tabular}{c|cccc}
			\toprule
			Metric          & Case1	    & Case2       & Case3 (Ours)	    & Case4  	    \\ \midrule
			PSNR/SSIM       & 28.62/0.93   	        &28.92/0.93             & \textit{29.01/0.93}            & 
		\textbf{30.07/0.95}      \\ 
			\bottomrule 
		\end{tabular}
		\label{ISM}
\end{table}

\begin{figure*}[t]
\begin{minipage}[c]{0.195\textwidth}
\includegraphics[width=3.58cm,height=2cm]{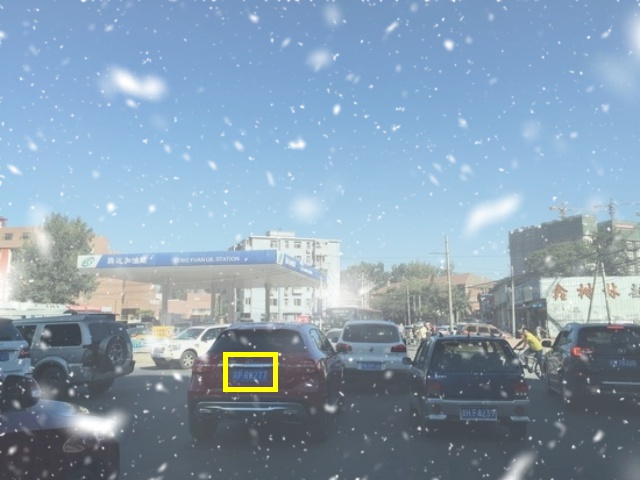}
\end{minipage}
\begin{minipage}[c]{0.195\textwidth}
\includegraphics[width=3.58cm,height=2cm]{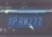}
\end{minipage}
\begin{minipage}[c]{0.195\textwidth}
\includegraphics[width=3.58cm,height=2cm]{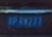}
\end{minipage}
\begin{minipage}[c]{0.195\textwidth}
\includegraphics[width=3.58cm,height=2cm]{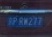}
\end{minipage}
\begin{minipage}[c]{0.195\textwidth}
\includegraphics[width=3.58cm,height=2cm]{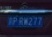}
\end{minipage}

\begin{minipage}[c]{1\textwidth}
\end{minipage}


\begin{minipage}[c]{0.195\textwidth}
\includegraphics[width=3.58cm,height=2cm]{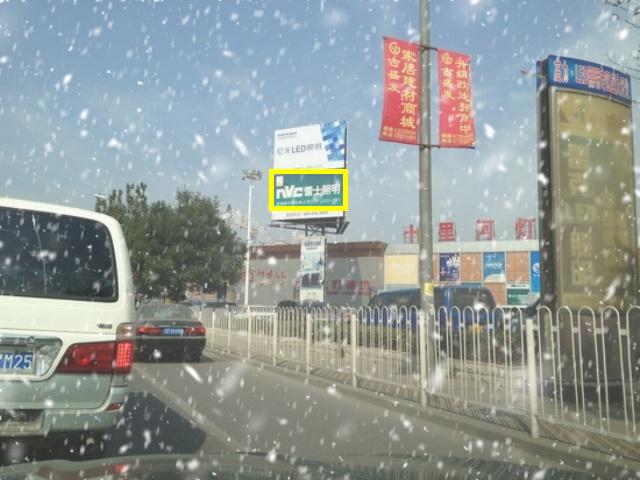}
\end{minipage}
\begin{minipage}[c]{0.195\textwidth}
\includegraphics[width=3.58cm,height=2cm]{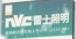}
\end{minipage}
\begin{minipage}[c]{0.195\textwidth}
\includegraphics[width=3.58cm,height=2cm]{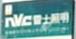}
\end{minipage}
\begin{minipage}[c]{0.195\textwidth}
\includegraphics[width=3.58cm,height=2cm]{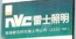}
\end{minipage}
\begin{minipage}[c]{0.195\textwidth}
\includegraphics[width=3.58cm,height=2cm]{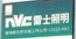}
\end{minipage}

\begin{minipage}[c]{1\textwidth}
\end{minipage}

\begin{minipage}[c]{0.195\textwidth}
\includegraphics[width=3.58cm,height=2cm]{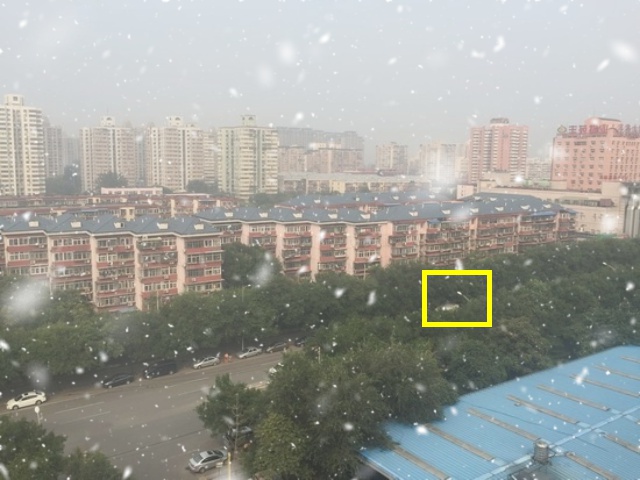}
\centerline{Original Image}
\centerline{}
\end{minipage}
\begin{minipage}[c]{0.195\textwidth}
\includegraphics[width=3.58cm,height=2cm]{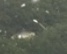}
\centerline{Snowy Image}
\centerline{}
\end{minipage}
\begin{minipage}[c]{0.195\textwidth}
\includegraphics[width=3.58cm,height=2cm]{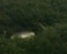}
\centerline{HDCWNet~\cite{chen2021all}}
\centerline{}
\end{minipage}
\begin{minipage}[c]{0.195\textwidth}
\includegraphics[width=3.58cm,height=2cm]{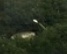}
\centerline{SMGARN (Ours)}
\centerline{}
\end{minipage}
\begin{minipage}[c]{0.195\textwidth}
\includegraphics[width=3.58cm,height=2cm]{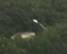}
\centerline{GT}
\centerline{}
\end{minipage}
\caption{Comparison with HDCWNet on the quality of snow-free images. Obviously, our SMGARN can reconstruct more clear images.}
\label{IQC}
\end{figure*}

\begin{table}[t]
	\centering
	\setlength{\tabcolsep}{2.2mm}
    \caption{Investigation of GF-Net.}
		\begin{tabular}{c|cccc}
			\toprule
			Metric 	    & Case1       & Case2	        & Case3    & Case4  	    \\ \midrule
			PSNR/SSIM  	&27.22/0.91  &27.31/0.91 &28.39/0.92          & 29.01/0.93      \\ 
			\bottomrule 
		\end{tabular}
		\label{IGF}
\end{table}

\begin{table}[t]
	\centering
	\setlength{\tabcolsep}{0.8mm}
    \caption{Comparison of Different Network Structures in the Proposed MARB. }
		\begin{tabular}{c|cccc|c}
			\toprule
			Model Case          & Single-Scale	    & Multi-Scale   & Single-Agg & Multi-Agg       &PSNR  	    \\ \midrule
			MARN-SS-SA              &    \checkmark	        &   $\times$       &\checkmark      &$\times$             &28.64 \\ 
		    MARN-MS-SA               &    $\times$            &   \checkmark      &\checkmark   & $\times$          &28.75\\
			MARN-SS-MA               &    \checkmark            & $\times$      & $\times$     & \checkmark           &28.72\\
			MARN-MS-MA               &  $\times$    & \checkmark      & $\times$        &\checkmark              & \textbf{29.01}\\ 
			\bottomrule 
		\end{tabular}
		\label{MARB}
\end{table}

\subsection{Study on GF-Net}
\label{sec:Investigation of GF-Net}
As the key structure of SMGARN, GF-Net plays an important role in guiding snow mask to remove snow. In TABLE~\ref{IGF}, we investigate the effectiveness of GF-Net. Specifically, in Case 1, we concatenate the snow mask with the features of the snow image and replace the GF-Net with 8 convolutional layers. In Case 2, we replace the feature connection operation with the residual operation based on Case 1. It can be seen that since the snow image is a weighted combination of the snow mask and the clean image, the performance of Case 2 is significantly higher than that of Case 1. To further verify the rationality of GF-Net, we replace the adaptive residual in the original GF-Net with the connection operation and represent it as Case 3. Compared with the original model Case 4, we can find that the performance of Case 3 is severely degraded. This means that the residual operation of adaptive coding is more in line with the snow image generation mechanism, and has better snow removal ability in high-dimensional feature space, which is very beneficial to performance improvement. \textbf{\textit{In summary, Case 1 and Case 2 prove the effectiveness of residual operation for snow removal. Case 3 and Case 4 demonstrate that adaptive residual structure is more beneficial to reconstruct snow-free images than feature connections.}} This experiment fully illustrate the effectiveness and necessity of the proposed GF-Net.


\begin{figure}[t]
\begin{center}
\includegraphics[width=1\linewidth]{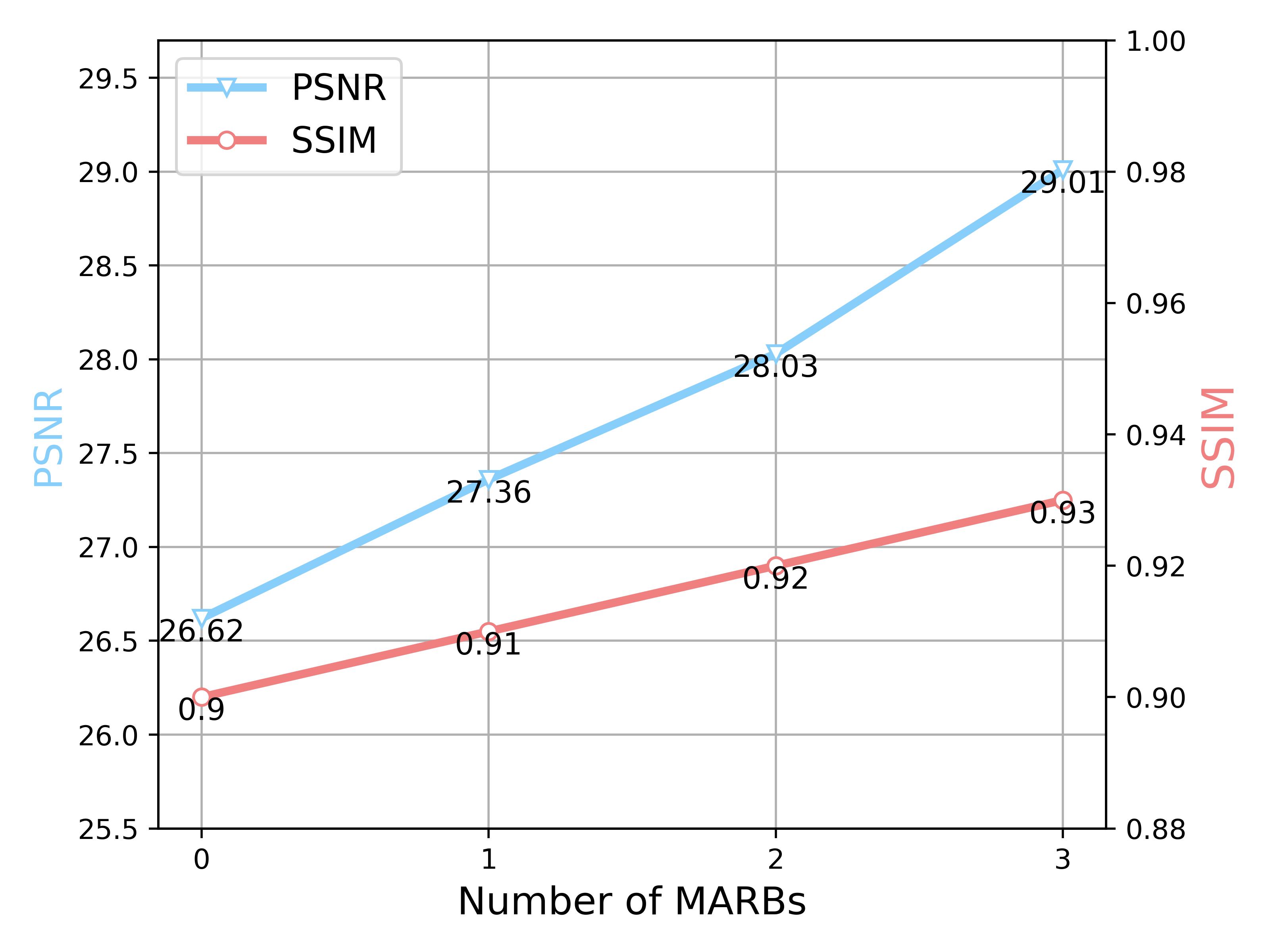}
\end{center}
\caption{Ablation study on different number of MARBs.} 
\label{IRN}
\end{figure}


\subsection{Study on Reconstruct-Net}
In this work, we propose a Reconstruct-Net to reconstruct the final snow-free image by the specially designed MARBs. In In TABLE~\ref{MARB}, we provide the results for several different MARB designs. Single-Scale (SS) means that the three input branches use $3\times3$ convolutional layers for feature extraction in MARB. Multi-Scale (MS) means that MARB uses convolutional layers with kernels of $1\times1$, $3\times3$, and $5\times5$ to process features in parallel. Single-Agg (SA) means that only one join operation is applied in MARB. Multi-Agg (MA) means that MARB aggregates features from different branches multiple times. It can be seen from the table that the multi-scale and multi-aggregation design can significantly improve the model performance, which proves the effectiveness of MARB.

\begin{figure*}[t]
\begin{minipage}[c]{0.16\textwidth}
\includegraphics[width=2.9cm,height=1.8cm]{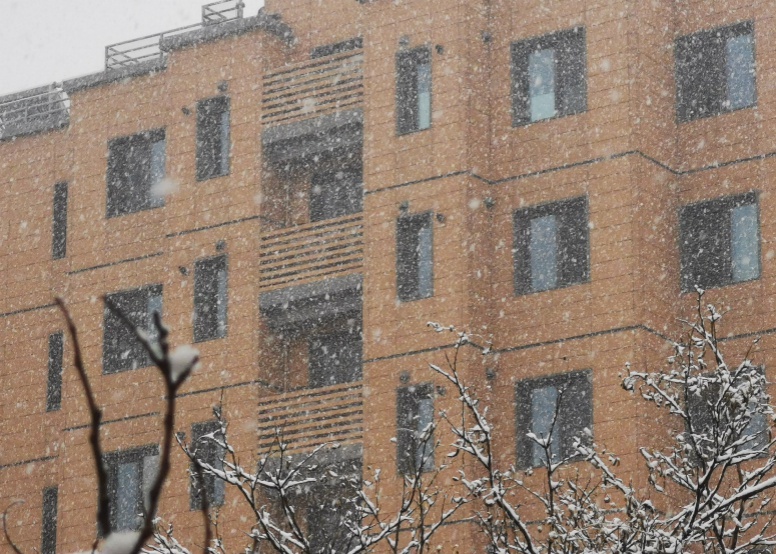}
\end{minipage}
\begin{minipage}[c]{0.16\textwidth}
\includegraphics[width=2.9cm,height=1.8cm]{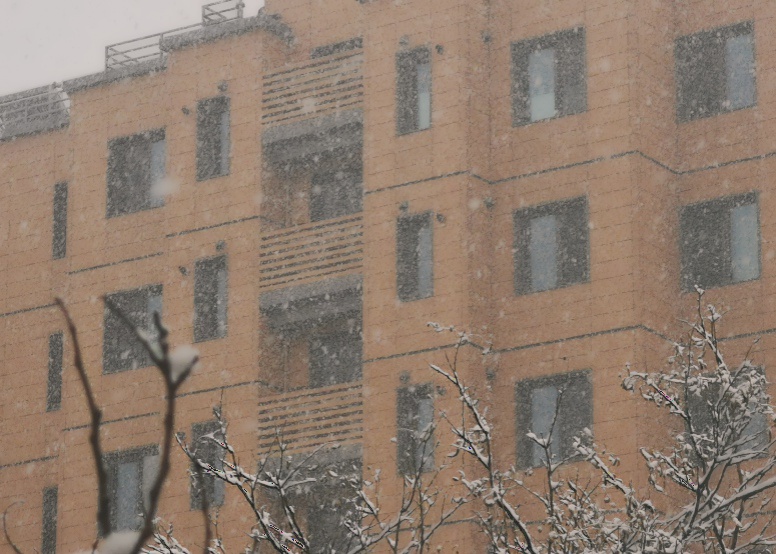}
\end{minipage}
\begin{minipage}[c]{0.16\textwidth}
\includegraphics[width=2.9cm,height=1.8cm]{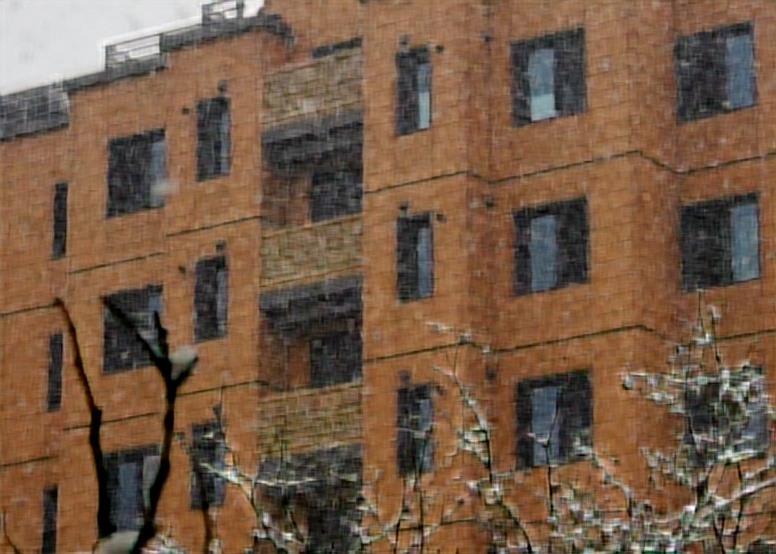}
\end{minipage}
\begin{minipage}[c]{0.16\textwidth}
\includegraphics[width=2.9cm,height=1.8cm]{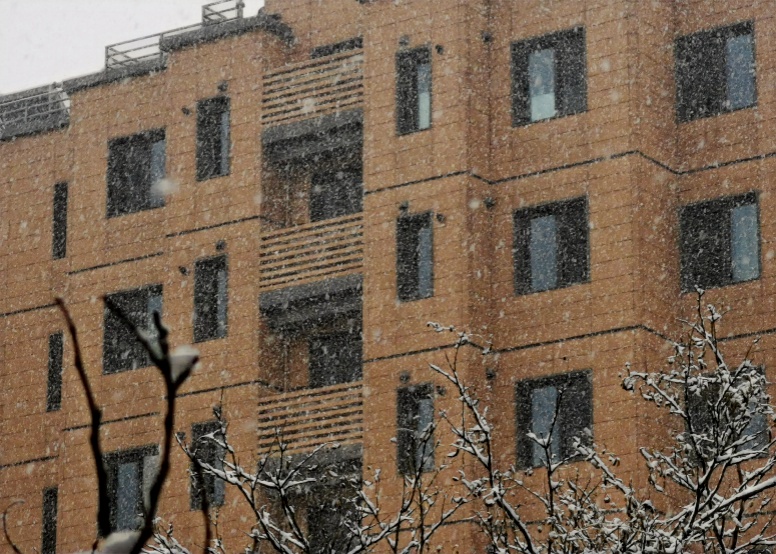}
\end{minipage}
\begin{minipage}[c]{0.16\textwidth}
\includegraphics[width=2.9cm,height=1.8cm]{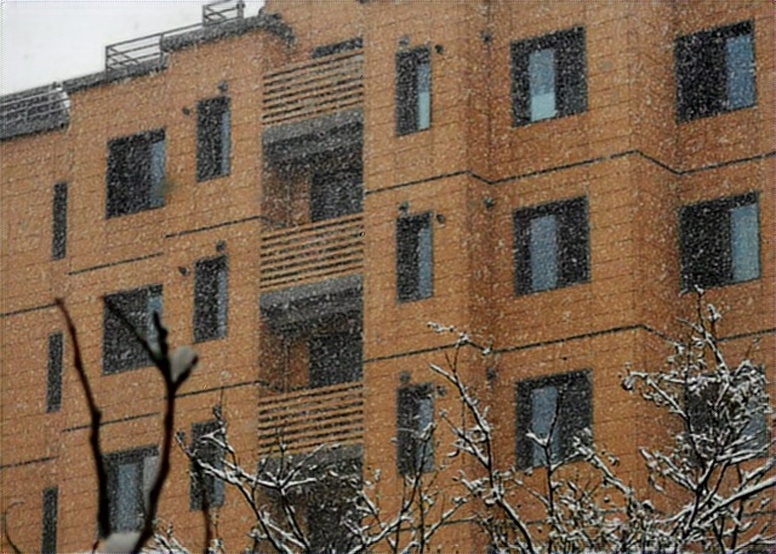}
\end{minipage}
\begin{minipage}[c]{0.16\textwidth}
\includegraphics[width=2.9cm,height=1.8cm]{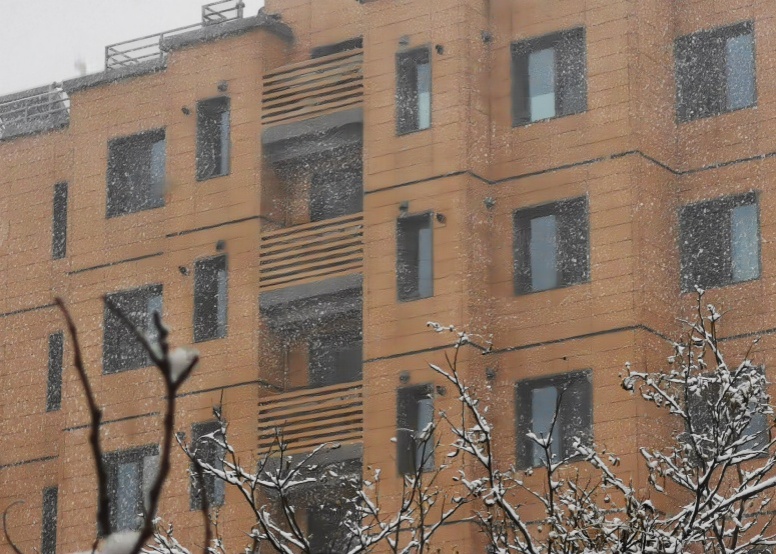}
\end{minipage}

\begin{minipage}[c]{1\textwidth}
\end{minipage}

\begin{minipage}[c]{0.16\textwidth}
\includegraphics[width=2.9cm,height=2.8cm]{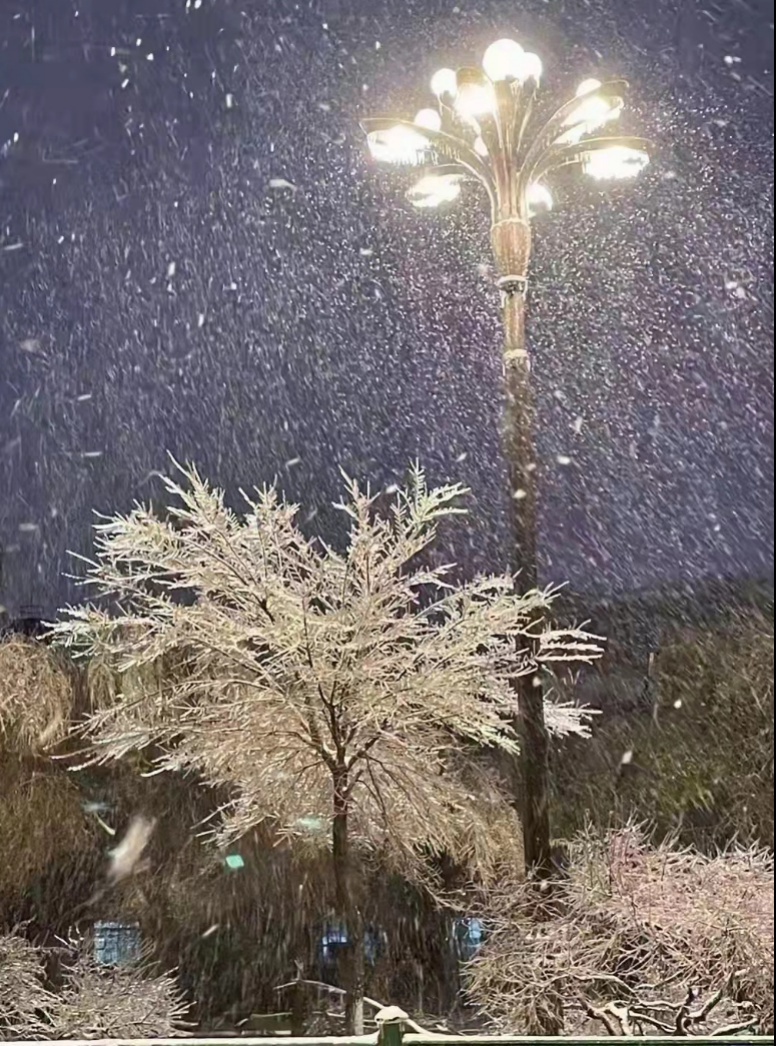}
\end{minipage}
\begin{minipage}[c]{0.16\textwidth}
\includegraphics[width=2.9cm,height=2.8cm]{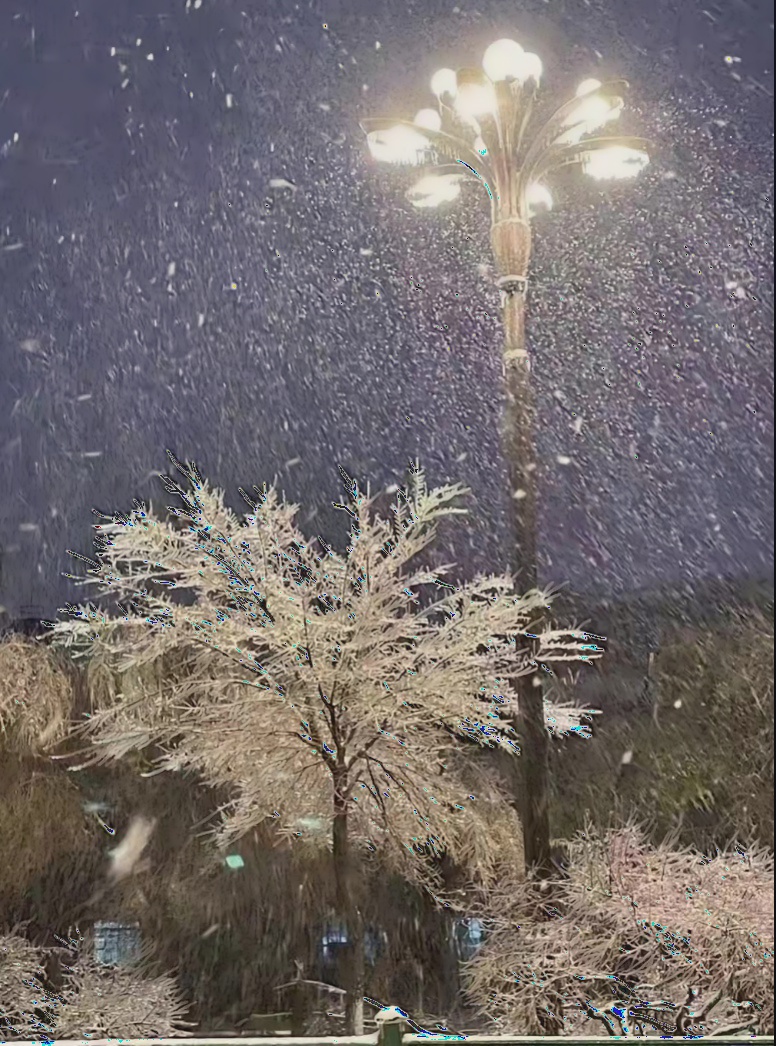}
\end{minipage}
\begin{minipage}[c]{0.16\textwidth}
\includegraphics[width=2.9cm,height=2.8cm]{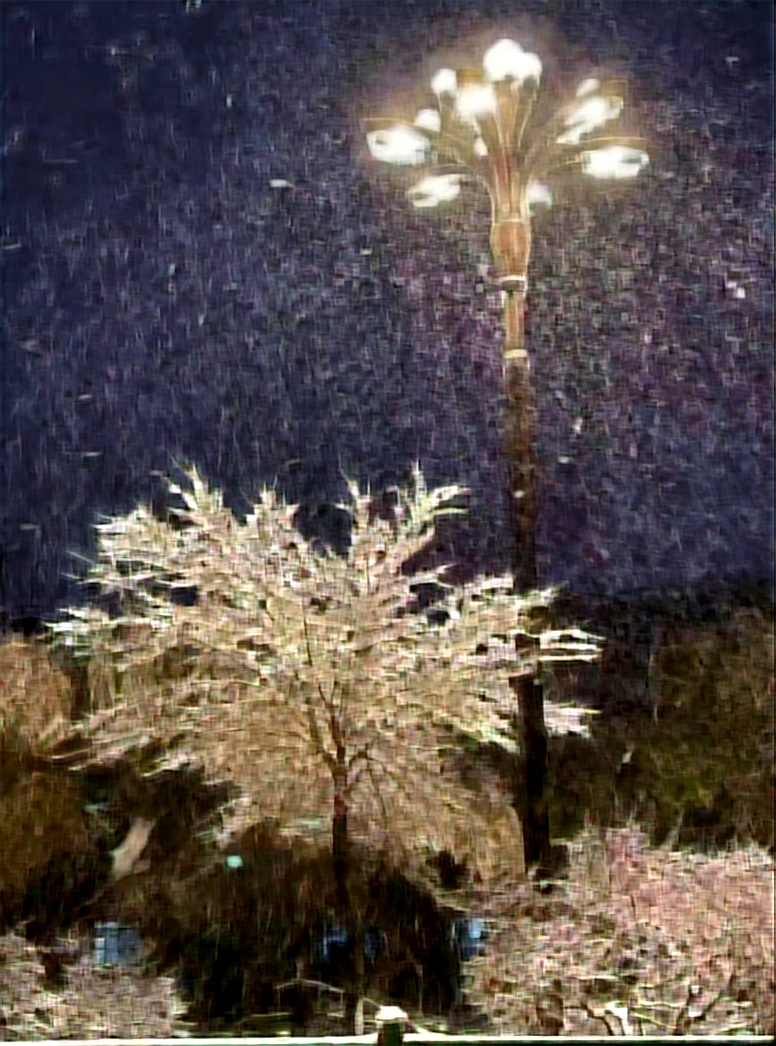}
\end{minipage}
\begin{minipage}[c]{0.16\textwidth}
\includegraphics[width=2.9cm,height=2.8cm]{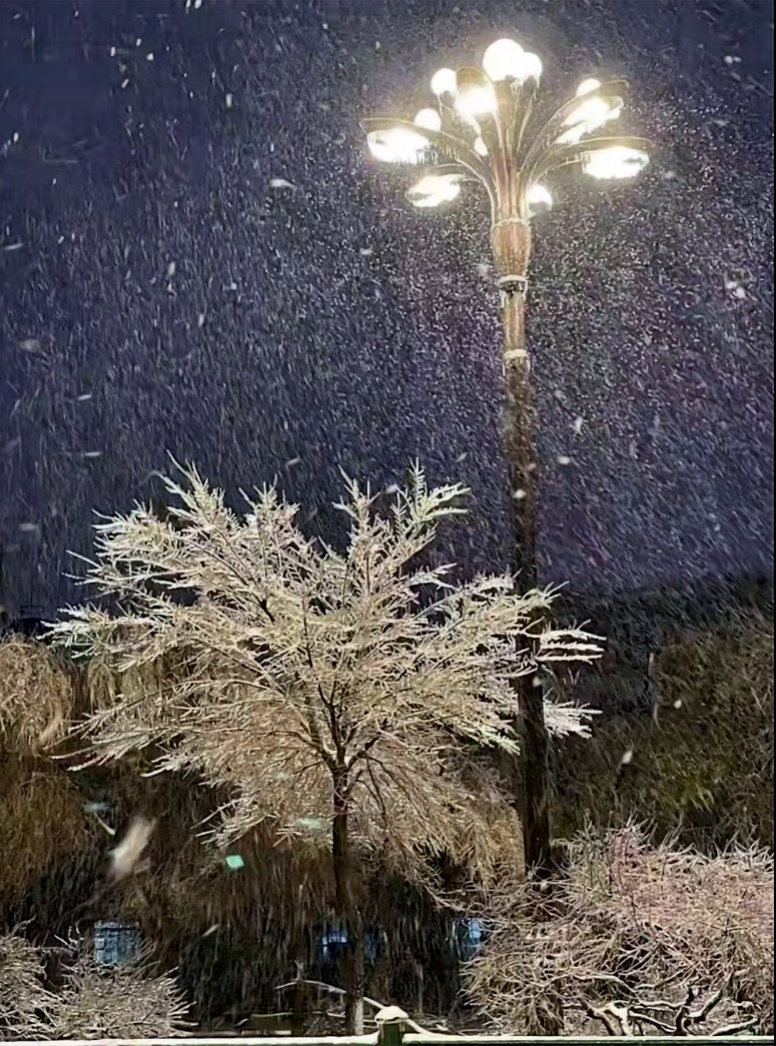}
\end{minipage}
\begin{minipage}[c]{0.16\textwidth}
\includegraphics[width=2.9cm,height=2.8cm]{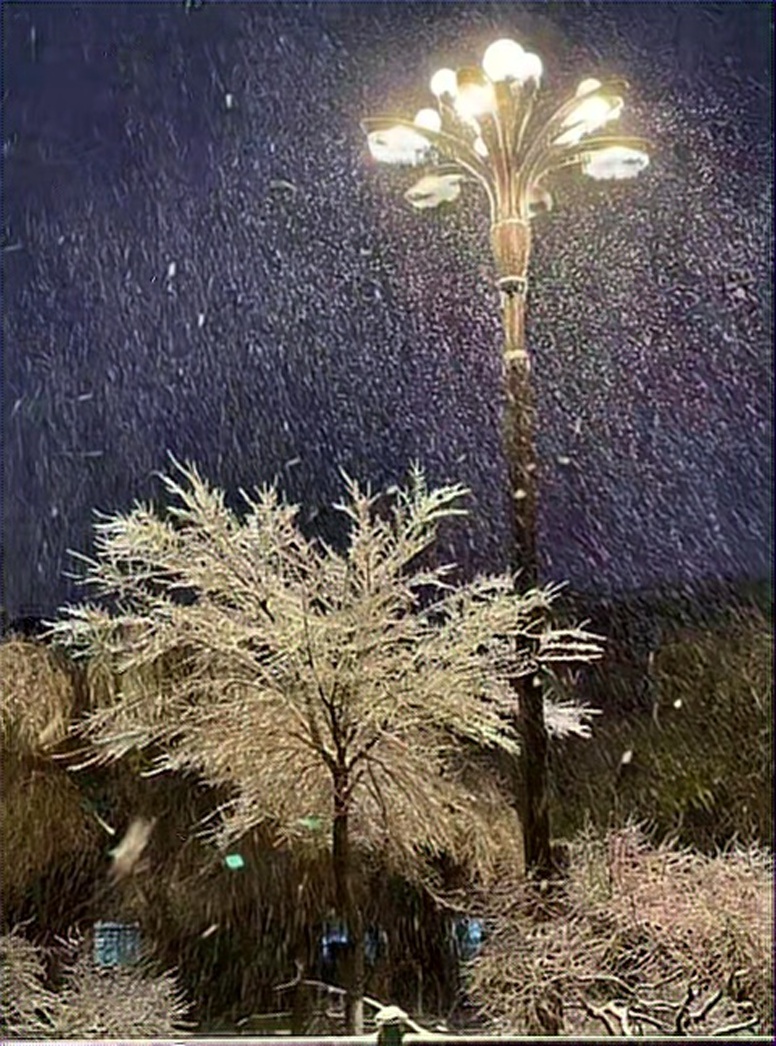}
\end{minipage}
\begin{minipage}[c]{0.16\textwidth}
\includegraphics[width=2.9cm,height=2.8cm]{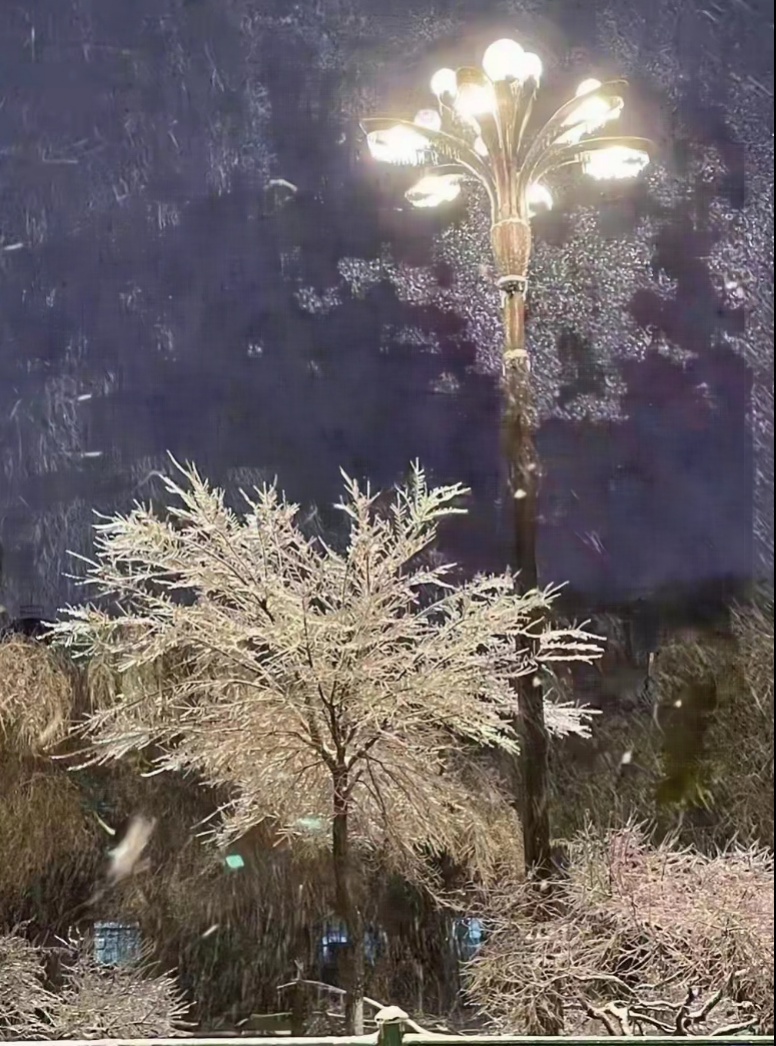}
\end{minipage}

\begin{minipage}[c]{1\textwidth}
\end{minipage}

\begin{minipage}[c]{0.16\textwidth}
\includegraphics[width=2.9cm,height=2.8cm]{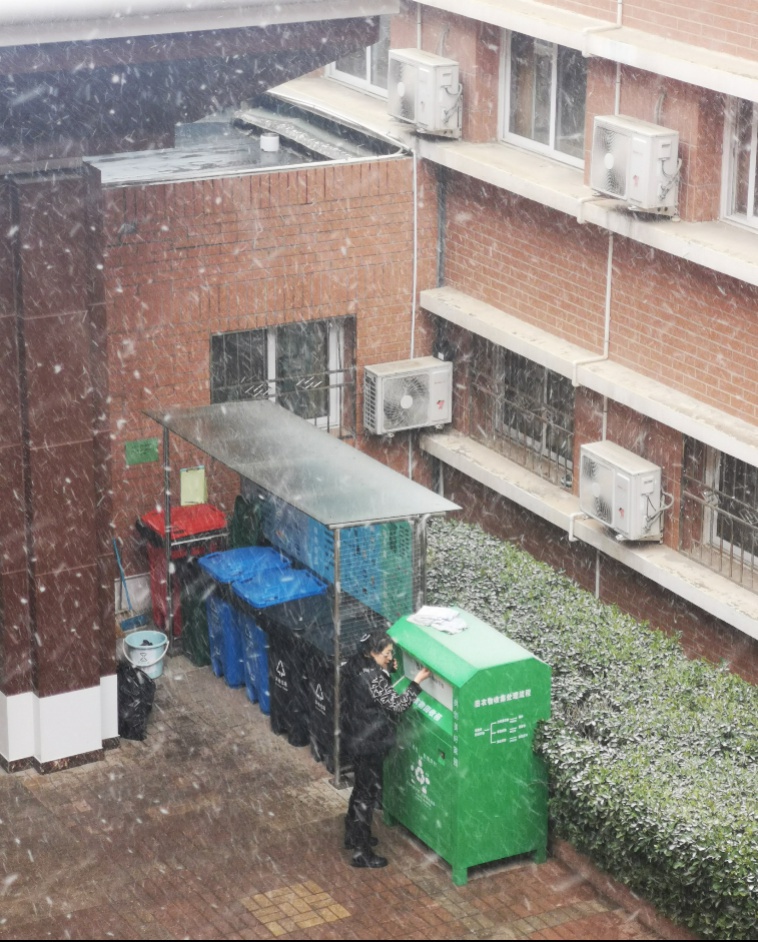}
\centerline{Snowy Image}
\centerline{}
\end{minipage}
\begin{minipage}[c]{0.16\textwidth}
\includegraphics[width=2.9cm,height=2.8cm]{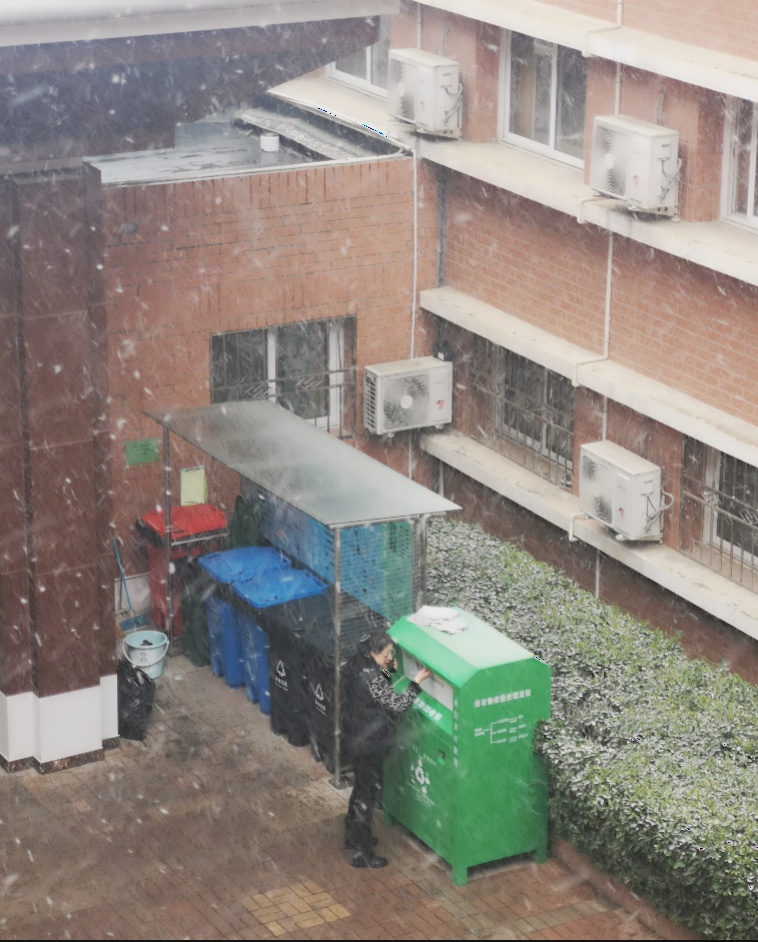}
\centerline{Zheng~\cite{zheng2013single}}
\centerline{}
\end{minipage}
\begin{minipage}[c]{0.16\textwidth}
\includegraphics[width=2.9cm,height=2.8cm]{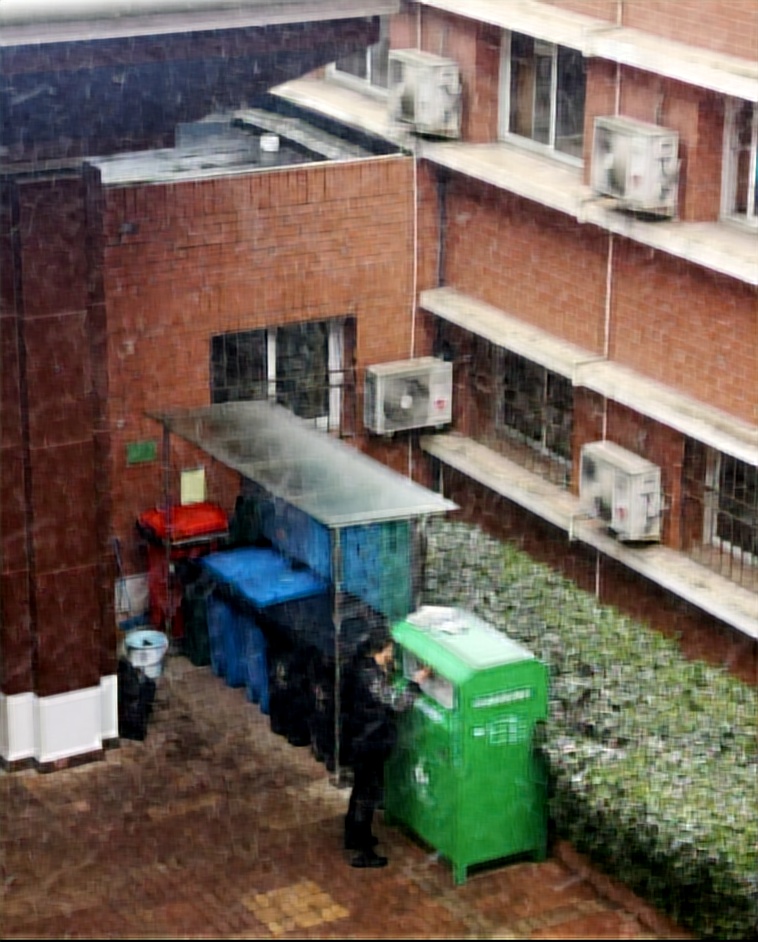}
\centerline{DehazeNet~\cite{cai2016dehazenet}}
\centerline{}
\end{minipage}
\begin{minipage}[c]{0.16\textwidth}
\includegraphics[width=2.9cm,height=2.8cm]{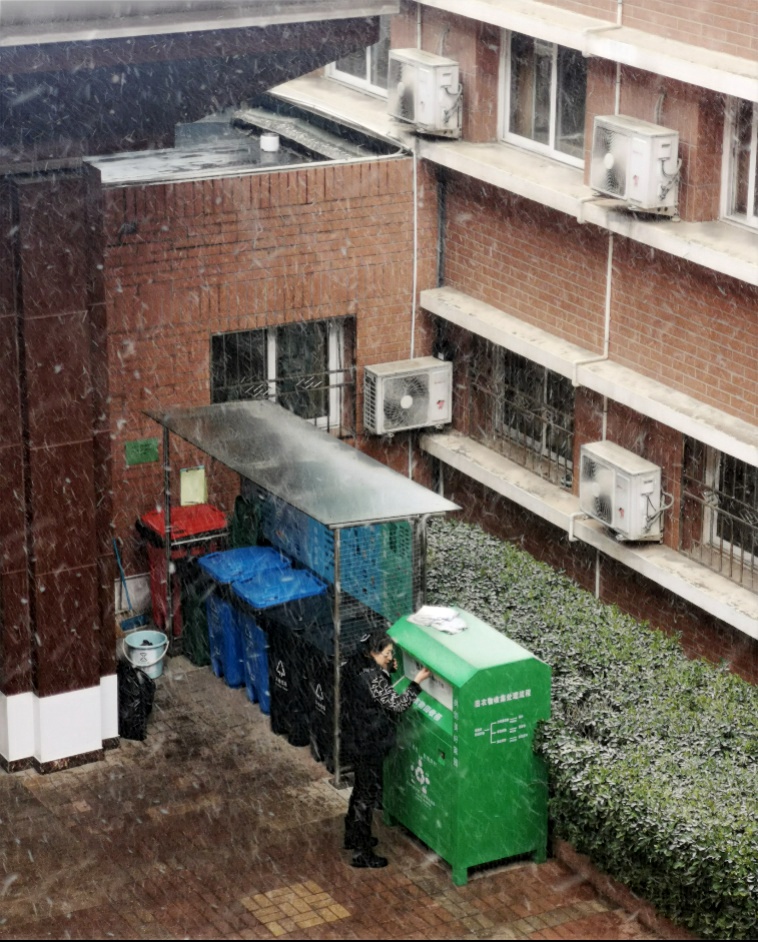}
\centerline{RESCAN~\cite{li2018recurrent}}
\centerline{}
\end{minipage}
\begin{minipage}[c]{0.16\textwidth}
\includegraphics[width=2.9cm,height=2.8cm]{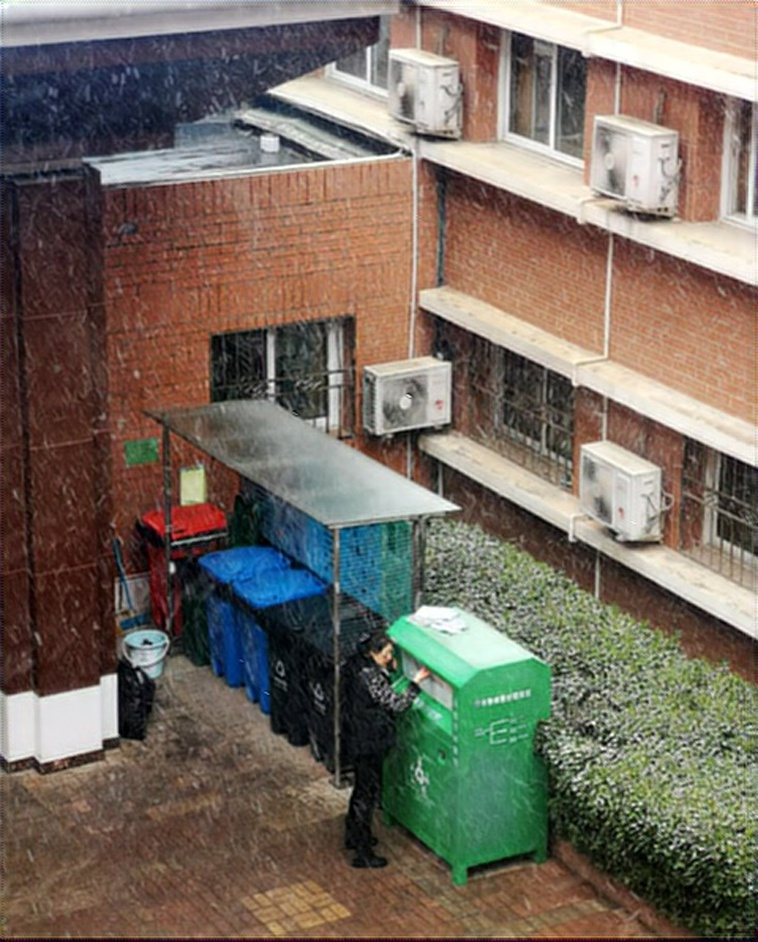}
\centerline{HDCWNet~\cite{chen2021all}}
\centerline{}
\end{minipage}
\begin{minipage}[c]{0.16\textwidth}
\includegraphics[width=2.9cm,height=2.8cm]{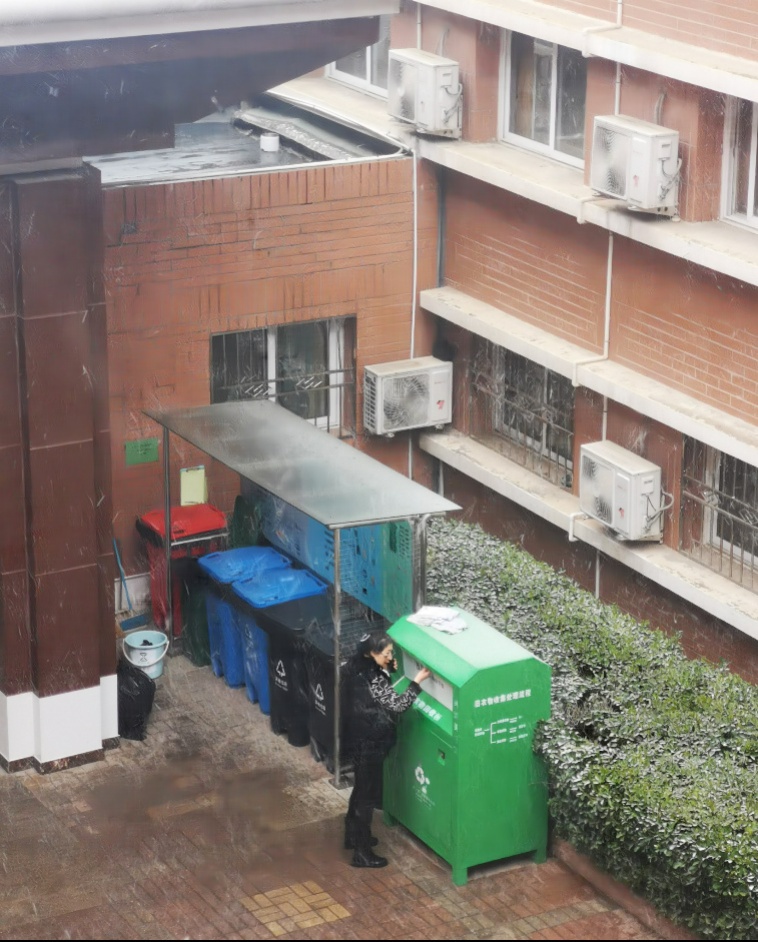}
\centerline{\textbf{SMGARN(Ours)}}
\centerline{}
\end{minipage}
\caption{Visual comparison on real snow images from SnowWorld24. Obviously, our proposed method can reconstruct clear images with less snow residue.}
\label{SnowWorld2}
\end{figure*}

To verify the effect of MARB on the model performance, we also tested the model with different numbers of MARBs. According to Fig.~\ref{IRN}, we can clearly observe that the performance of the model increases as the number of MARBs increases. This illustrates the effectiveness of MARB and the potential of Reconstructe-Net. However, we also noticed that as the number of MARBs increases, the number of parameters of the model also increases. Therefore, we used three MARBs in the final model to achieve a good balance between the model performance and size. 

\subsection{Image Quality Comparison}
For the image snow removal task, it is not only necessary to eliminate the interference of snow in the image, but also to retain the information in the original image to the maximum extent possible for the reconstructed image. In this part, we evaluate the image quality reconstructed by the proposed method to further verify the effectiveness of SMGARN. In TABLE~\ref{NIQE}, we show the NIQE results of our SMGARN and other advanced methods. NIQE is a fully blind image quality analyzer that can be used to assess the authenticity of images. \textbf{\textit{It is worth noting that lower NIQE values represent better perceptual quality. }} According to the table, we can found that our method has a lower NIQE, proving that it can generate higher quality snow-free images. In addition, we also evaluated the image quality from the visual aspect. According to the first two lines of images in Figure~\ref{IQC}, it can be clearly seen that the images generated by SMGARN well retain the clear text information in the original images. In contrast, the images generated by HDCWNet blur the details of text regions, which affects the image quality. In addition, the last line of Figure~\ref{IQC} shows that SMGARN can accurately distinguish the original information of the image and the snow, and will not remove other objects as snow particles. This further demostrated that our proposed SMGARN can generate high-quality images and have better robustness to be compatible with more scenes.

\begin{table}[t]
	\centering
	\setlength{\tabcolsep}{2.6mm}
    \caption{Performance comparison on real-world dataset. $\downarrow$ indicates that the lower the NIQE value, the higher the quality of the reconstructed image.}
		\begin{tabular}{c|ccc}
			\toprule
			Model         & SnowWorld24 (NIQE$\downarrow$)	    &Snow100K (NIQE$\downarrow$)    \\\midrule
			Zheng~\cite{zheng2013single}                           & 4.3762 	                & 3.9886            \\ 
			DehazeNet~\cite{cai2016dehazenet}                           & 4.6954 	                & 4.5731           \\ 
		    RESCAN~\cite{li2018recurrent}                              & 3.0795                   & 3.3893                \\
			HDCWNet~\cite{chen2021all}                             & 3.8623                   & 4.0448           \\
			SMGARN(Ours)                        & \textbf{2.9371}  	            & \textbf{3.0751}         \\ 
			\bottomrule 
		\end{tabular}
		\label{NIQE}
\end{table}

\subsection{Study on SnowWorld24}
To further verify the snow removal ability of the model in the real snow scene, we test SMGARN on our proposed SnowWorld24. SnowWorld24 is real snow image dataset that can fully verify the model performance in the real scene. According to Fig.~\ref{SnowWorld2}, we can clearly see that HDCWNet can only remove part of the snow with large particles, and can do nothing for dense snowflakes. Although HDCWNet can improve the contrast of the image, it does not achieve the original intention of removing snow from the image. In contrast, our SMGARN can remove almost snow in the image and fully retain the texture information of the image. All the above results further illustrate the effectiveness and robustness of the proposed SMGARN.

\begin{table}[t]
	\centering
	\setlength{\tabcolsep}{2.2mm}
    \caption{Questionnaire results. Large Snowflake,  Dense Snowflakes, and Quality scores range from 1 to 5, with higher scores indicating better images. The first and second metrics represent the ability of the model to remove large and dense snowflakes, respectively. Quality represents the clarity and fidelity of the image.}
		\begin{tabular}{c|ccc}
			\toprule
			Model Name         & Big Snowflake	    &Dense Snowflakes   &Quality  \\\midrule
			DehazeNet~\cite{cai2016dehazenet}                           & 2.38 	                & 1.74            & 0.46        \\ 
		    RESCAN~\cite{li2018recurrent}                              & 2.75                   & 2.06           & 1.21         \\
			HDCWNet~\cite{chen2021all}                             & 3.41                   & 2.96           & 3.16        \\
			SMGARN(Ours)                        & \textbf{4.28}  	            & \textbf{3.84}        & \textbf{3.92}           \\ 
			\bottomrule 
		\end{tabular}
		\label{Questionnaire_t}
\end{table}

\subsection{User Study}
\begin{figure}[t]
\begin{minipage}[c]{0.09\textwidth}
\includegraphics[width=1.6cm, height=1.4cm]{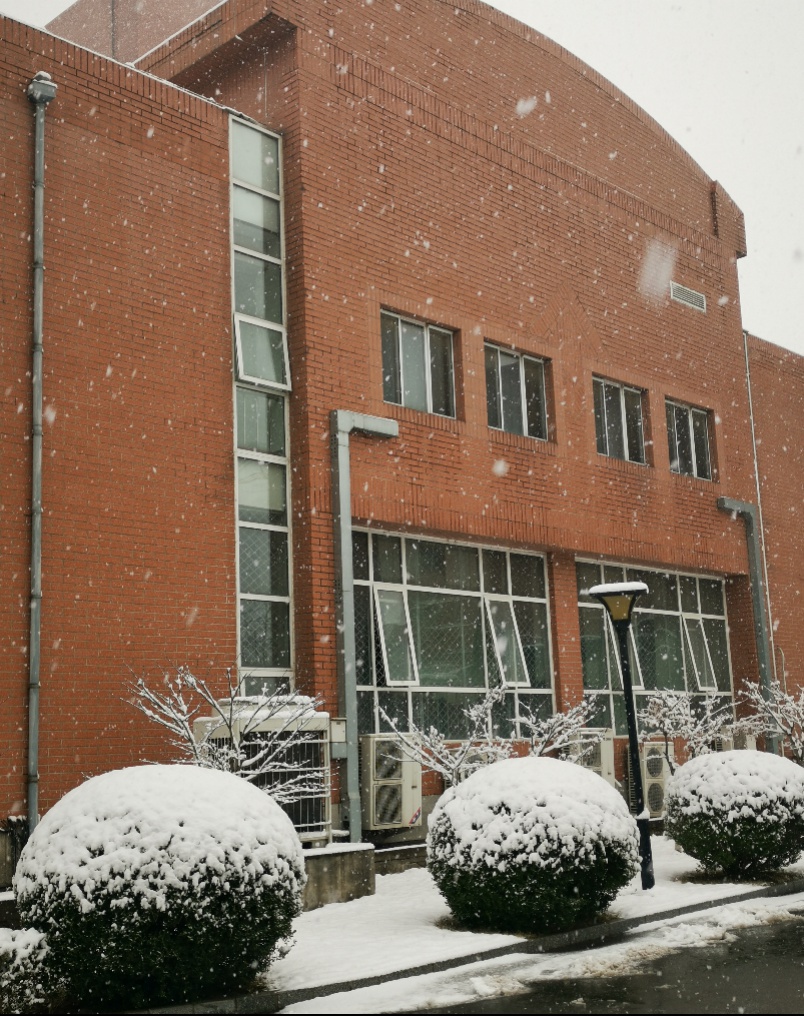}
\end{minipage}
\begin{minipage}[c]{0.09\textwidth}
\includegraphics[width=1.6cm, height=1.4cm]{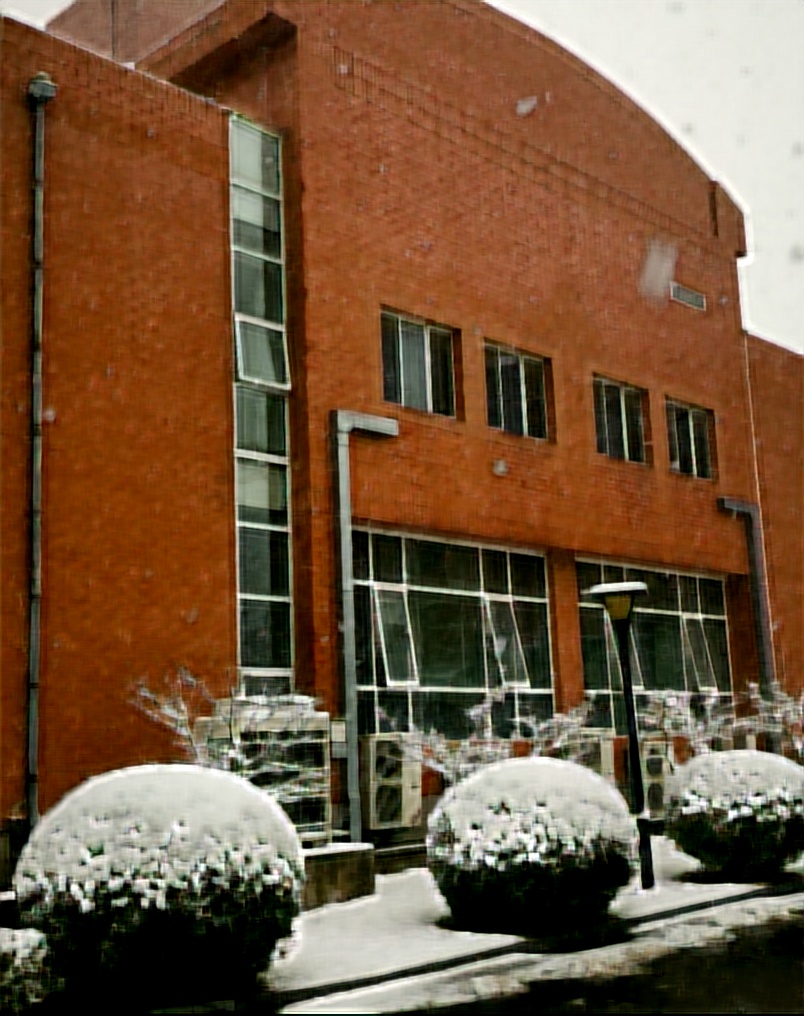}
\end{minipage}
\begin{minipage}[c]{0.09\textwidth}
\includegraphics[width=1.6cm, height=1.4cm]{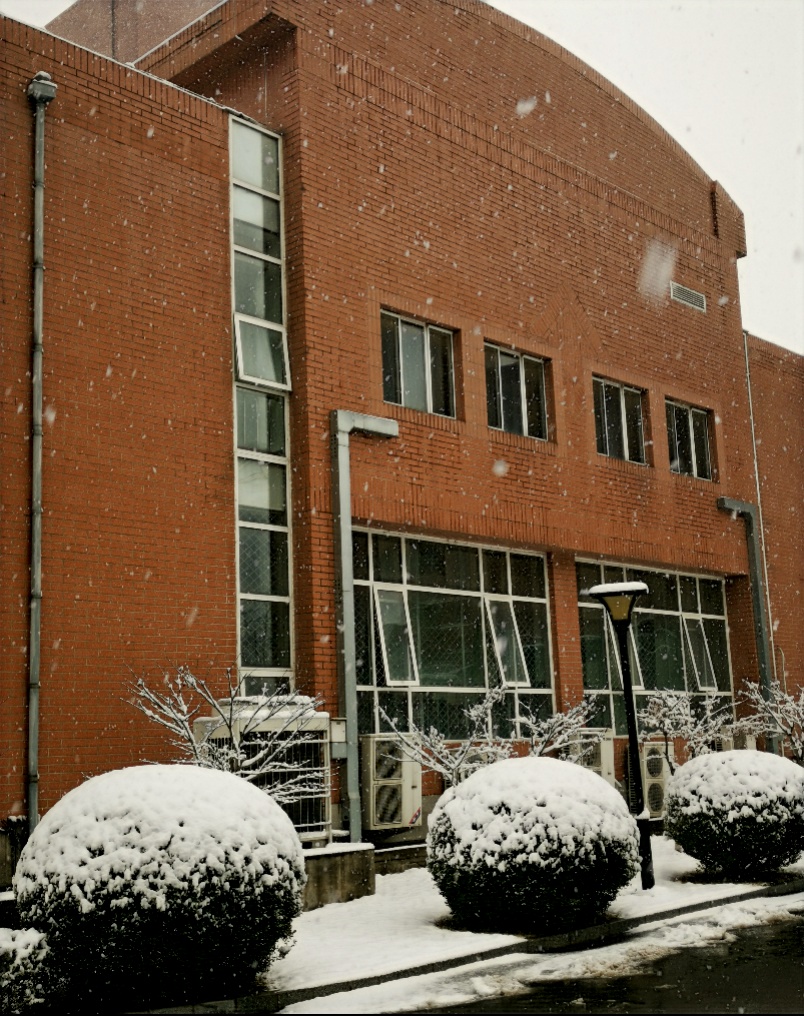}
\end{minipage}
\begin{minipage}[c]{0.09\textwidth}
\includegraphics[width=1.6cm, height=1.4cm]{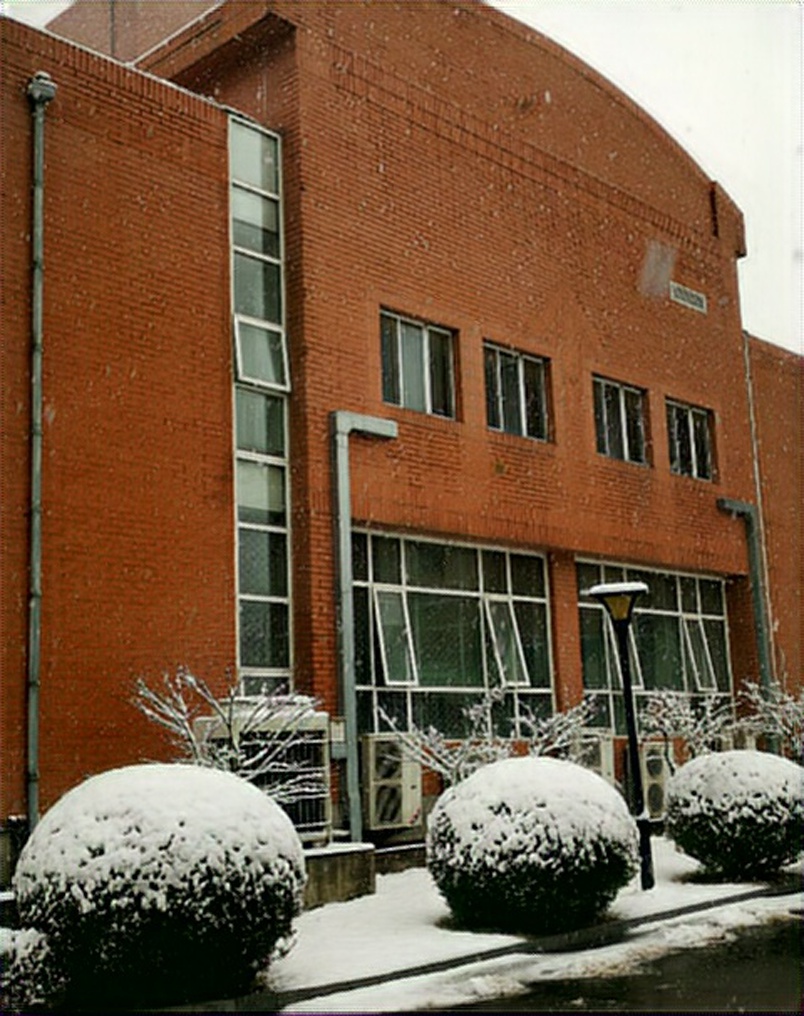}
\end{minipage}
\begin{minipage}[c]{0.09\textwidth}
\includegraphics[width=1.6cm, height=1.4cm]{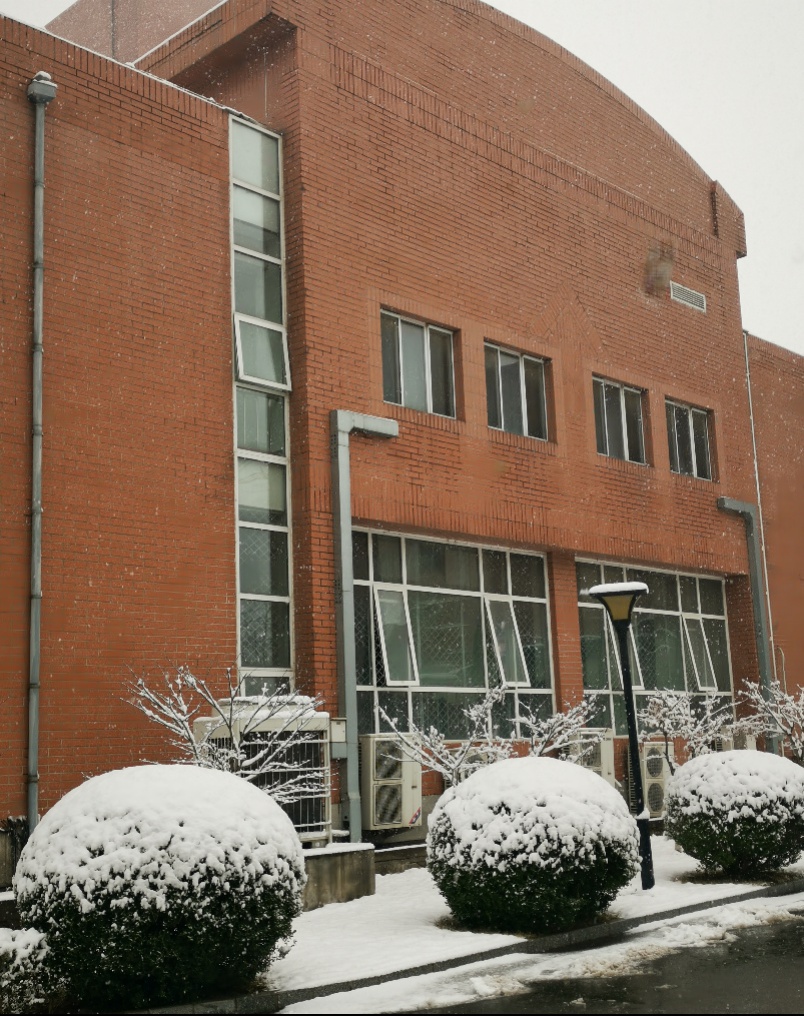}
\end{minipage}

\begin{minipage}[c]{1\textwidth}
\end{minipage}

\begin{minipage}[c]{0.09\textwidth}
\includegraphics[width=1.6cm, height=1.4cm]{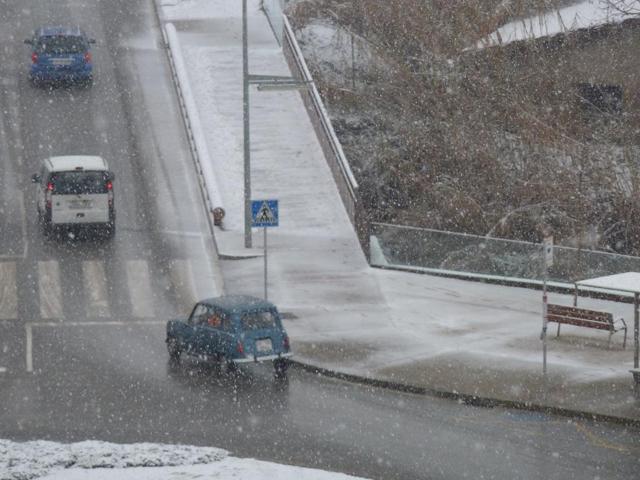}
\end{minipage}
\begin{minipage}[c]{0.09\textwidth}
\includegraphics[width=1.6cm, height=1.4cm]{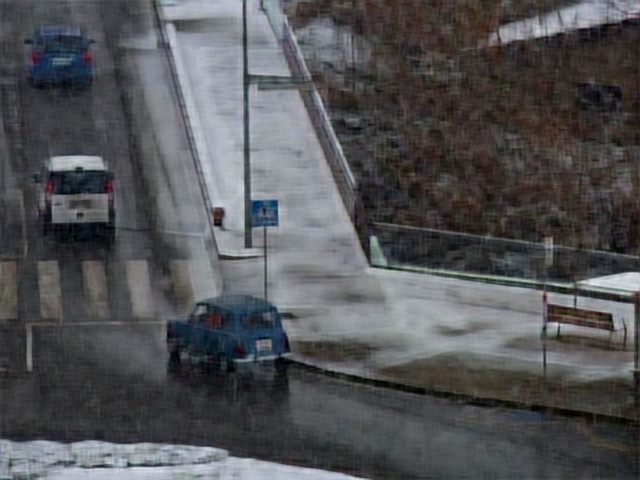}
\end{minipage}
\begin{minipage}[c]{0.09\textwidth}
\includegraphics[width=1.6cm, height=1.4cm]{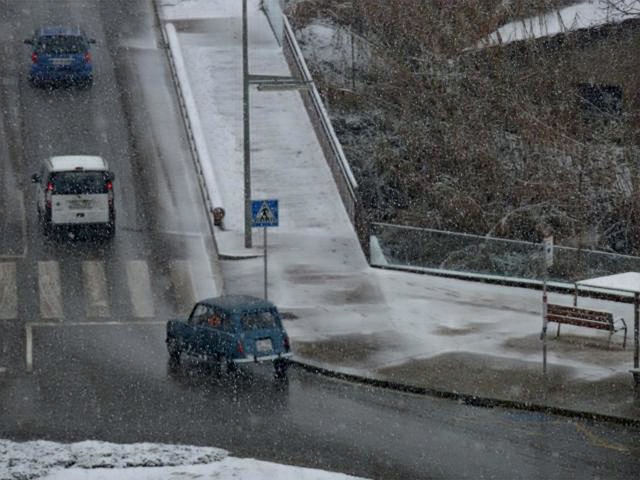}
\end{minipage}
\begin{minipage}[c]{0.09\textwidth}
\includegraphics[width=1.6cm, height=1.4cm]{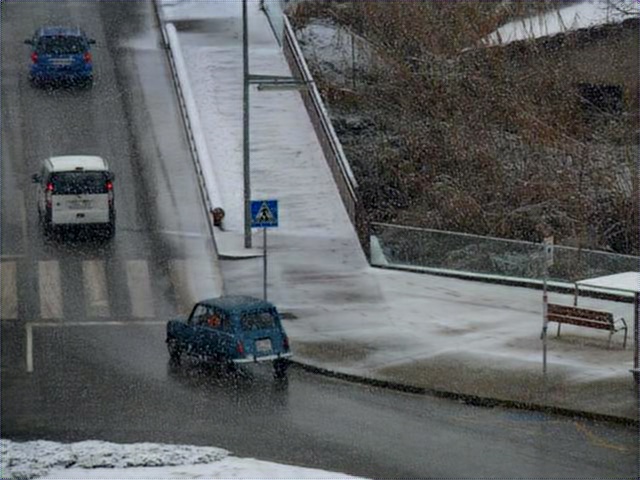}
\end{minipage}
\begin{minipage}[c]{0.09\textwidth}
\includegraphics[width=1.6cm, height=1.4cm]{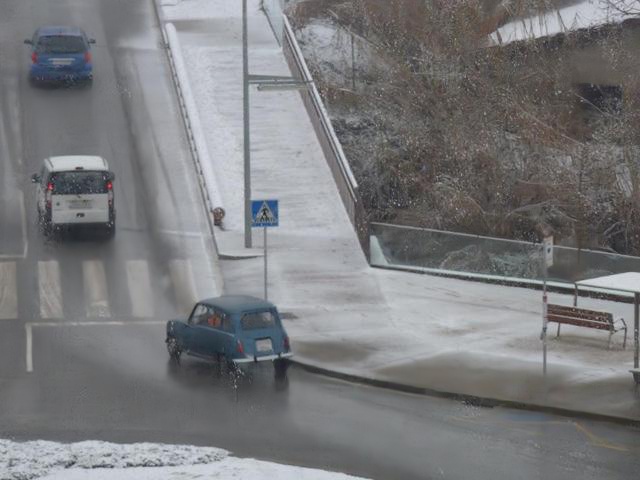}
\end{minipage}

\begin{minipage}[c]{1\textwidth}
\end{minipage}

\begin{minipage}[c]{0.09\textwidth}
\includegraphics[width=1.6cm, height=1.4cm]{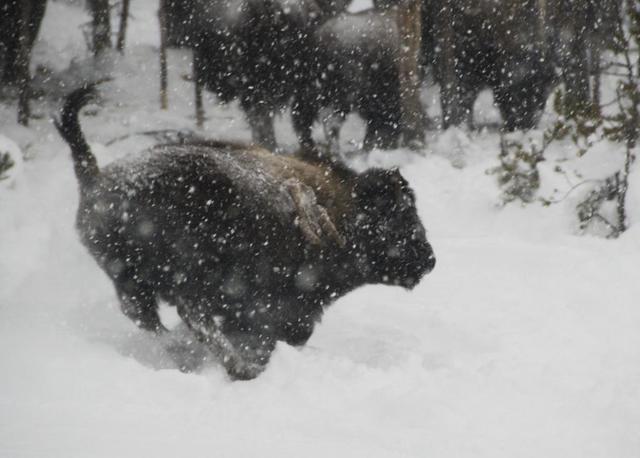}
\end{minipage}
\begin{minipage}[c]{0.09\textwidth}
\includegraphics[width=1.6cm, height=1.4cm]{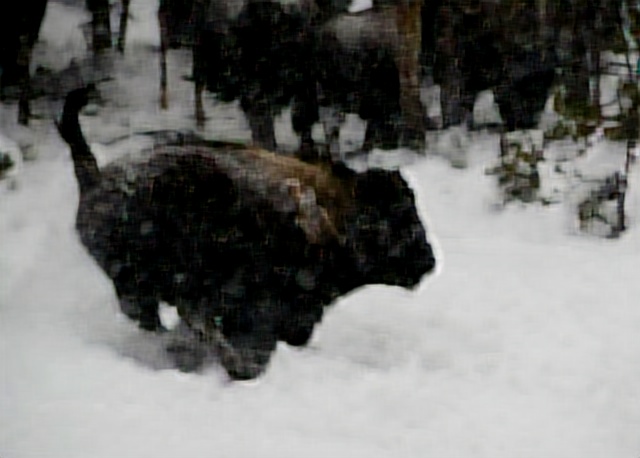}
\end{minipage}
\begin{minipage}[c]{0.09\textwidth}
\includegraphics[width=1.6cm, height=1.4cm]{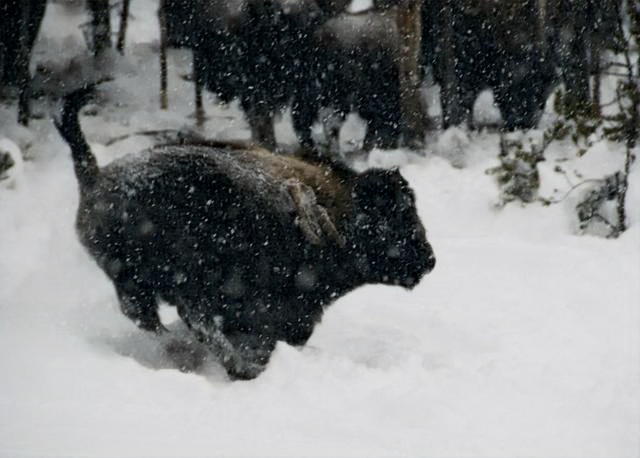}
\end{minipage}
\begin{minipage}[c]{0.09\textwidth}
\includegraphics[width=1.6cm, height=1.4cm]{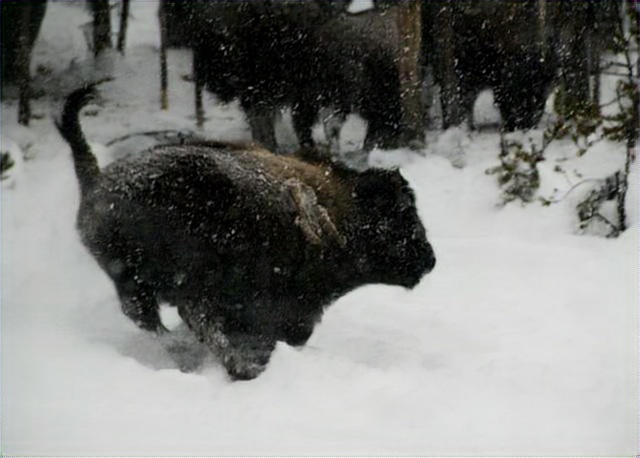}
\end{minipage}
\begin{minipage}[c]{0.09\textwidth}
\includegraphics[width=1.6cm, height=1.4cm]{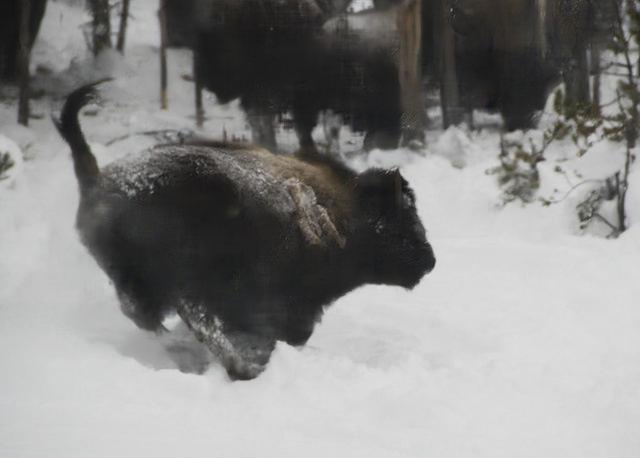}
\end{minipage}

\begin{minipage}[c]{1\textwidth}
\end{minipage}

\begin{minipage}[c]{0.09\textwidth}
\includegraphics[width=1.6cm, height=1.4cm]{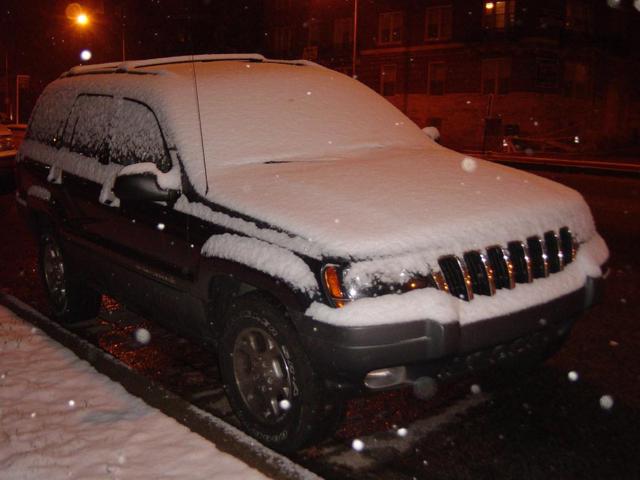}
\end{minipage}
\begin{minipage}[c]{0.09\textwidth}
\includegraphics[width=1.6cm, height=1.4cm]{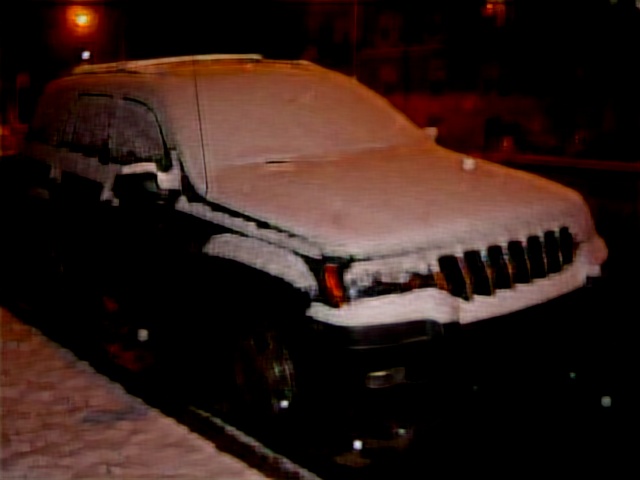}
\end{minipage}
\begin{minipage}[c]{0.09\textwidth}
\includegraphics[width=1.6cm, height=1.4cm]{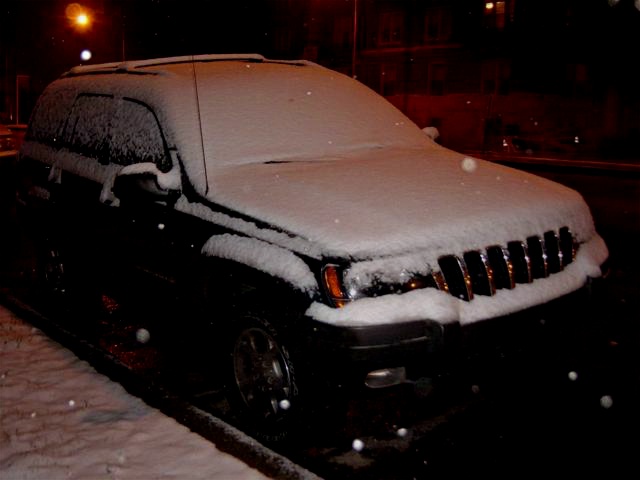}
\end{minipage}
\begin{minipage}[c]{0.09\textwidth}
\includegraphics[width=1.6cm, height=1.4cm]{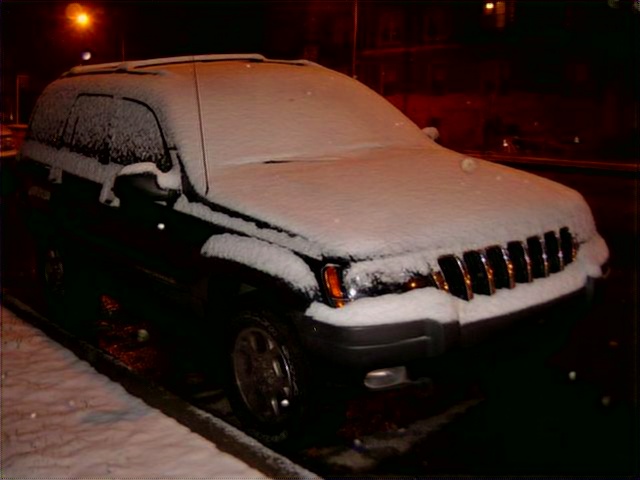}
\end{minipage}
\begin{minipage}[c]{0.09\textwidth}
\includegraphics[width=1.6cm, height=1.4cm]{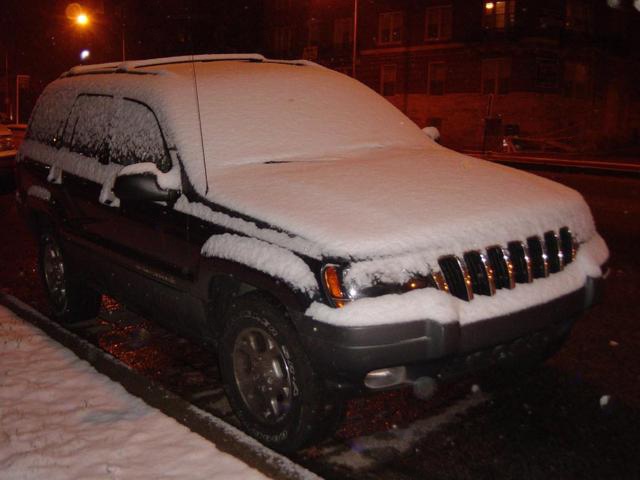}
\end{minipage}
\caption{Examples of real snow image categories in the questionnaire survey and the snow removal results corresponding to SMGARN. Column 1 represents snow images, and columns 2 to 5 correspond to DehazeNet, RESCAN, HDCWNet and our proposed SMGARN, respectively.} 
\label{Questionnaire}
\end{figure}

To evaluate the proposed model more objectively from the visual aspect, we conduct a user study on real snow images. Specifically, we collected 80 images from the Snow100K real snow dataset and our proposed SnowWorld24 dataset, and invited 25 participants for evaluation. We set three evaluation indicators as the basis for the participants to score, namely, the ability to remove large snowflakes, the ability to remove dense snowflakes, and the image quality. The 80 test images cover various scenes in the real world, including buildings, streets, crowds, and animals. In Fig.~\ref{Questionnaire}, we shows four sets of examples. In addition, we evaluate DehazeNet, RESCAN, HDCWNet, and our SMGARN as objects. Specific results are listed in TABLE~\ref{Questionnaire_t}. In the survey, more than half of the participants believed that the snow-free images reconstructed by SMGARN are closer to the real situation. Some participants believed that the snow-free image of SMGARN will not destroy the scene in the original image, such as the fourth row in Fig.~\ref{Questionnaire}. Meanwhile, the snow-free image generated by the four methods only SMGARN retains the details of the wheel hub in the lower right corner of the image . Therefore, in TABLE ~\ref{Questionnaire_t}, our method achieves the highest score. This proves that SMGARN has excellent performance in real scenes and can greatly preserve the original information of the image.

\begin{figure}
\centering
\begin{minipage}[c]{0.23\textwidth}
\includegraphics[width=4cm, height=2.8cm]{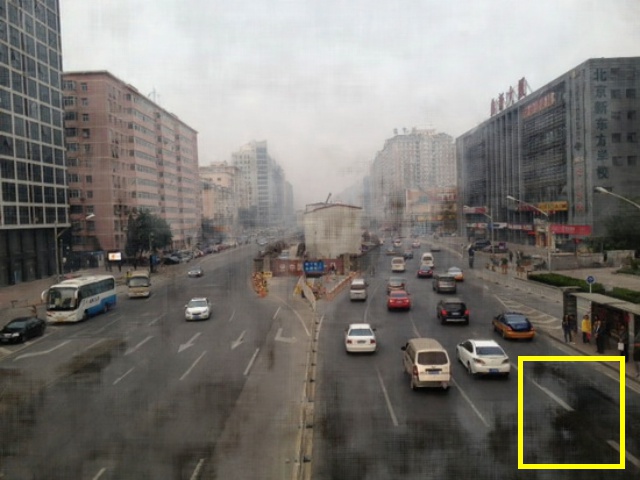}
\end{minipage}
\begin{minipage}[c]{0.23\textwidth}
\includegraphics[width=4cm, height=2.8cm]{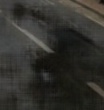}
\end{minipage}
\caption{Example of our reconstructed image with artifacts.} 
\label{discuss}
\end{figure}

\subsection{Application on High-Level Tasks}
\begin{figure}
\begin{minipage}[c]{0.24\textwidth}
\includegraphics[width=4.2cm]{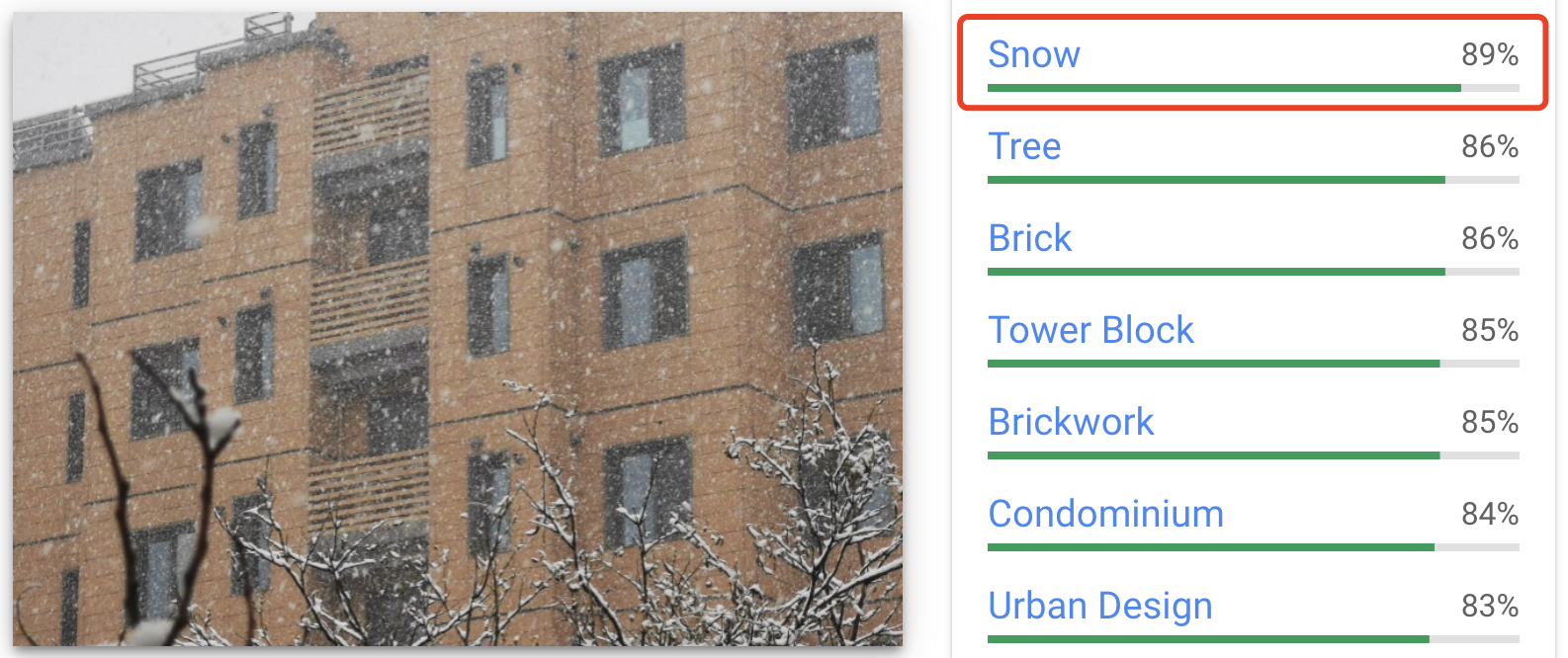}
\end{minipage}
\begin{minipage}[c]{0.24\textwidth}
\includegraphics[width=4.2cm]{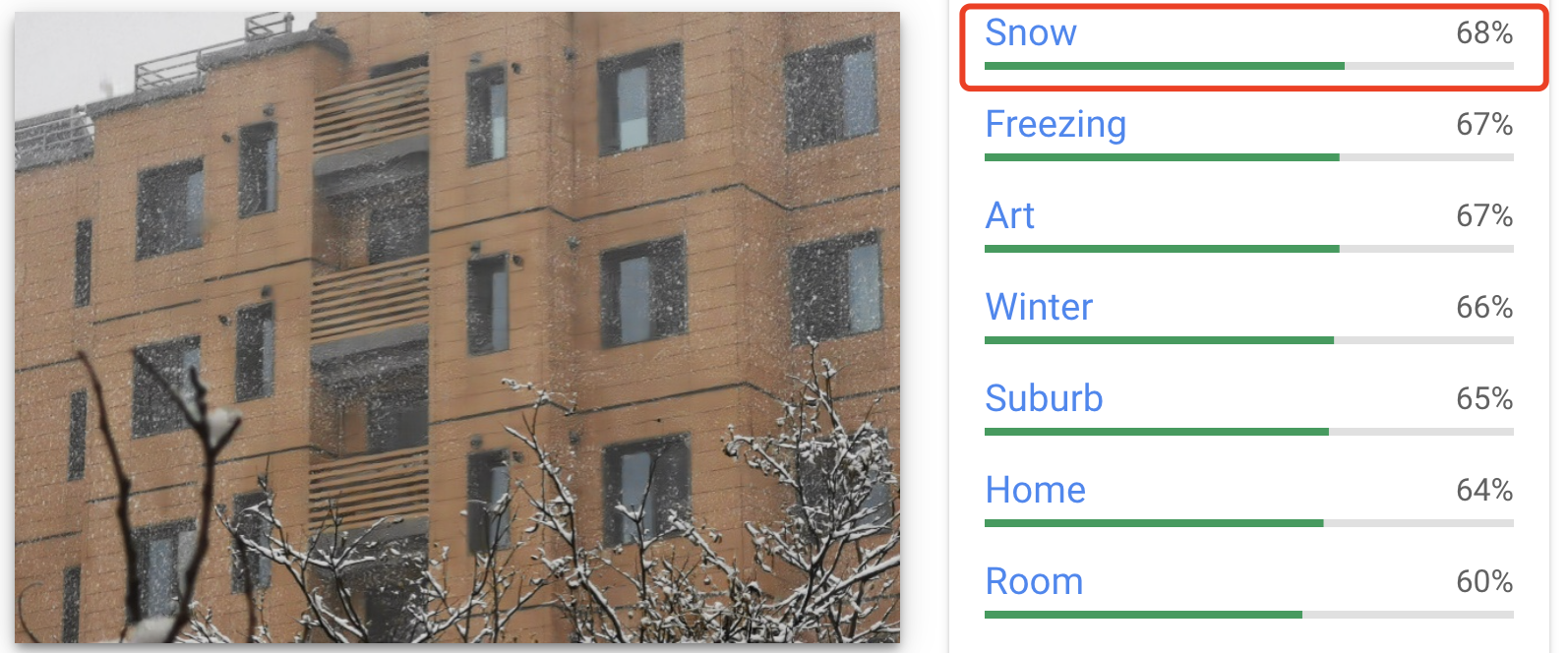}
\end{minipage}

\begin{minipage}[c]{\textwidth}
\end{minipage}

\begin{minipage}[c]{0.24\textwidth}
\includegraphics[width=4.2cm]{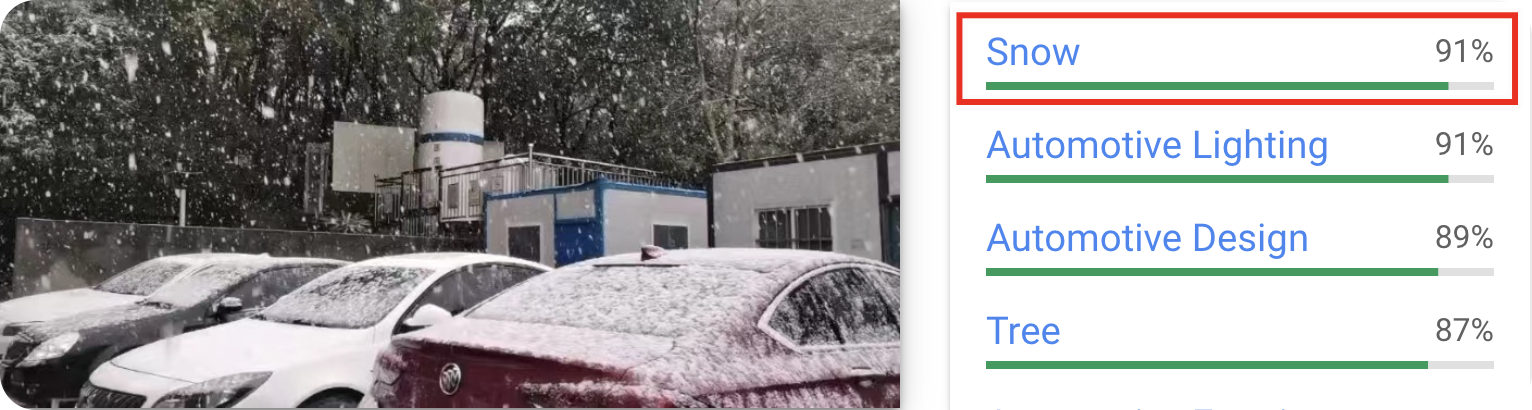}
\centerline{Snowy Image}
\centerline{}
\end{minipage}
\begin{minipage}[c]{0.24\textwidth}
\includegraphics[width=4.2cm]{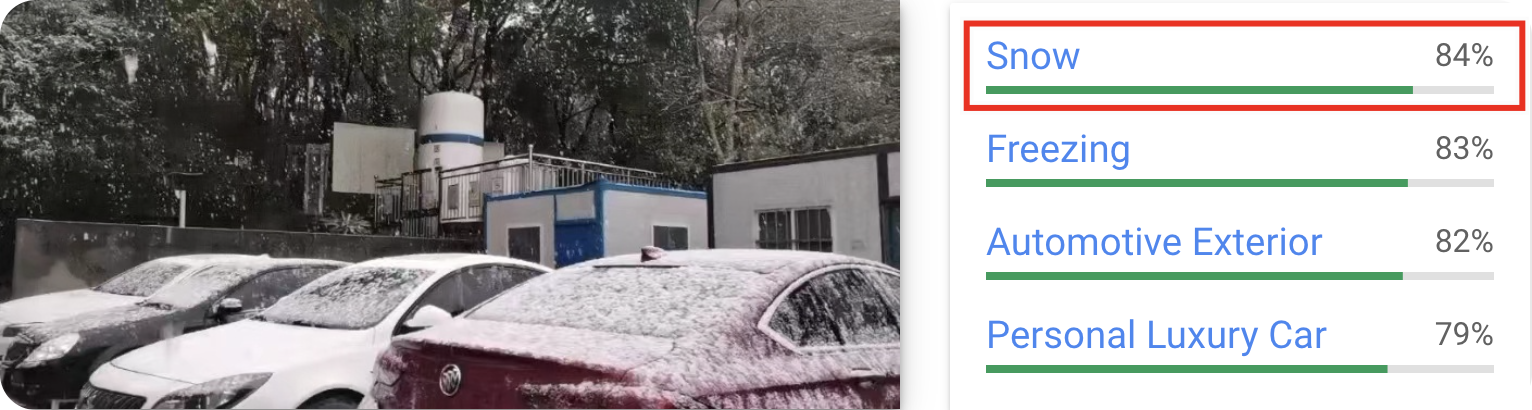}
\centerline{SMGARN (Ours)}
\centerline{}
\end{minipage}
\caption{The desnowing results are tested on the Google Vision API (snow classification). Please zoom in to see the details.}
\label{API-1}
\end{figure}

\begin{figure}
\begin{minipage}[c]{0.24\textwidth}
\includegraphics[width=4.2cm]{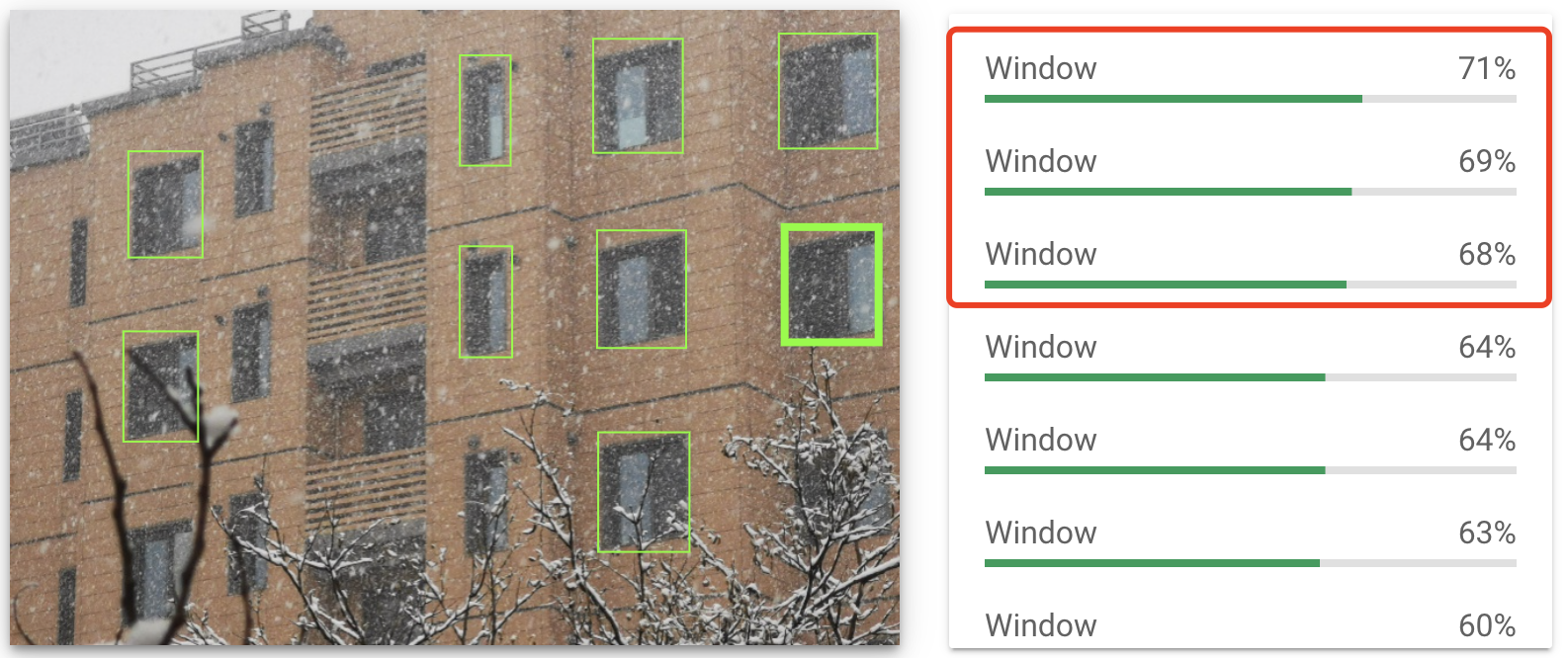}
\end{minipage}
\begin{minipage}[c]{0.24\textwidth}
\includegraphics[width=4.2cm]{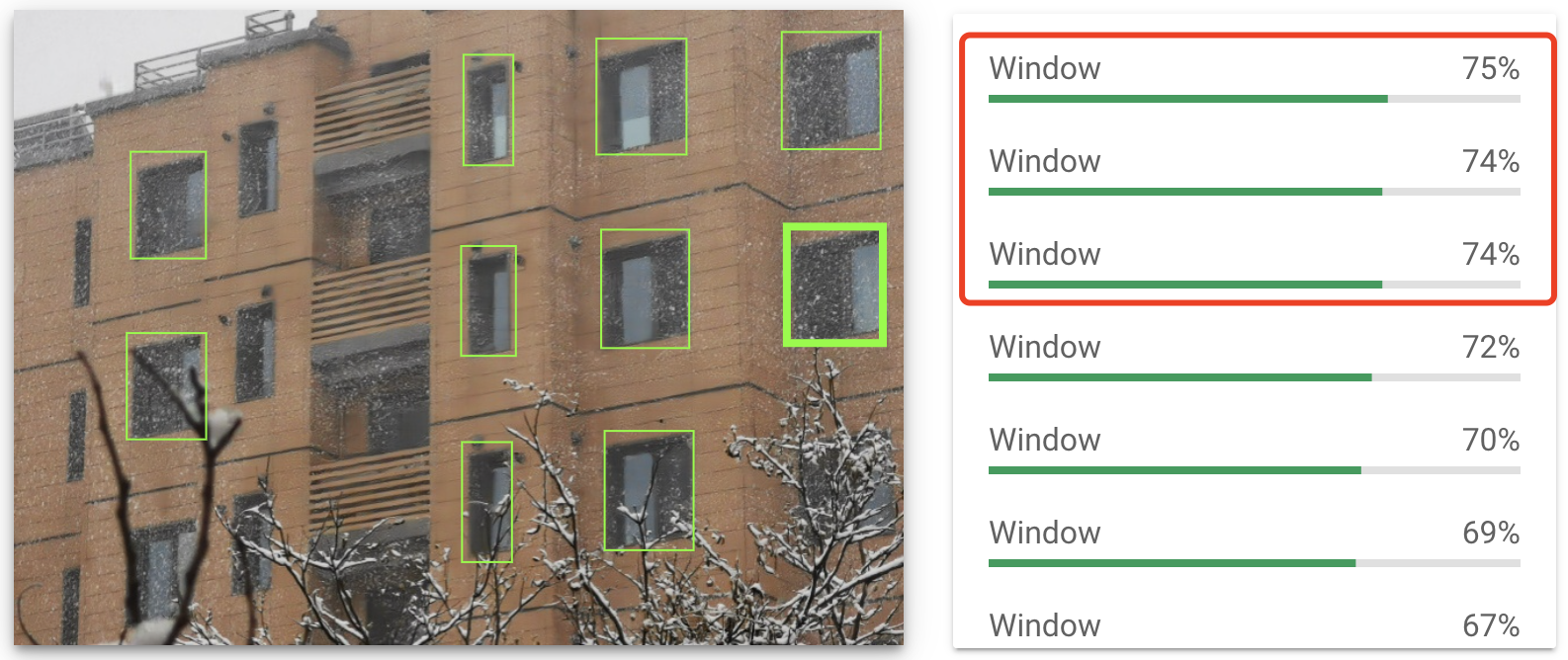}
\end{minipage}

\begin{minipage}[c]{1\textwidth}
\end{minipage}

\begin{minipage}[c]{0.24\textwidth}
\includegraphics[width=4.2cm]{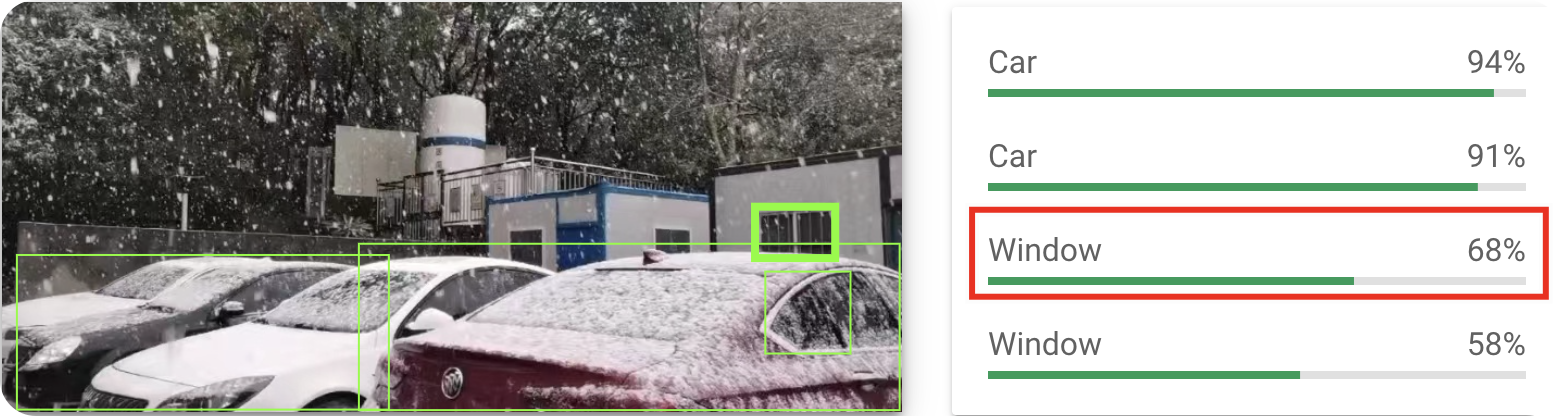}
\centerline{Snowy Image}
\end{minipage}
\begin{minipage}[c]{0.24\textwidth}
\includegraphics[width=4.2cm]{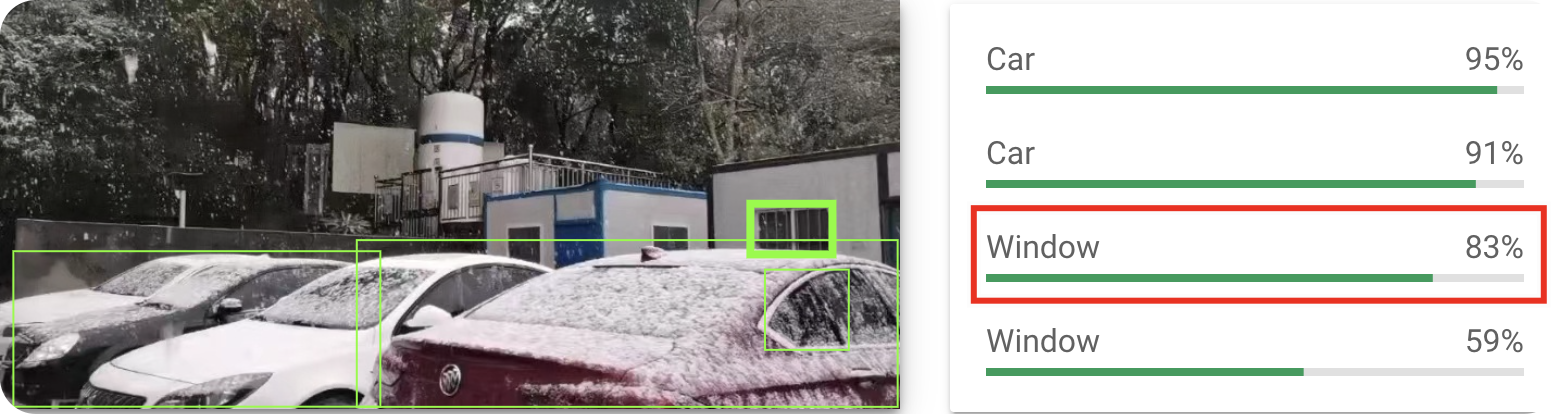}
\centerline{SMGARN (Ours)}
\end{minipage}
\caption{The desnowing results are tested on the Google Vision API (object detection). Please zoom in to see the details.}
\label{API-2}
\end{figure}

Image snow removal aims to restore clear images from images disturbed by snow phenomenon. It is not only a part of image processing, but also serves as a data preprocessing method for other vision tasks. To demonstrate that our proposed SMGARN can benefit high-level computer vision tasks, we use Google Vision API to evaluate the snow removal results. As a new evaluation method, it has been applied to image rain and snow removal to verify model performance from an application point of view. From Fig.~\ref{API-1}, it can be seen that the probability of SMGARN-processed images being judged as snow drops significantly. Meanwhile, according to Fig.~\ref{API-2}, the images reconstructed by our SMGARN can improve the accuracy of the object detection model. The confidence represents the probability of snowy weather and the probability of target, respectively. According to these results, we can observe that the results generated by our SMGARN effectively promote the performance of high-level vision tasks, which further demonstrates the effectiveness of SMGARN.

\section{Discussion}~\label{DS}
Although our SMGARN has obvious improvement in both quantification performance and visual performance, there are artifacts that cannot be removed in some images heavily affected by the veiling effect, as shown in Fig.~\ref{discuss}. This is because SMGARN lacks sufficient constraints for the haze phenomenon caused by the veiling effect. However, if too many constraints are imposed, the phenomenon of image blurring in HDCWNet will occur. Therefore, we believe that image desnowing should be regarded as an image restoration task covering a variety of extreme weather phenomena, which needs to be modeled by combining the characteristics of image dehazing and image deraining. Meanwhile, a new training dataset is also critical to the image snow removal field. In future works, we will make more contributions to this field.

\section{Conclusion}
In this paper, we proposed a Snow Mask Guided Adaptive Residual Network (SMGARN) for image snow removal. Specifically, we build a Mask-Net to directly predict the snow mask from the snowy image with the help of Self-pixel Attention (SA) and Cross-pixel Attention (CA) mechanisms. Meanwhile, a multi-level Guidance-Fusion Network (GF-Net) is designed to adaptively remove snow from the image with the guidance of the predicted snow mask. In addition, a Reconstruct-Net is proposed to reconstruct the final snow-free image by the specially designed Multi-scale Aggregated Residual Blocks (MARBs). Extensive experimental results show that the proposed SMGARN achieves excellent results on multiple datasets, and also shows great potential in practical applications. Although the current models and strategies have achieved excellent results, further improvements can be made in terms of visual presentation and training methods. In the future, we will explore better image prior guidance strategies to further improve model performance. 

\bibliographystyle{unsrt}
\bibliography{ref}

\end{document}